\documentclass[12pt]{article}
\usepackage{arxiv}

\usepackage[utf8]{inputenc} %
\usepackage[T1]{fontenc}    %
\usepackage{booktabs}       %
\usepackage{amsfonts}       %
\usepackage{nicefrac}       %
\usepackage{microtype}      %

\usepackage[numbers,sort&compress]{natbib}
\usepackage{doi}
\usepackage{algorithm,algpseudocode}
\usepackage{amsmath}
\usepackage{amssymb}
\usepackage{mathtools}
\usepackage{amsthm}
\usepackage{booktabs,multirow,threeparttable,tabularx}
\usepackage{tabularray}
\usepackage[font=normalsize,skip=4pt]{caption}
\usepackage{subcaption}
\usepackage{enumerate}

\DeclareMathOperator{\diag}{diag}
\DeclareMathOperator{\cov}{cov}
\DeclareMathOperator*{\argmax}{arg\,max}

\theoremstyle{thmstyleone}%
\newtheorem{theorem}{Theorem}%

\theoremstyle{thmstyletwo}

\theoremstyle{thmstylethree}

\usepackage[dvipsnames]{xcolor}
\usepackage{graphicx}   %
\usepackage{hyperref}   %
\hypersetup{
    colorlinks,
    linkcolor={red!60!black},
    citecolor={green!60!black},
    urlcolor={blue!60!black}
}
\usepackage[capitalize,nameinlink]{cleveref}

\title{Sampling from Gaussian Processes: A Tutorial and Applications in Global Sensitivity Analysis and Optimization}

\author{{\hspace{1mm}Bach Do} \\
	University of Houston\\
	\texttt{bdo3@uh.edu} \\
	\And
{\hspace{1mm}Nafeezat Ajenifuja} \\
	University of Houston\\
	\texttt{naajenif@uh.edu} \\
        \And
{\hspace{1mm}Taiwo Adebiyi} \\
	University of Houston\\
	\texttt{taadebiyi2@uh.edu} \\
        \And
 {\hspace{1mm}Ruda Zhang} \\
	University of Houston\\
	\texttt{rudaz@uh.edu} \\
}

\date{}

\renewcommand{\headeright}{ }
\renewcommand{\shorttitle}{Sampling from Gaussian Processes: A Tutorial}

\begin{document}
\maketitle

\begin{abstract}
High-fidelity simulations and physical experiments are essential for engineering analysis and design, yet their high cost often makes two critical tasks--global sensitivity analysis (GSA) and optimization--prohibitively expensive.
This limitation motivates the common use of Gaussian processes (GPs) as proxy regression models that provide uncertainty-aware predictions from a limited number of high-quality observations. GPs naturally enable efficient sampling strategies that support informed decision-making under uncertainty by extracting information from a subset of possible functions for the model of interest.
However, direct sampling from GPs is inefficient due to their infinite-dimensional nature and the high cost associated with large covariance matrix operations.
Despite their popularity in machine learning and statistics communities, sampling from GPs has received little attention in the community of engineering optimization.
In this paper, we present the formulation and detailed implementation of two notable sampling methods--random Fourier features and pathwise conditioning--for generating posterior samples from GPs at reduced computational cost.
Alternative approaches are briefly described.
Importantly, we detail how the generated samples can be applied in GSA, single-objective optimization, and multi-objective optimization.
We show successful applications of these sampling methods through a series of numerical examples.

\end{abstract}


\def\thefootnote{}\footnotetext{This is the accepted version of the following article: Do, B., Ajenifuja N., Adebiyi T., Zhang, R. Sampling from Gaussian processes: A tutorial and applications in global sensitivity analysis and optimization. Structural and Multidisciplinary Optimization (2025).}\def\thefootnote{\arabic{footnote}}

\clearpage

\section{Introduction}
\label{Sec1}

Gaussian process (GP) regression \citep{Rasmussen2006}, known as Kriging in geostatistics \citep[Chapter 3]{Chiles1999}, has become a valuable tool for modeling complex systems across various disciplines of science and engineering \citep[see e.g.,][]{Kennedy2001,Simpson2001,Forrester2008,Solak2002,Higdon2004,Raissi2017,ChenY2021,ZhangRD2022gps}.
Its strength lies in the ability to construct analytical probabilistic regression models that provide uncertainty-aware predictions under the lack of reliable observations, while requiring only a few hyperparameters.
This modeling capability makes GP useful for many critical tasks in engineering analysis and design, including global sensitivity analysis \citep[see e.g.,][]{Oakley2004,ChenW2004,Marrel2009,LeGratiet2014,LeGratiet2016,Belakaria2024} and optimization  \citep[see e.g.,][]{Jones1998,Shahriari2016,Frazier2018,Do2025photonics}.

We consider a dataset that
consists of several pairs of input variable values
and the corresponding observations of an engineering model. This model is considered a black-box function of the input variables.
While many possible functions can fit the dataset for a mapping between the input variables and the model output, GPs learn a distribution over these functions, assuming that any finite set of function values follows a multivariate Gaussian distribution, which is known as the GP prior.
The GP prior is determined by a mean function and a covariance function.
By combining their prior with the observations via Bayes' rule, GPs refine the set of possible functions, resulting in posterior mean and covariance functions that define GP posteriors.
The GP posteriors not only allow for predictions of the function values at unseen input values but also quantify uncertainty in these predictions.

A successful application of GPs in the engineering sphere is for \textit{global sensitivity analysis} (GSA) that studies how variations in the input variables influence a model output, considering all possible values of the input variables \citep{Saltelli2008,Saltelli2010}.
One of the notable forms of GSA is that of %
Sobol' indices (i.e., variance-based indices) which rely on the decomposition of the output variance into the contributions of the input variables and their potential interactions \citep{Sobol1993,Sobol2001}.
Due to the substantial amount of model evaluations required for computing Sobol' indices, using GPs in place of high-fidelity models can reduce the computational cost while providing a sense of estimation accuracy \citep[see e.g.,][]{Marrel2009,LeGratiet2016,ChengK2020,Chauhan2024}.
Another form of GSA, which is not the focus of this work, is that of derivative-based global sensitivity
measures \citep{Sobol2009,Kucherenko2009} which rely on the global variability of the gradient of output functions. In this axis, GPs can also serve as surrogates for expensive-to-evaluate output functions to estimate their gradients with and without an active learning strategy \citep[see e.g.,][]{Matthias2016,Belakaria2024}.

There are two primary approaches to calculating Sobol' indices via GPs: the GP mean-based and global GP-based approaches \citep{Marrel2009}.
The former uses the GP mean as a proxy for the model that provides a deterministic estimate of each Sobol's index for each input variable.
For this estimate to be reliable, the GP mean function must accurately capture the model behavior over the entire input domain.
This however requires a large number of observations, leading to a high training cost.
The latter uses both the prediction and uncertainty quantification from GPs for the mean and variance of each Sobol' index. Specifically, the indices are computed from multiple realizations of the model output at a set of grid points of input variables generated via the GP mean and the Cholesky decomposition of the GP covariance matrix at these points \citep{Marrel2009}. 
However, as we will show in \Cref{sec4.4}, this approach is computationally demanding when a large set of input variables is required for accurately capturing their variations.

The strength of GPs also enables the development of many GP-based \textit{optimization methods} that enjoy strong convergence guarantees \citep[Chapter 10]{Garnett2023}.
These methods are members of a big family of Bayesian optimization (BO) \citep[see e.g.,][]{Jones1998,Shahriari2016,Frazier2018,Garnett2023,ZhangRD2024mfml} and multi-objective Bayesian optimization (MOBO) algorithms \citep[see e.g.,][]{Knowles2006,Couckuyt2012,Bradford2018,KonakovicLukovic2020,Daulton2020,Mathern2021,Daulton2022}.
Each of them iteratively recommends new candidate solutions for optimization stages via optimizing an acquisition function that quantifies how valuable each candidate point in the input variable space is, based on what we value in the current observations.

Each acquisition function typically addresses the classic tension between exploitation and exploration in optimization. Here, exploitation focuses the search on regions where good solutions have been observed, while exploration directs the search to unexplored regions that may hide the global solution.
In the context of GP-based optimization, exploitation often refers to the posterior mean function, which interpolates the observations, while exploration is associated with the posterior covariance function, which can detect the regions that lack observations.
Many notable acquisition functions have been developed based on the posterior mean function and/or the posterior covariance function.
For example, the prediction-based acquisition function is defined as the posterior mean function.
The active learning--MacKay and active learning--Cohn acquisition functions \citep{Gramacy2009} are based on the posterior covariance function.
The expected improvement acquisition function \citep{Jones1998} balances exploitation and exploration using an equal weighted-sum of a posterior mean-based term and a posterior standard deviation-based term.
The GP upper confidence bound acquisition function \citep{Srinivas2010} uses the upper prediction bound by adding a scheduled proportion of the posterior standard deviation to the posterior mean. 

Apart from the posterior mean and covariance functions, sample paths drawn from GP posteriors have promised sensible approaches to GSA and optimization.
For GSA, a sample path from the GP posterior of the model of interest can be used as a proxy for the model to compute the output realizations at a set of input variable points. With these realizations, the mean
and variance of each sensitivity index can be evaluated \citep{LeGratiet2014,LeGratiet2016}.
This approach overcomes the computational limitation associated with sampling a large number of output realizations directly from the corresponding GP mean vector and covariance matrix.
For optimization, sample paths from the GP posterior of the objective function can serve as stochastic acquisition functions in BO, which are known as GP Thompson sampling (GP-TS) acquisition functions \citep{May2012,Russo2014,Chowdhury2017,Mutny2018}.
GP-TS naturally addresses the exploration-exploitation dilemma by accumulating observations during the optimization process \citep{Do2024jcise} and is known for its asymptotic convergence with some regularity \citep{May2012}.
It underpins scalable BO algorithms \citep[see e.g.,][]{Eriksson2019,Daulton2022} for high-dimensional optimization problems and efficient evolutionary strategies for multi-objective optimization \citep{Bradford2018}.
GP-TS also serves as the engine for computing powerful information-theoretic acquisition functions \citep{HernandezLobato2014,WangZ2017,Hvarfner2022}, and for parallelizing BO \citep{Kandasamy2018}.

Although the applications of the aforementioned sampling approaches in GSA and optimization arise naturally because GPs are inherently distributions of functions, they are underutilized within the community of engineering optimization.
Therefore, our hope in writing this educational paper is to catalyze research on these simple yet powerful approaches and their exciting applications in GSA and engineering optimization.

This paper presents the mathematical foundation and detailed implementation of two notable sampling methods, namely the random Fourier feature method \citep{Rahimi2007} and pathwise conditioning \citep{Wilson2020}, to generate approximate sample paths from GP posteriors.
Alternative sampling methods are briefly described.
The strengths and weaknesses of the random Fourier feature and pathwise conditioning methods are discussed.
Importantly, their applications in GSA, single-objective optimization, and multi-objective optimization are detailed and empirically investigated.

The remainder of this paper progresses as follows.
\Cref{sec2} describes sampling from multivariate Gaussian distributions and the concept of conditional Gaussian distributions in finite-dimensional spaces.
The latter forms the foundation of GPs presented in \cref{sec3}, where their function- and weight-space views are discussed. 
Built upon \cref{sec2,sec3}, \Cref{sec4} presents the random Fourier feature method and pathwise conditioning while briefly discussing other sampling methods. 
This section then compares the prediction performance and computational costs of the random Fourier feature method and pathwise conditioning against those of the standard exhaustive sampling method, which samples the GP posterior at a discrete set of locations.
Sections~\ref{sec5}, \ref{sec6}, and \ref{sec7} discuss how GP sample paths can be used for 
GSA, single-objective optimization, and multi-objective optimization, respectively.
\Cref{sec8} presents a series of numerical examples to show successful applications of GP sample paths in these tasks.
\Cref{sec9} summarizes key takeaways from this paper.
\Cref{sec10} discusses additional implementation considerations for the presented GP sampling methods.
Finally, \Cref{sec11} concludes the paper with several potential research directions.

\section{Sampling from multivariate and conditional Gaussian distributions}
\label{sec2}

This section presents sampling from multivariate Gaussian distributions and the concept of conditional Gaussians distributions in finite-dimensional spaces.
The former and latter build the foundation of sampling from GPs in \cref{sec4} and conditioning of GPs in \cref{sec3}, respectively.

\subsection{Sampling from multivariate Gaussian distributions}
\label{sec21}

Let ${\bf f} \sim \mathcal{N}\left( {\bf m}, \boldsymbol{\Sigma}\right) \in \mathbb{R}^{N_\text{f}}$ be a real-valued vector that follows an $N_\text{f}$-variate Gaussian distribution, where ${\bf m} \in \mathbb{R}^{N_\text{f}}$ is the mean vector and $\boldsymbol{\Sigma} \in \mathbb{R}^{N_\text{f} \times N_\text{f}}$ the positive-definite covariance matrix.
A standard method to draw random samples of 
${\bf f}$ is via the Cholesky decomposition of $\boldsymbol{\Sigma}$, which consists of three steps.
(1) Randomly generate samples ${\bf e}$ from the $N_\text{f}$-variate standard Gaussian distribution $\mathcal{N}\left( {\bf 0}, {\bf I}\right)$, where ${\bf I}$ is the identity matrix.
(2) Factorize $\boldsymbol{\Sigma}$ using the Cholesky decomposition.
This results in a lower triangular matrix ${\bf L}$ such that ${\bf L} {\bf L}^\intercal = \boldsymbol{\Sigma}$, where $(\cdot)^\intercal$ denotes the transpose operator.
(3) Obtain samples of ${\bf f}$ via the following composition of affine transformations:
\begin{equation}\label{eqn:SamplingGaussian}
    {\bf f} = {\bf m} + {\bf L} {\bf e}.
\end{equation}

The proof of \cref{eqn:SamplingGaussian} is straightforward. We can recover the mean and covariance of ${\bf f}$ by computing the mean and covariance of ${\bf m} + {\bf L} {\bf e}$.
Specifically, $\mathbb{E}\left[{\bf m}+ {\bf L} {\bf e} \right] = {\bf m} + {\bf L} \mathbb{E}\left[ {\bf e}\right] = {\bf m}$ and $\cov\left({\bf m}+ {\bf L} {\bf e}, {\bf m}+ {\bf L} {\bf e}\right) = {\bf L}  \cov({\bf e},{\bf e}) {\bf L}^\intercal = {\bf L} {\bf L}^\intercal = \boldsymbol{\Sigma}$, where $\mathbb{E}[\cdot]$ and $\cov(\cdot,\cdot)$ are the mean and covariance operators, respectively.

\begin{figure*}[t]
	\centering
	\includegraphics[width=\textwidth]{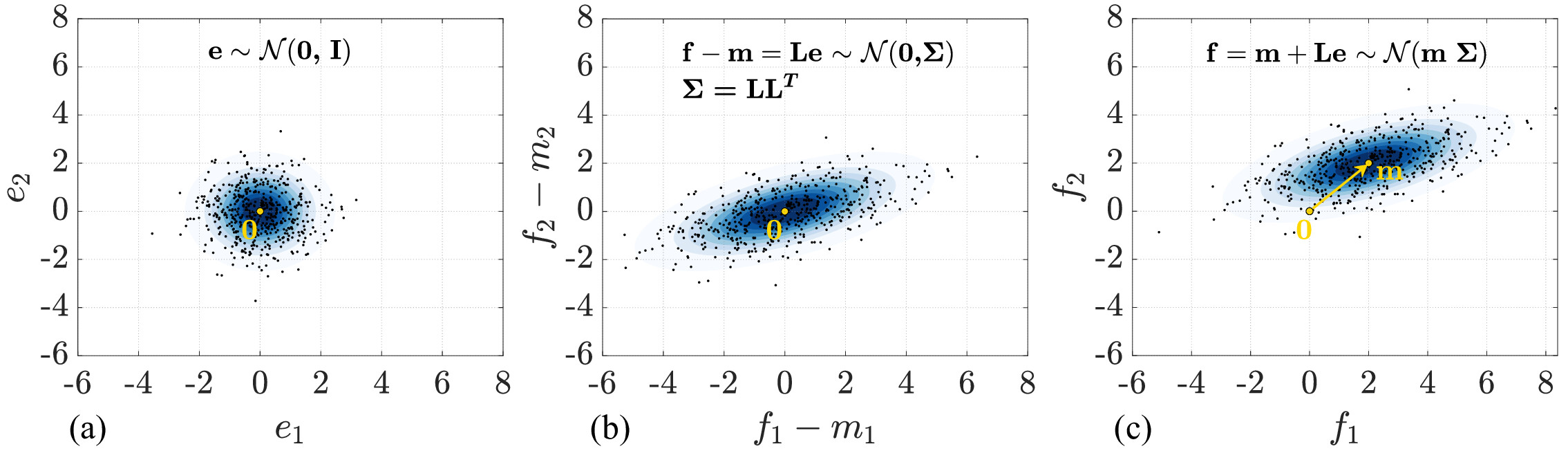}
	\caption{Samples from a bivariate Gaussian distribution $\mathcal{N}\left( {\bf m}, \boldsymbol{\Sigma}\right)$ generated by a composition of affine transformations of samples from the standard Gaussian distribution $\mathcal{N}\left( {\bf 0}, {\bf I}\right)$. (a) Samples from $\mathcal{N}\left( {\bf 0}, {\bf I}\right)$. (b) Samples from $\mathcal{N}\left( {\bf 0}, \boldsymbol{\Sigma}\right)$ obtained by a linear transformation of samples in (a), where the transformation matrix is from the Cholesky decomposition of $\boldsymbol{\Sigma}$. (c) Samples from $\mathcal{N}\left( {\bf m}, \boldsymbol{\Sigma}\right)$ obtained by shifting the samples from (b) in the direction of ${\bf m}$.}
    \label{fig:GaussianSampling}
\end{figure*}

\Cref{fig:GaussianSampling} illustrates the aforementioned three steps to generate samples from a bivariate  Gaussian distribution $\mathcal{N}\left( {\bf m}, \boldsymbol{\Sigma}\right)$.
Samples ${\bf e}$ are drawn from $\mathcal{N}\left( {\bf 0}, {\bf I}\right)$ (\cref{fig:GaussianSampling}(a)).
These samples are then transformed by the linear mapping ${\bf L} {\bf e}$, resulting in samples from $\mathcal{N}\left( {\bf 0}, \boldsymbol{\Sigma}\right)$ (\cref{fig:GaussianSampling}(b)).
Adding ${\bf m}$ to the transformed samples finally provides the samples from $\mathcal{N}\left( {\bf m}, \boldsymbol{\Sigma}\right)$ (\cref{fig:GaussianSampling}(c)).

\subsection{Conditional Gaussian distributions}
\label{sec22}

Without loss of generality, let ${\bf f}_1$ denote the first $N_{\text{f}_1}$ components of ${\bf f}$ and ${\bf f}_2$ the remaining $N_{\text{f}_2}$ components, where $N_{\text{f}_1} + N_{\text{f}_2} = N_\text{f}$. 
We can rewrite ${\bf f} \sim \mathcal{N}\left( {\bf m}, \boldsymbol{\Sigma}\right)$ as the following joint Gaussian distribution of ${\bf f}_1$ and ${\bf f}_2$:
\begin{equation} \label{eq:jointGaussian}
    {\bf f} = \begin{bmatrix}
    {\bf f}_1 \\
    {\bf f}_2
    \end{bmatrix}
    \sim
    \mathcal{N} \left( \begin{bmatrix}
     {\bf m}_1\\
    {\bf m}_2
    \end{bmatrix}, \begin{bmatrix}
    \boldsymbol{\Sigma}_{11} &  \boldsymbol{\Sigma}_{12}\\
    \boldsymbol{\Sigma}_{21} &  \boldsymbol{\Sigma}_{22}
    \end{bmatrix}\right),
\end{equation}
where ${\bf m}_1 \in \mathbb{R}^{N_{\text{f}_1}}$ and ${\bf m}_2 \in \mathbb{R}^{N_{\text{f}_2}}$ are marginal mean vectors of ${\bf f}_1$ and ${\bf f}_2$, respectively, $\boldsymbol{\Sigma}_{11} \in \mathbb{R}^{N_{\text{f}_1} \times N_{\text{f}_1}}$ and $\boldsymbol{\Sigma}_{22} \in \mathbb{R}^{N_{\text{f}_2} \times N_{\text{f}_2}}$ the corresponding marginal covariance matrices, and $\boldsymbol{\Sigma}_{12} = \boldsymbol{\Sigma}_{21}^\intercal \in \mathbb{R}^{N_{\text{f}_1} \times N_{\text{f}_2}}$ the cross-covariance matrix between ${\bf f}_1$ and ${\bf f}_2$.
Here, ``marginal" arises from what we call the marginalization of the joint Gaussian distribution in \cref{eq:jointGaussian} that allows for extracting the distributions of ${\bf f}_1$ and ${\bf f}_2$ using the corresponding means and covariance matrices.
Accordingly, ${\bf f}_1 \sim \mathcal{N}\left( {\bf m}_1, \boldsymbol{\Sigma}_{11}\right)$ and ${\bf f}_2 \sim \mathcal{N}\left( {\bf m}_2, \boldsymbol{\Sigma}_{22}\right)$.

From \cref{eq:jointGaussian}, we can also determine the distribution of ${\bf f}_2$ given ${\bf f}_1$, known as the conditional distribution.
This conditional distribution, denoted as ${\bf f}_2|{\bf f}_1 \sim \mathcal{N} \left( 
\boldsymbol{\mu}_{2|1}, \boldsymbol{\Sigma}_{2|1}\right)$, is an $N_{\text{f}_2}$-variate Gaussian distribution whose mean vector and covariance matrix are given as follows:
\begin{equation} \label{eqn:conditionalmeancovariance}
	\begin{aligned}
		\boldsymbol{\mu}_{2|1} & = {\bf m}_2 + \boldsymbol{\Sigma}_{21} \boldsymbol{\Sigma}_{11}^{-1} ({\bf f}_1-{\bf m}_1), \\
		\boldsymbol{\Sigma}_{2|1} & = \boldsymbol{\Sigma}_{22} - \boldsymbol{\Sigma}_{21} \boldsymbol{\Sigma}_{11}^{-1} \boldsymbol{\Sigma}_{12}.
	\end{aligned}
\end{equation}
A formal derivation of these expressions can be found in \cite{Bishop2006}, Section 2.3.1.
Appendix~\ref{AppA} presents a geometric way to derive the conditional mean and conditional variance from bivariate Gaussian distributions.
We see from \cref{eqn:conditionalmeancovariance} that $\boldsymbol{\mu}_{2|1}$ depends on the conditioned variables ${\bf f}_1$, while $\boldsymbol{\Sigma}_{2|1}$ does not.

Once the conditional Gaussian distribution is settled, we are ready draw random samples from it.
Consider the conditional Gaussian distribution of ${\bf f}_2|{\bf f}_1 = \boldsymbol{\beta}$, where $\boldsymbol{\beta} \in \mathbb{R}^{N_{\text{f}_1}}$ is a specific value of the random variable ${\bf f}_1$.
We can follow two main approaches to generating samples from this conditional distribution. The first approach relies on \cref{eqn:SamplingGaussian} that requires computing the conditional mean and the conditional covariance matrix in \cref{eqn:conditionalmeancovariance}, followed by the Cholesky decomposition of the conditional covariance matrix.

The second approach is via Matheron's rule that updates samples generated from the joint Gaussian distribution ${\bf f} \sim \mathcal{N}\left( {\bf m}, \boldsymbol{\Sigma}\right)$ to obtain the corresponding samples from the conditional Gaussian distribution of ${\bf f}_2|{\bf f}_1 = \boldsymbol{\beta}$ (see e.g., \cite{Chiles1999}, Section 7.3; \cite{Wilson2020}). This approach has its roots in the following distribution equality:
\begin{equation}\label{eqn:distributionequality}
 	({\bf f}_2|{\bf f}_1 = \boldsymbol{\beta}) \overset{\text{d}}{=} {\bf f}_2 + \boldsymbol{\Sigma}_{21} \boldsymbol{\Sigma}_{11}^{-1} (\boldsymbol{\beta} -{\bf f}_1),
\end{equation}
where $\overset{\text{d}}{=}$ denotes an equality in distribution, the left-hand side represents a sample from the distribution of ${\bf f}_2|{\bf f}_1 = \boldsymbol{\beta}$, and ${\bf f}_1$ and ${\bf f}_2$ on the right-hand side are the marginal components of a sample generated from $\mathcal{N}\left( {\bf m}, \boldsymbol{\Sigma}\right)$.
We can prove this distribution equality by computing the mean and covariance of ${\bf f}_2 + \boldsymbol{\Sigma}_{21} \boldsymbol{\Sigma}_{11}^{-1} (\boldsymbol{\beta} -{\bf f}_1)$, such that 
\begin{equation}
	\begin{aligned}
		 \mathbb{E}\left[ {\bf f}_2 + \boldsymbol{\Sigma}_{21} \boldsymbol{\Sigma}_{11}^{-1} (\boldsymbol{\beta} -{\bf f}_1) \right] 
		& = \mathbb{E}[{\bf f}_2] + \boldsymbol{\Sigma}_{21} \boldsymbol{\Sigma}_{11}^{-1} (\boldsymbol{\beta} -\mathbb{E}[{\bf f}_1]) \\
		& = {\bf m}_2 + \boldsymbol{\Sigma}_{21} \boldsymbol{\Sigma}_{11}^{-1} (\boldsymbol{\beta} -{\bf m}_1) = \boldsymbol{\mu}_{{\bf f}_2|{\bf f}_1 = \boldsymbol{\beta}} ,
	\end{aligned}
\end{equation}
\begin{equation}
	\begin{aligned}
		& \cov \left( {\bf f}_2 + \boldsymbol{\Sigma}_{21} \boldsymbol{\Sigma}_{11}^{-1} (\boldsymbol{\beta} -{\bf f}_1), {\bf f}_2 + \boldsymbol{\Sigma}_{21} \boldsymbol{\Sigma}_{11}^{-1} (\boldsymbol{\beta} -{\bf f}_1) \right) \\
		& = \cov({\bf f}_2,{\bf f}_2) - 2 \boldsymbol{\Sigma}_{21} \boldsymbol{\Sigma}_{11}^{-1} \cov({\bf f}_1,{\bf f}_2)  
		 + \boldsymbol{\Sigma}_{21} \boldsymbol{\Sigma}_{11}^{-1} \cov({\bf f}_1,{\bf f}_1) \boldsymbol{\Sigma}_{11}^{-1} \boldsymbol{\Sigma}_{12} \\
		& = \boldsymbol{\Sigma}_{22} - 2 \boldsymbol{\Sigma}_{21} \boldsymbol{\Sigma}_{11}^{-1} \boldsymbol{\Sigma}_{12} + \boldsymbol{\Sigma}_{21} \boldsymbol{\Sigma}_{11}^{-1} \boldsymbol{\Sigma}_{12} \\
		& = \boldsymbol{\Sigma}_{22} -  \boldsymbol{\Sigma}_{21} \boldsymbol{\Sigma}_{11}^{-1} \boldsymbol{\Sigma}_{12}= \boldsymbol{\Sigma}_{{\bf f}_2|{\bf f}_1 = \boldsymbol{\beta}}.
	\end{aligned}
\end{equation}
The Matheron's rule approach does not require computing the conditional mean and conditional covariance matrix to draw conditional samples.

\section{Two perspectives on Gaussian processes}
\label{sec3}

Given a dataset $\mathcal{D} = \{ ({\bf X}, {\bf y}) \} = \left\{\left( {\bf x}^i, y^i \right) \right\}_{i=1}^N$, where ${\bf x}^i \in \mathbb{R}^d$ is an observation of the input variables and $y^i \in \mathbb{R}$ the observation at ${\bf x}^i$ of a model of interest $c({\bf x})$, which is considered a black-box function of ${\bf x}$.
We wish to learn $c({\bf x})$ from $\mathcal{D}$ and use the resulting model to predict $c({\bf x}_\xi)$, where ${\bf x}_\xi$ is any unseen input variable point.
Gaussian processes (GPs) allow us to build flexible and tractable probabilistic regression models that provide uncertainty-aware prediction of $c({\bf x}_\xi)$.
This section reviews GPs from both function- and weight-space perspectives \citep[Chapter 2]{Rasmussen2006}, each offering a distinct way to generate sample paths from GP posteriors. 
Hereafter, \textit{conditional} and \textit{posterior} are used interchangeably.

\subsection{Function-space view}
\label{sec31}
The function-space view interprets a GP as the generalization of a multivariate Gaussian distribution. While the multivariate Gaussian distribution describes the distribution of a finite-dimensional vector of random variables, the GP extends this concept to describe a distribution over functions. 
In other words, the GP assumes that the function values at any finite set of input variable points have a joint Gaussian, which is known as the GP prior.

Let $f({\bf x})$ denote a prior stochastic function for modeling $c({\bf x})$. 
Like a multivariate Gaussian distribution defined by its mean vector and covariance matrix, the GP prior is fully determined by its mean function $m({\bf x}) = \mathbb{E}\left[ f({\bf x}) \right]$ and covariance function $\kappa ({\bf x},{\bf x}')= \cov (f({\bf x}),f({\bf x}'))$, 
where ${\bf x}$ and ${\bf x}'$ are two arbitrary input points.
Notationally, the GP prior is written as
\begin{equation}
    f({\bf x}) \sim \mathcal{GP} \left( m({\bf x}), \kappa ({\bf x},{\bf x}') \right).
\end{equation}

While it can be any function, $m({\bf x})$ is typically set to zero, i.e., $m({\bf x}) = 0$, unless there is strong prior knowledge about the function suggesting a different form.
Meanwhile, the choice of $\kappa({\bf x},{\bf x}')$ determines important characteristics of the function such as its smoothness or periodicity. 
Different types of covariance functions and their properties are discussed in \cite{Rasmussen2006}, Chapter 4.
Illustrations and examples in this work use squared exponential (SE) (\cref{SEcovariancefunction}) and Matérn (\cref{Materncovariancefunction}) covariance functions.
\begin{align}
    \kappa({\bf x},{\bf x}'|\boldsymbol{\phi}_\kappa) & = \kappa(\boldsymbol{\delta}) = \sigma_\text{f}^2 \exp \left( -\frac{1}{2} \boldsymbol{\delta}^\intercal \boldsymbol{\Lambda}^{-1} \boldsymbol{\delta} \right), \label{SEcovariancefunction}\\
    \kappa({\bf x},{\bf x}'|\boldsymbol{\phi}_\kappa) & = \kappa_\nu(\boldsymbol{\delta}) = \sigma_\mathrm{f}^2 \frac{2^{1-\nu}}{\Gamma(\nu)} \left( \sqrt{2\nu} \sqrt{\boldsymbol{\delta}^\intercal \boldsymbol{\Lambda}^{-1} \boldsymbol{\delta}} \right)^\nu 
    K_\nu\left( \sqrt{2\nu} \sqrt{\boldsymbol{\delta}^\intercal \boldsymbol{\Lambda}^{-1} \boldsymbol{\delta}} \right), \label{Materncovariancefunction}
\end{align}
where $\boldsymbol{\phi}_\kappa = \left[ \sigma_\text{f}, l_1,\dots,l_d \right]^\intercal$ is the vector of hyperparameters, with the output scale $\sigma_\text{f}>0$ and the characteristic length scales $l_i>0$, $i \in \{1,\dots,d\}$, $\boldsymbol{\delta} = {\bf x}-{\bf x}' \in \mathbb{R}^d$, $\boldsymbol{\Lambda} = \mathrm{diag}\left([l_1^2, \dots, l_d^2]^\intercal\right)$, $\nu > 0$ the smoothness parameter, $\Gamma$ the gamma function, and $K_\nu$ the modified Bessel function of the second kind. Common choices for the smoothness parameter are $\nu=3/2$ and $\nu=5/2$.

Once the GP prior is defined, we further assume that the observed output points ${\bf y}$ arise from an observation model $y({\bf x}^i)  = f({\bf x}^i) + \varepsilon^i$, where $\varepsilon^i \sim \mathcal{N}(0,\sigma_\text{n}^2)$ is the observation noise with $\cov(\varepsilon^i,\varepsilon^j) = \sigma_\text{n}^2 \delta_{ij}$. Here, $\sigma_\text{n}$ and $\delta_{ij}$ represent the standard deviation of $\varepsilon^i$ and the Kronecker delta, respectively.
Combining this observation model with the GP prior leads to the following distribution of the observed output points ${\bf y}$, which is known as the marginal likelihood: 
\begin{equation} \label{eqn:marginallikelihood}
    p({\bf y}|{\bf X},\boldsymbol{\phi}_\kappa) = \mathcal{N}({\bf 0},{\bf K} + \sigma_\text{n}^2 {\bf I}) ,
\end{equation}
where the mean vector is ${\bf 0}$ because $m({\bf x}) = 0$, and the $(i,j)$th coordinate of ${\bf K}$ reads ${\bf K}_{ij}  = \kappa({\bf x}^i,{\bf x}^j|\boldsymbol{\phi}_\kappa)$, $i,j \in \{1\dots,N\}$.

At this point, an important question is how to select an observation model that well explains ${\bf y}$. The answer is straightforward if we agree to rank the candidate models using the model posterior.
It follows from Bayes' rule that we can select the maximum a posteriori (MAP) model that corresponds to the MAP of hyperparameters defined as
\begin{equation}
    \boldsymbol{\phi}^\star_\kappa = \underset{\boldsymbol{\phi}_\kappa}{\argmax} \, p(\boldsymbol{\phi}_\kappa) p({\bf y}|{\bf X},\boldsymbol{\phi}_\kappa),
\end{equation}
where $p(\boldsymbol{\phi}_\kappa)$ represents the prior of $\boldsymbol{\phi}_\kappa$.
If $p(\boldsymbol{\phi}_\kappa)$ is flat, then $\boldsymbol{\phi}^\star_\kappa$ is the maximum likelihood estimate (MLE) of $\boldsymbol{\phi}_\kappa$.
This inference approach has been used in many GP toolboxes such as DACE~\citep{Lophaven2002}, GPML~\citep{Rasmussen2010}, GPstuff~\citep{Vanhatalo2013}, pyGPs~\citep{Neumann2015}, GPflow~\citep{Matthews2017jmlr}, GPyTorch \citep{Gardner2018}, and GP+ \citep{Yousefpour2024}.

With the GP prior and the observation model established, we now return to the central question: how do we make predictions using a GP?
The answer lies in the conditioning of multivariate Gaussian distributions discussed in \cref{sec22}.
Specifically, we first, based on the GP prior, form the joint distribution between the function value at the query point ${\bf x}_\xi$ and the observed output points ${\bf y}$, which results in the following $(N+1)$-variate Gaussian distribution:
\begin{equation}\label{jointGaussianGP}
    \begin{bmatrix}
	f({\bf x}_\xi)\\
	{\bf y}
    \end{bmatrix}
    \sim
    \mathcal{N} \left( 
    \begin{bmatrix} 0 \\ 
    {\bf 0}
    \end{bmatrix},
    \begin{bmatrix}
    \kappa({\bf x}_\xi,{\bf x}_\xi) & \boldsymbol{\kappa}_\xi^\intercal\\
    \boldsymbol{\kappa}_\xi &  {\bf C}
    \end{bmatrix}\right),
\end{equation}
where $\boldsymbol{\kappa}_\xi = \left[ \kappa({\bf x}_\xi,{\bf x}^1), \dots, \kappa({\bf x}_\xi,{\bf x}^N) \right]^\intercal$ and ${\bf C} = {\bf K} + \sigma_\text{n}^2 {\bf I}$.
We then apply the conditioning formulas in \cref{eqn:conditionalmeancovariance} to the joint Gaussian distribution in \cref{jointGaussianGP} to find the distribution of
$\widehat{f}({\bf x}_\xi) = f({\bf x}_\xi)|{\bf y}$, which contains both the prediction of $c({\bf x}_\xi)$ and the uncertainty in that prediction.
As described in \cref{sec22}, $\widehat{f}({\bf x}_\xi)$ follows a Gaussian distribution, which reads
\begin{equation} \label{eqn:predictiveGP}
    \widehat{f}({\bf x}_\xi) \sim \mathcal{N} \left( \mu_{\widehat{f}}({\bf x}_\xi), \Sigma_{\widehat{f}}({\bf x}_\xi) \right),
\end{equation}
where $\mu_{\widehat{f}}({\bf x}_\xi)$ and $\Sigma_{\widehat{f}}({\bf x}_\xi)$ are derived from \cref{eqn:conditionalmeancovariance}, as
\begin{equation}\label{eqn:postmeancovariance} 
    \begin{aligned}
        \mu_{\widehat{f}}({\bf x}_\xi) & = \boldsymbol{\kappa}_\xi^\intercal {\bf C}^{-1} {\bf y},\\
        \Sigma_{\widehat{f}}({\bf x}_\xi) & = \kappa({\bf x}_\xi,{\bf x}_\xi) - \boldsymbol{\kappa}_\xi^\intercal {\bf C}^{-1} \boldsymbol{\kappa}_\xi.
    \end{aligned}
\end{equation}

The conditional mean $\mu_{\widehat{f}}$ is a weighted sum of the observed output points ${\bf y}$, where the weights depend on the similarity between the query point ${\bf x}_\xi$ and each of the observed input points ${\bf X}$, which is encapsulated in $\boldsymbol{\kappa}_\xi$.
Thus, $\mu_{\widehat{f}}$ in the interpolation regions between the observations is determined by the data. Away from these regions, it is devolved to the prior mean $m({\bf x})$, which is $0$ in the above expression of $\mu_{\widehat{f}}({\bf x}_\xi)$.
Meanwhile, the conditional covariance $\Sigma_{\widehat{f}}$ is smaller than the prior covariance $\kappa({\bf x}_\xi,{\bf x}_\xi)$.
This indicates that our knowledge of the function has improved after we observe the data.

Having the GP prior and the observation model in \cref{eqn:marginallikelihood}, we can generalize that the function values evaluated at any finite collection of input points follow a Gaussian distribution.
Thus, the posterior distribution of the functions over the entire input space is a GP.
In other words, the GP posterior is itself a GP, which reads 
\begin{equation} \label{eqn:GPposterior}
    \widehat{f}({\bf x}) \sim \mathcal{GP} \left( \widehat{m}({\bf x}), \widehat{\kappa} ({\bf x},{\bf x}') \right),
\end{equation}
where $\widehat{m}({\bf x}) = \boldsymbol{\kappa}_{\bf x}^\intercal {\bf C}^{-1} {\bf y}$ and $ \widehat{\kappa} ({\bf x},{\bf x}') = \kappa({\bf x},{\bf x}') - \boldsymbol{\kappa}_{\bf x}^\intercal {\bf C}^{-1} \boldsymbol{\kappa}_{{\bf x}'}$ are the posterior mean and posterior covariance functions, respectively, and $\boldsymbol{\kappa}_{\bf x} = \left[ \kappa({\bf x},{\bf x}^1), \dots, \kappa({\bf x},{\bf x}^N) \right]^\intercal$.
Computing ${\bf C}^{-1}$ generally costs $\mathcal{O}(N^3)$ operations,
which becomes slow with over thousands of data points.
To handle large datasets, one may use efficient computation strategies such as
inducing points \citep{Hensman2013}, Vecchia approximation \citep{Katzfuss2022}, and approximate inference \citep{Gardner2018}.

\subsection{Weight-space view}
\label{sec32}
The weight-space view interprets GPs as Bayesian generalized linear models (GLMs), which read
\begin{equation} \label{eqn:BayesGLM}
    f({\bf x}) = \sum_{k=1}^{\infty} w_k \phi_k({\bf x}),
\end{equation}
where $w_k$ are the weights and $\phi_k({\bf x})$ the corresponding features.
The weights are treated as Gaussian random variables whose prior and posterior determine the GP prior and GP posterior, respectively.
The features $\phi_k({\bf x})$ are constructed based on the characteristics of the covariance functions and the accuracy and scalability of the approximation.
\Cref{sec4} provides different methods to construct $\phi_k({\bf x})$.

In practice, the weight-space view models GPs using a finite number of features. Accordingly,
the Bayesian GLM in \cref{eqn:BayesGLM} is approximated by
\begin{equation} \label{eqn:TruncatedBayesGLM}
    f({\bf x}) \approx \sum_{k=1}^{N_\phi} w_k \phi_k({\bf x}) = {\bf w}^\intercal \boldsymbol{\phi}({\bf x}),
\end{equation}
where $N_\phi$ is the number of features, ${\bf w} \in \mathbb{R}^{N_\phi}$, and $\boldsymbol{\phi}({\bf x})$: $\mathbb{R}^d \mapsto \mathbb{R}^{N_\phi}$.

Suppose $\boldsymbol{\phi}({\bf x})$ is known.
We start the inference of the Bayesian GLM with a prior of the weights, for example, $p({\bf w}) = \mathcal{N}({\bf 0}, {\bf I})$. 
The equivalence between the prior models $f({\bf x}) = {\bf w}^\intercal \boldsymbol{\phi}({\bf x})$ and $f({\bf x}) \sim \mathcal{GP} \left( 0, \kappa ({\bf x},{\bf x}') \right)$ leads to
\begin{equation}\label{kerneltrick}
\begin{aligned}
    \cov(f({\bf x}),f({\bf x}')) & = \cov({\bf w}^\intercal \boldsymbol{\phi}({\bf x}),{\bf w}^\intercal \boldsymbol{\phi}({\bf x}')) = \boldsymbol{\phi}^\intercal({\bf x}) \cov({\bf w},{\bf w}) \boldsymbol{\phi}({\bf x}') \\
    &  = \boldsymbol{\phi}^\intercal({\bf x})  \boldsymbol{\phi}({\bf x}') = \kappa({\bf x},{\bf x}'),
\end{aligned}
\end{equation}
which is known as the kernel trick \citep[Chapter 6]{Bishop2006} that defines the covariance function as an inner product in the feature space.

We now define an observation model and combine it with the prior $p({\bf w})$ to infer a posterior for the weights, which in turn determines the posterior of the Bayesian GLM.
Like what is assumed in the function-space view, we use the observation model $y({\bf x}^i)  = f({\bf x}^i) + \varepsilon^i$, where $\varepsilon^i \sim \mathcal{N}(0,\sigma_\text{n}^2)$ and $\cov(\varepsilon^i,\varepsilon^j) = \sigma_\text{n}^2 \delta_{ij}$.
By incorporating $f({\bf x})$ in \cref{eqn:TruncatedBayesGLM} into this observation model, we have the following marginal likelihood:
\begin{equation}
    p({\bf y}|{\bf w},{\bf X}) = \mathcal{N}(\boldsymbol{\Phi} {\bf w}, \sigma_\text{n}^2{\bf I}), 
\end{equation}
where $\boldsymbol{\Phi} = [\boldsymbol{\phi}({\bf x}),\dots,\boldsymbol{\phi}({\bf x}^N)]^\intercal \in \mathbb{R}^{N \times N_\phi}$. 

With $p({\bf y}|{\bf w},{\bf X})$ and the prior $p({\bf w})$, the posterior of ${\bf w}$, denoted as $p(\widehat{{\bf w}})$, can be obtained via Bayes’ rule, as $p(\widehat{{\bf w}}) \propto  p({\bf w}) p({\bf y}|{\bf w},{\bf X})$, where $\propto$ is proportional-to symbol.
Since $p({\bf w})$ and $p({\bf y}|{\bf w},{\bf X})$ are Gaussian distributions, $p(\widehat{{\bf w}}) = \mathcal{N}\left( \boldsymbol{\mu}_{\widehat{{\bf w}}}, \boldsymbol{\Sigma}_{\widehat{{\bf w}}} \right)$, where 
\begin{equation} \label{eqn:postweightmeancov}
	\begin{aligned}
		\boldsymbol{\mu}_{\widehat{\bf w}} & = \left(\boldsymbol{\Phi}^\intercal \boldsymbol{\Phi} +\sigma_\text{n}^2 {\bf I}\right)^{-1} \boldsymbol{\Phi}^\intercal {\bf y},\\
		\boldsymbol{\Sigma}_{\widehat{\bf w}} & = \left(\boldsymbol{\Phi}^\intercal \boldsymbol{\Phi} +\sigma_\text{n}^2 {\bf I}\right)^{-1} \sigma_\text{n}^2. 
	\end{aligned}
\end{equation}
As $\boldsymbol{\Phi}^\intercal \boldsymbol{\Phi} +\sigma_\text{n}^2 {\bf I} \in \mathbb{R}^{N_\phi \times N_\phi}$, computing $\left(\boldsymbol{\Phi}^\intercal \boldsymbol{\Phi} +\sigma_\text{n}^2 {\bf I}\right)^{-1}$ generally costs $\mathcal{O}\left( N^3_\phi \right)$ operations. To solve it at $\mathcal{O}\left( \min \{N^3_\phi, N^3 \} \right)$ operations, we can use the Sherman-Morrison-Woodbury (SMW) formula $\left(\boldsymbol{\Phi}^\intercal \boldsymbol{\Phi} +\sigma_\text{n}^2 {\bf I}\right)^{-1} = \sigma_\text{n}^{-2} \left( {\bf I} - \boldsymbol{\Phi}^\intercal \left( \sigma_\text{n}^{2} {\bf I}_N + \boldsymbol{\Phi} \boldsymbol{\Phi}^\intercal \right)^{-1} \boldsymbol{\Phi} \right)$ \citep{Hager1989}, where ${\bf I}_N$ is the $N$-by-$N$ identity matrix. This formula is advantageous when $N_\phi \gg N$.

\section{Sampling from Gaussian processes}
\label{sec4}

Let $\widetilde{f}({\bf x})$ represent a sample path drawn from a GP posterior,
which is called a \textit{posterior sample} or a \textit{posterior sample function} hereafter.
This section describes various methods to generate $\widetilde{f}({\bf x})$.
The prediction performance and computational cost of notable sampling methods are also compared.

\subsection{Exhaustive sampling}
\label{sec4.1}

The exhaustive sampling method (see e.g., \cite{Garnett2023}, Section 8.7), from the function-space view, considers the posterior sample as a set of values ${\bf f}(\boldsymbol{\xi})$ evaluated at $N_\xi$ discrete input variable points $\boldsymbol{\xi}$.
As the nature of GP posteriors, we have ${\bf f}(\boldsymbol{\xi}) \sim \mathcal{N} \left( \boldsymbol{\mu}(\boldsymbol{\xi}), \boldsymbol{\Sigma}(\boldsymbol{\xi})\right)$, where $\boldsymbol{\mu}(\boldsymbol{\xi}) \in \mathbb{R}^{N_\xi}$ is the posterior mean and $\boldsymbol{\Sigma}(\boldsymbol{\xi}) \in \mathbb{R}^{N_\xi \times N_\xi}$ the posterior covariance associated with $\boldsymbol{\xi}$, see \cref{eqn:GPposterior}. 
Thus, samples of ${\bf f}(\boldsymbol{\xi})$ can be generated by
\begin{equation}
    \widetilde{{\bf f}}(\boldsymbol{\xi}) = \boldsymbol{\mu}(\boldsymbol{\xi}) + {\bf L}(\boldsymbol{\xi}) {\bf e},
\end{equation}
where ${\bf L}(\boldsymbol{\xi}){\bf L}^\intercal(\boldsymbol{\xi}) = \boldsymbol{\Sigma}(\boldsymbol{\xi})$ and ${\bf e} \sim \mathcal{N}\left( {\bf 0}, {\bf I}\right) \in \mathbb{R}^{N_\xi}$.

The exhaustive sampling method costs $\mathcal{O}(N_\xi ^3)$ operations to compute the Cholesky decomposition of $\boldsymbol{\Sigma}(\boldsymbol{\xi})$.
This computational cost restricts the method to input domains that can be accurately discretized using a small number of points.
Consequently, the method is expensive for GSA and continuous optimization because these tasks generally require a large number of discrete points to capture the input variable domain.

\subsection{Random Fourier feature}
\label{sec4.2}

The random Fourier feature (RFF) method \citep{Rahimi2007} adopts the weight-space view to approximate a posterior sample using the truncated Bayesian GLM in \cref{eqn:TruncatedBayesGLM}, where the features are called RFFs. Accordingly, we have
\begin{equation}
    \widetilde{f}({\bf x}) = {\bf w}^\intercal \boldsymbol{\phi}({\bf x}),
\end{equation}
where ${\bf w} \sim p(\widehat{{\bf w}}) = \mathcal{N}(\boldsymbol{\mu}_{\widehat{{\bf w}}}, \boldsymbol{\Sigma}_{\widehat{{\bf w}}}) \in \mathbb{R}^{N_\phi}$, as given in \cref{eqn:postweightmeancov}, and $\boldsymbol{\phi}({\bf x}): \mathbb{R}^d \mapsto \mathbb{R}^{N_\phi}$ is the map of RFFs, as described below.

The RFFs $\boldsymbol{\phi}({\bf x})$ are constructed based on the Monte Carlo approximation of a continuous, stationary (shift- or translation-invariant) covariance function on $\mathbb{R}^d$ that, according to Bochner’s theorem (see e.g., \cite{Bochner1933};\cite{Rudin1990}, Chapter 1; \cite{Rahimi2007}), is the Fourier transform of a finite nonnegative symmetric measure on $\mathbb{R}^d$.
Let $\kappa({\bf x},{\bf x}')  = \kappa(\boldsymbol{\delta})$ with $\boldsymbol{\delta} = {\bf x}-{\bf x}'$ represent the stationary covariance function, $\boldsymbol{\omega} \in \mathbb{R}^d$ be a vector of frequencies, and $S(\boldsymbol{\omega})$ denote a way to represent the measure on $\mathbb{R}^d$.

\begin{theorem} (Bochner’s theorem)
A continuous, stationary covariance function $\kappa({\bf x},{\bf x}')  = \kappa(\boldsymbol{\delta})$ defined over a compact set $\mathcal{X} \subset \mathbb{R}^d$ is positive semi-definite if and only if there exists the Fourier duality of $\kappa(\boldsymbol{\delta})$ and $S(\boldsymbol{\omega})$. That is,
\begin{equation} \label{eqn:Bochner}
    \begin{aligned}
        \kappa(\boldsymbol{\delta}) & = \frac{1}{(2\pi)^d} \int_\mathcal{X} S(\boldsymbol{\omega}) e^{i \boldsymbol{\omega}^\intercal \boldsymbol{\delta}} \mathrm{d}\boldsymbol{\omega},\\
        S(\boldsymbol{\omega}) & = \int_\mathcal{X} \kappa(\boldsymbol{\delta}) e^{-i \boldsymbol{\omega}^\intercal \boldsymbol{\delta}} \mathrm{d}\boldsymbol{\delta},
    \end{aligned}
\end{equation}
where $i$ represents the imaginary unit.
\end{theorem} \medskip
These expressions show how to determine $S(\boldsymbol{\omega})$ from $\kappa(\boldsymbol{\delta})$ or vice versa.
If $S(\boldsymbol{\omega})$ is a density associated with a probability measure on $\mathbb{R}^d$, known as the spectral density of $\kappa(\boldsymbol{\delta})$, then it can be determined from the second expression of \cref{eqn:Bochner} for a specific stationary covariance $\kappa(\boldsymbol{\delta})$.
Since $\kappa({\bf 0}) = \sigma_\text{f}^2$, $S(\boldsymbol{\omega})$ can be written as $S(\boldsymbol{\omega}) = \sigma_\text{f}^2 (2 \pi)^d p(\boldsymbol{\omega})$, where $p(\boldsymbol{\omega})$ is the probability density function (PDF) associated with $S(\boldsymbol{\omega})$.
In this way, the spectral PDF for the SE covariance function in \cref{SEcovariancefunction} can be derived as 
\begin{equation} \label{eqn:SEspectral}
    p(\boldsymbol{\omega}) = \left(\sqrt{2\pi}\right)^{-d} \left(\prod_{i=1}^{d}l_i\right) \exp{\left(-\frac{1}{2} \sum_{i=1}^{d}l_i^2 \omega_i^2\right)} = \mathcal{N} \left( {\bf 0}, \boldsymbol{\Sigma}\right),
\end{equation}
where $\boldsymbol{\Sigma} = \mathrm{diag}\left([l_1^{-2}, \dots, l_d^{-2}]^\intercal\right)$.
The spectral PDF for the class of Matérn covariance functions in \cref{Materncovariancefunction} is provided in Appendix~\ref{AppB}.

Given $S(\boldsymbol{\omega})$ or $p(\boldsymbol{\omega})$, $\kappa(\boldsymbol{\delta})$ can be determined from the first expression of \cref{eqn:Bochner} as
\begin{equation} \label{eqn:realCovariance}
    \begin{aligned}
        \kappa(\boldsymbol{\delta}) & = \sigma_\text{f}^2  \mathrm{Re}\left[  \int_\mathcal{X} p(\boldsymbol{\omega}) e^{i \boldsymbol{\omega}^\intercal \boldsymbol{\delta}} \mathrm{d}\boldsymbol{\omega} \right]
         = \sigma_\text{f}^2 \int_\mathcal{X} p(\boldsymbol{\omega}) \cos \left( \boldsymbol{\omega}^\intercal \boldsymbol{\delta} \right)  \mathrm{d}\boldsymbol{\omega}\\
        & = \sigma_\text{f}^2 \int_\mathcal{X} p(\boldsymbol{\omega}) \cos \left( \boldsymbol{\omega}^\intercal ({\bf x}-{\bf x}') \right)  \mathrm{d}\boldsymbol{\omega}.
    \end{aligned}
\end{equation}
Here, $\mathrm{Re}[\cdot]$ denotes the
real part.
Since $\int_{0}^{2\pi} \cos\left( a + 2b\right) \mathrm{d}b = 0, \forall a \in \mathbb{R}$, we can rewrite \cref{eqn:realCovariance} as
\begin{equation} \label{eqn:rcovariance}
    \kappa(\boldsymbol{\delta})  = \sigma_\text{f}^2 \int_\mathcal{X} p(\boldsymbol{\omega}) \cos \left( \boldsymbol{\omega}^\intercal{\bf x}- \omega^\intercal{\bf x}' \right)  \mathrm{d}\boldsymbol{\omega} 
     + \sigma_\text{f}^2 \int_{0}^{2\pi} \cos\left( \boldsymbol{\omega}^\intercal {\bf x} +\boldsymbol{\omega}^\intercal {\bf x}' + 2b\right) \mathrm{d}b.
\end{equation}

From the product-sum trigonometric identities, we also have
\begin{equation} \label{eqn:trigidentity} 
      \cos\left( \boldsymbol{\omega}^\intercal {\bf x}  -\boldsymbol{\omega}^\intercal {\bf x}' \right) + \cos\left( \boldsymbol{\omega}^\intercal {\bf x} +\boldsymbol{\omega}^\intercal {\bf x}' + 2b \right)
     =  2 \cos\left( \boldsymbol{\omega}^\intercal {\bf x} + b \right) \cos\left( \boldsymbol{\omega}^\intercal {\bf x}' + b \right).
\end{equation}

By applying the Monte Carlo integration to \cref{eqn:rcovariance} and incorporating \cref{eqn:trigidentity} in the resulting approximation, we obtain the following approximation of $\kappa(\boldsymbol{\delta})$ \citep{Rahimi2007}:
\begin{equation} \label{eqn:RFFcovariance}
     \kappa(\boldsymbol{\delta}) = \kappa({\bf x},{\bf x}') \approx \frac{2 \sigma_\text{f}^2}{N_\phi} \sum_{k=1}^{N_\phi} \boldsymbol{\zeta}^\intercal({\bf x}) \boldsymbol{\zeta}({\bf x}'),
\end{equation}
where $\boldsymbol{\zeta}({\bf x}) = \left[ \cos(\boldsymbol{\omega}_1^\intercal {\bf x}+ b_1),\dots, \cos(\boldsymbol{\omega}_{N_\phi}^\intercal {\bf x} + b_{N_\phi}) \right]^\intercal$. Here, $\boldsymbol{\omega}_k \sim p(\boldsymbol{\omega})$ and $b_k \sim \mathcal{U}(0,2 \pi)$ can be viewed as samples drawn from the joint distribution of $\boldsymbol{\omega}$ and $b$, with $\mathcal{U}$ representing the uniform distribution and $N_\phi$ the number of Monte Carlo samples.

From \cref{kerneltrick,eqn:RFFcovariance}, we define the following features as RFFs:
\begin{equation} \label{eqn:RFF}
    \boldsymbol{\phi}({\bf x}) \approx \frac{\sqrt{2}\sigma_\text{f}}{\sqrt{N_\phi}}\left[ \cos(\boldsymbol{\omega}_1^\intercal {\bf x}+ b_1,)\dots, \cos(\boldsymbol{\omega}_{N_\phi}^\intercal {\bf x}+ b_{N_\phi})\right]^\intercal.
\end{equation}

\begin{figure}[t]
	\centering
	\includegraphics[width=\textwidth]{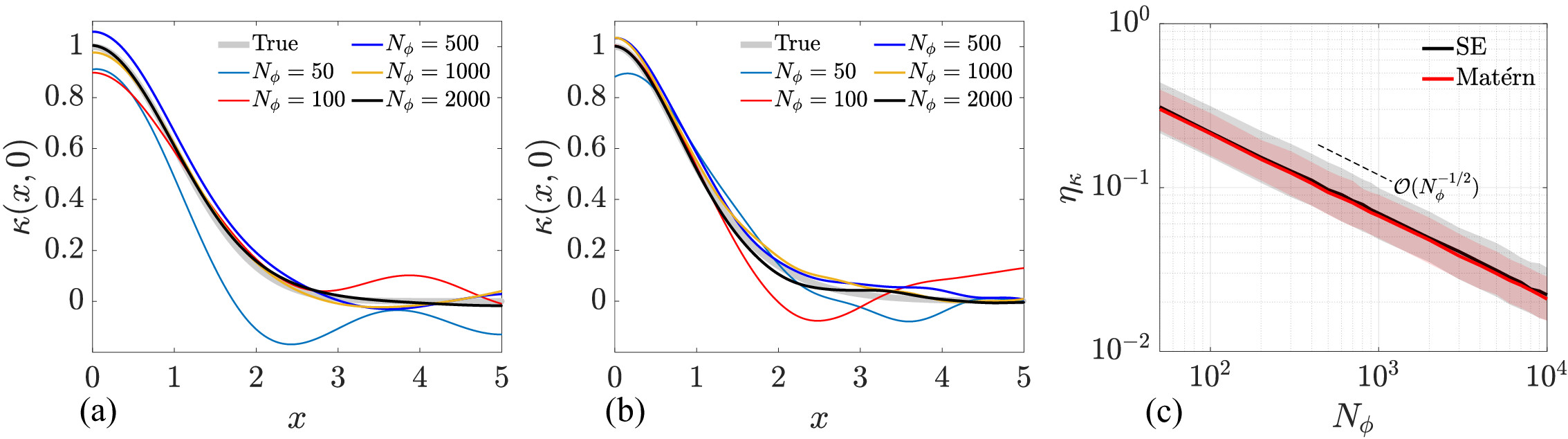}
	\caption{ Approximations of SE and Matérn 5/2 covariance functions using different numbers of random features. (a) SE. (b) Matérn 5/2. (c) Error means (solid lines) and 90\% error intervals (shaded areas).}
    \label{fig:approximateSE}
\end{figure}

\Cref{fig:approximateSE} shows the approximations of a univariate SE covariance function and a Matérn 5/2 covariance function for different numbers of features, and the corresponding convergence rates.
In Figs.~\ref{fig:approximateSE}(a) and (b), we set $l=1$, $\sigma_\text{f} =1$, and $x'=0$. In \cref{fig:approximateSE}(c), we plot the mean and 90\% interval of the relative approximation error, defined as $ \eta_\kappa  = \left\|\boldsymbol{\kappa}({\bf X},{\bf X}')-\widetilde{\boldsymbol{\kappa}}({\bf X},{\bf X}') \right\|_2/\left\| \boldsymbol{\kappa}({\bf X},{\bf X}') \right\|_2$, for different the numbers $N_\phi$ of features, where $\widetilde{\boldsymbol{\kappa}}$ is the approximation of vectorized $\boldsymbol{\kappa}$, $\left\| \cdot \right\|_2$ denotes the $L^2$-norm of a vector, ${\bf X}$ and ${\bf X}'$ are sets of 2000 points evenly spaced on $[-5, 5]$, and $N_\phi \in \{50, 100, 200, \dots, 900, 1000, 2000, \dots, 9000, 10000\}$. Each approximation trial is repeated 1000 times to estimate the mean and 90\% interval of the approximation error. We see that the approximations by RFFs converge at the Monte Carlo rate of $\mathcal{O}\left( N_\phi^{-1/2} \right)$, as proven by \cite{Sutherland2015}.

\Cref{{alg:rffs}}, i.e., $\texttt{GP-RFF}$, summarizes the implementation of the RFF method to draw GP posterior samples.
Given $\boldsymbol{\phi}({\bf x})$ in \cref{eqn:RFF}, $\texttt{GP-RFF}$ generally costs $\mathcal{O}\left( N^3_\phi\right)$ operations to generate ${\bf w}$ from its posterior and $\mathcal{O}\left( N_\phi N_\xi d \right)$ operations for matrix multiplication in Line 12, where $N_\xi$ is the number of discrete points used to capture the input variable domain.
When $ N_\phi \gg N$, we may generate ${\bf w}$ from its posterior at $\mathcal{O}\left( N^3 \right)$ operations using the SMW formula and the Cholesky decomposition of an $N$-by-$N$ matrix \citep{Seeger2007}, which is not implemented in $\texttt{GP-RFF}$.  

\begin{algorithm}[t]
	\caption{\texttt{GP-RFF}: GP posterior samples via random Fourier features}
	\label{alg:rffs}
	\begin{algorithmic}[1]
		\State \textbf{Input:} dataset $\mathcal{D}$, stationary covariance function $\kappa(\cdot,\cdot)$, number of features $N_\phi$, standard deviation of observation noise $\sigma_\text{n}$
		\State 
        Optimize hyperparameters $\boldsymbol{\phi}_\kappa =\left[ \sigma_\text{f},l_1,\dots,l_d\right]^\intercal$ from $\mathcal{D}$, $\kappa(\cdot,\cdot)$, and $\sigma_\text{n}$ \label{alg:rffs-line2}
		\State Formulate $p(\boldsymbol{\omega})$, see \cref{eqn:SEspectral} or \cref{MaternspectralPDF} \Comment{spectral PDF}
		
		\For {$k=1:N_\phi$} 
		\State $\boldsymbol{\omega}_k \sim p(\boldsymbol{\omega})$ \Comment{points in frequency domain}
		\State $b_k \sim \mathcal{U}(0,2\pi)$ \Comment{points in phase domain}
		\EndFor
		
		\State Formulate $\boldsymbol{\phi}({\bf x})$, see \cref{eqn:RFF} \Comment{random Fourier features}
		\State $\boldsymbol{\Phi} \gets [\boldsymbol{\phi}^1({\bf x}),\dots,\boldsymbol{\phi}({\bf x}^N)]^\intercal$
		\State Compute $\boldsymbol{\mu}_{\widehat{{\bf w}}}$ and $\boldsymbol{\Sigma}_{\widehat{{\bf w}}}$, see \cref{eqn:postweightmeancov} 
		\State ${\bf w} \sim \mathcal{N}(\boldsymbol{\mu}_{\widehat{{\bf w}}}, \boldsymbol{\Sigma}_{\widehat{{\bf w}}})$, see \cref{sec21}
		\State \textbf{return} $\widetilde{f}({\bf x}) \gets {\bf w}^\intercal \boldsymbol{\phi}({\bf x})$ \label{alg:rffs-line12}
	\end{algorithmic}
\end{algorithm}

\begin{figure}[h]
	\centering
	\includegraphics[width=\textwidth]{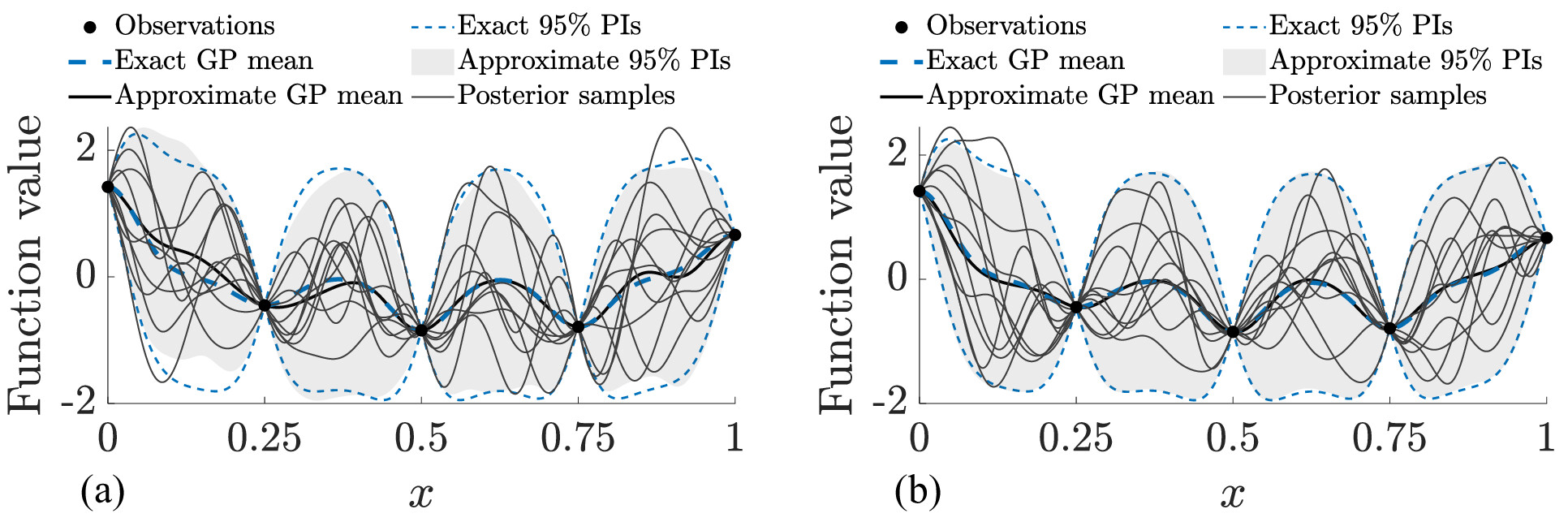}
	\caption{Mean and predictive intervals of the exact posterior compared with those of approximations by RFF method with (a) $N_\phi=100$ and (b) $N_\phi=2000$ random features.}
    \label{fig:rffsampling}
\end{figure}

To show how the number of RFFs influences the prediction performance of $\texttt{GP-RFF}$, we set $N_\phi=100$ and $N_\phi=2000$ and use $\texttt{GP-RFF}$ to compute the approximate GP mean and approximate 95\% prediction intervals (PIs) of a univariate GP. As shown in \cref{fig:rffsampling}, the approximation accuracy significantly improves as the number of RFFs increases from $N_\phi=100$ to $N_\phi=2000$.

\begin{figure}[t]
	\centering
	\includegraphics[width=\textwidth]{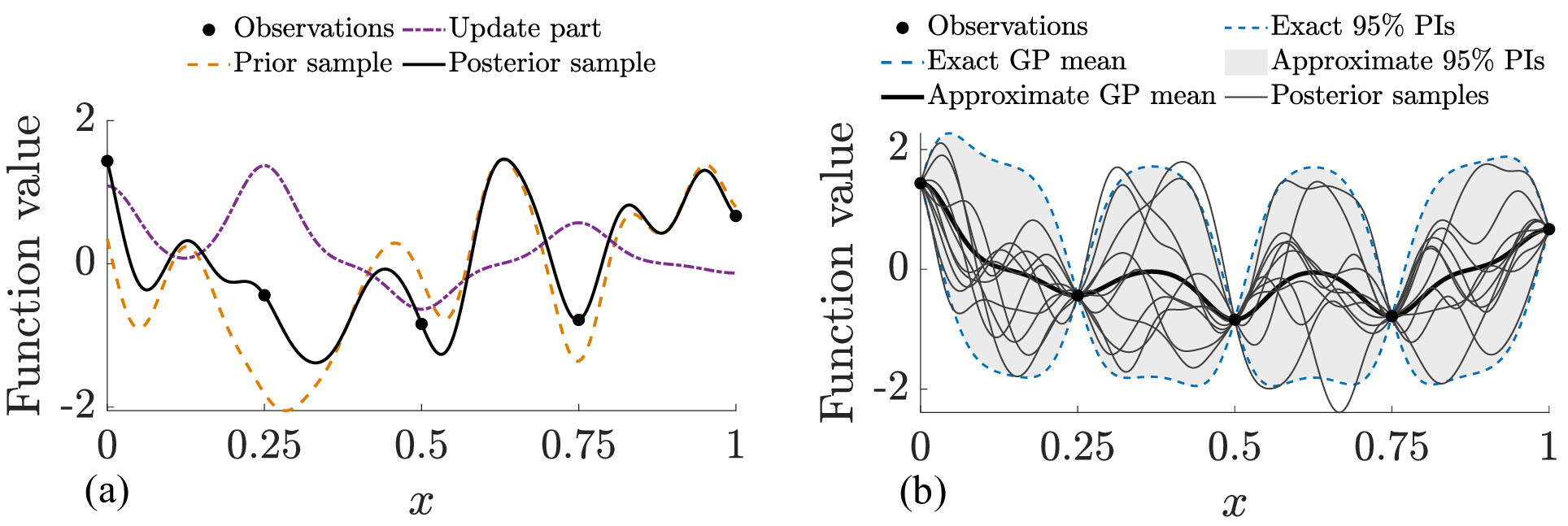}
	\caption{Illustration of the PC method. (a) A posterior sample is formed by adding an update part to a prior sample. (b) Mean and predictive intervals of the exact posterior compared with those of approximations by the PC method with RFF prior of $N_\phi=2000$.}
    \label{fig:decoupledsampling}
\end{figure}

\subsection{Pathwise conditioning}
\label{sec4.3}

The pathwise conditioning (PC) method \citep{Wilson2020,Wilson2021} draws a GP posterior sample by updating the corresponding \textit{prior sample} using the distribution equality of Matheron’s rule given in \cref{eqn:distributionequality}.
When the observations ${\bf y}$ are corrupted by independent and identically distributed Gaussian noise, the posterior sample can be expressed as
\begin{equation} \label{eqn:pathwise}
    \underbrace{\widetilde{f}({\bf x})}_\mathrm{posterior} = \underbrace{f({\bf x})}_\mathrm{prior} + \underbrace{\boldsymbol{\kappa}_{\bf x}^\intercal {\bf C}^{-1} ({\bf y}-{\bf f} - \boldsymbol{\varepsilon})}_\mathrm{update},
\end{equation}
where $f({\bf x})$ is the prior sample associated with $\widetilde{f}({\bf x})$, ${\bf f} = [f({\bf x}^1),\dots,f({\bf x}^N)]^\intercal$ the prior sample values at the observed input points, and $\boldsymbol{\varepsilon} = [\varepsilon^1,\dots,\varepsilon^N]^\intercal$ with $\varepsilon^i \sim \mathcal{N}(0,\sigma^2_\text{n})$.
In an analogy to \cref{eqn:distributionequality}, we can interpret $f({\bf x})$ and ${\bf f} + \boldsymbol{\varepsilon}$ as the marginal components of a sample drawn from their joint distribution.
We call the second term on the right-hand side of \cref{eqn:pathwise} the update (or data adjustment) part, which is a weighted-sum of canonical basis functions $\boldsymbol{\kappa}_{\bf x}$ as in the first expression of \cref{eqn:postmeancovariance} from the function-space perspective.

How can we generate the prior sample $f({\bf x})$? From the weight-space perspective, the answer is straightforward as $f({\bf x}) = {\bf w}^\intercal \boldsymbol{\phi}({\bf x})$, where ${\bf w} \sim p({\bf w})$.
Sampling $f({\bf x})$ thus only requires the samples of ${\bf w}$ drawn from its prior $p({\bf w})$ and the feature map $\boldsymbol{\phi}({\bf x})$. 
In the seminal paper, \cite{Wilson2020} generated $f({\bf x})$ using $p({\bf w})= \mathcal{N}({\bf 0}, {\bf I})$ and $\boldsymbol{\phi}({\bf x})$ as RFFs of stationary covariance functions.
Alternative feature maps $\boldsymbol{\phi}({\bf x})$, such as those provided in \cref{otherapproaches}, can also be used.

\Cref{fig:decoupledsampling} illustrates the concept of the PC method.
\Cref{{alg:pc}}, i.e., $\texttt{GP-PC}$, summarizes its implementation, where $f({\bf x})$ in Line 13 is drawn from RFFs formulated in Line 11.
$\texttt{GP-PC}$ costs $\mathcal{O}(N^3)$ operations to generate a posterior sample function $\widetilde{f}({\bf x})$, and scales linearly in $N_\phi$.

\begin{algorithm}[t]
	\caption{\texttt{GP-PC}: GP posterior samples via pathwise conditioning}
	\label{alg:pc}
	\begin{algorithmic}[1]
		\State \textbf{Input:} dataset $\mathcal{D}$, stationary covariance function $\kappa(\cdot,\cdot)$, number of features $N_\phi$, standard deviation of observation noise $\sigma_\text{n}$
		\State Find an optimal set of hyperparameters $\boldsymbol{\phi}_\kappa =\left[ \sigma_\text{f},l_1,\dots,l_d\right]^\intercal$ from $\mathcal{D}$, $\kappa(\cdot,\cdot)$, and $\sigma_\text{n}$ \label{alg:pc-line2}
        \State ${\bf C}^{-1} \gets \left({\bf K} + \sigma^2_\text{n} {\bf I} \right)^{-1}$
        \State $\boldsymbol{\kappa}_{\bf x} \gets \left[\kappa({\bf x},{\bf x}^1),\dots,\kappa({\bf x},{\bf x}^N)\right]^\intercal$
        \State $\boldsymbol{\varepsilon} \sim \mathcal{N}({\bf 0}, \sigma^2_\text{n} {\bf I})$
		\State Formulate $p(\boldsymbol{\omega})$, see \cref{eqn:SEspectral} or \cref{MaternspectralPDF} \Comment{spectral PDF}
		
		\For {$k=1:N_\phi$} 
		\State $\boldsymbol{\omega}_k \sim p(\boldsymbol{\omega})$ \Comment{points in frequency domain}
		\State $b_k \sim \mathcal{U}(0,2\pi)$ \Comment{points in phase domain}
		\EndFor
		
		\State Formulate $\boldsymbol{\phi}({\bf x})$, see \cref{eqn:RFF} \Comment{random Fourier features} \label{alg:pc-line11}
		\State ${\bf w} \sim \mathcal{N}({\bf 0}, {\bf I})$ \Comment{sample ${\bf w}$ from its prior $p({\bf w})$}
        \State $f({\bf x}) \gets {\bf w}^\intercal \boldsymbol{\phi}({\bf x})$ \Comment{prior sample} \label{alg:pc-line13}
        \State ${\bf f} \gets [f({\bf x}^1),\dots,f({\bf x}^N)]^\intercal$ \Comment{prior values at observed input points}
		\State \textbf{return} $\widetilde{f}({\bf x}) \gets f({\bf x}) + \boldsymbol{\kappa}_{\bf x}^\intercal {\bf C}^{-1} ({\bf y}-{\bf f} - \boldsymbol{\varepsilon})$
	\end{algorithmic}
\end{algorithm}

\subsection{Other sampling methods}
\label{otherapproaches}

This section provides a brief overview of three approaches for constructing the feature map
$\boldsymbol{\phi}(\mathbf{x})$ in the weight-space view:
orthonormal expansions, Hilbert space approximations, and quasi-Monte Carlo methods.
These approaches can be used as alternatives to the RFF method,
either to generate GP posterior samples directly
or to generate GP prior samples to be used in the PC method.

\subsubsection{Orthonormal expansion}
Consider a univariate input domain $\mathcal{X}$. Every positive semi-definite covariance function $\kappa(x,x'): \mathcal{X} \times \mathcal{X} \mapsto \mathbb{R}$ induces a reproducing kernel Hilbert space (RKHS), which is a Hilbert space of functions in which evaluating a function at any point is the same as taking an inner product with the covariance function.
If the RKHS induced by a covariance function is separable, then the RKHS has a countable orthonormal basis, and the covariance function can be written as a pointwise convergent
orthonormal expansion (OE) \citep{Tronarp2024}:
\begin{equation}\label{orthonormalexpansion}
    \kappa(x,x') = \sum_{k \in \mathbb{N}} \phi_k(x) \phi_k(x'), \ \ \forall x,x' \in \mathcal{X},
\end{equation}
where $\phi_k(x)$ form the orthonormal basis of the RKHS.
In practice, truncated expansions $\sum_{k = 0}^{N_\phi-1} \phi_i(x) \phi_i(x')$ are used as approximate covariance functions, which are usually not stationary. 

A notable example of OEs is the spectral representation of a covariance function per Mercer’s theorem \citep[Section 4.3]{Rasmussen2006}.
Accordingly, if a continuous, bounded covariance function is positive semi-definite, then it can be expressed as $\kappa (x,x') = \sum_{k=0}^{\infty} \lambda_k \psi_k(x) \psi_k(x')$, where $\lambda_k \geq \lambda_{k+1} \geq 0$ are the eigenvalues of the covariance operator $\mathcal{K}$ defined by $\mathcal{K}f = \int \kappa(\cdot,x')f(x')\mathrm{d}x'$, and $\psi_k(x)$ are the corresponding orthonormal eigenfunctions.
In this case, the orthonormal bases in \cref{orthonormalexpansion} are $\phi_k(x) = \sqrt{\lambda_k} \psi_k(x)$.

\cite{ZhuHY1998} derived analytical eigenfunctions and eigenvalues for univariate SE covariance functions $\kappa(x, x'| l) = \exp(-0.5 s^2)$, where $s = |x - x'| / l$ and length scale $l>0$. Let $\mu = \mathcal{N}(0,\sigma^2)$ be a Gaussian measure on $\mathcal{X}$, $a = (2 \sigma^{2})^{-1}$, $b = (2 l)^{-1}$, $c = \sqrt{a^2 + 4 a b}$, and $A = 0.5 a + b + 0.5 c$.
The $k$th eigenvalue is $\lambda_k = \sqrt{a/A} \left( b/A \right)^k$ and the corresponding eigenfunction is
$\psi_k(x) = \left( {\pi c}/{a} \right)^{1/4} \varphi_k (\sqrt{c} x) \exp\left( 0.5 a x^2 \right)$,
where $\varphi_k (x) = \left( \pi^{1/2} 2^k k! \right)^{-1/2} H_k(x) \exp\left( -0.5 x^2 \right)$ with $H_k(x)$ representing the $k$th-order physicist's Hermite polynomial.
Since the $d$-variate SE covariance function in \cref{SEcovariancefunction} has a tensor product form, the $d$-dimensional eigenfunctions and 
eigenvalues are $\psi_k({\bf x}) = \prod_{i=1}^d \psi_{k,i}(x_i)$ and $\lambda_k = \prod_{i=1}^d \lambda_{k,i}$ \citep{Adebiyi2025tsroots}. This, however, causes the approximation to suffer from the curse of dimensionality.

\cite{Tronarp2024} used Matérn-Laguerre functions as the orthonormal bases for univariate Matérn covariance functions.
A collection of Mercer's expansions for non-stationary covariance functions can be found in \cite{Fasshauer2015}, Appendix A.

\subsubsection{Hilbert space approximation}

We now consider an isotropic covariance function of the form $\kappa({\bf x},{\bf x}')  = \kappa(\left\|\boldsymbol{\delta}\right\|_2)$, where $\boldsymbol{\delta} = {\bf x}-{\bf x}'$.
Its spectral density from \cref{eqn:Bochner} can be written as $S(\boldsymbol{\omega}) = S(\left\|\boldsymbol{\omega} \right\|_2)$.
Since the Fourier transform of the Laplace operator $\nabla^2$ is $-\left\|\boldsymbol{\omega} \right\|_2$, the covariance operator $\mathcal{K}$, defined by $\mathcal{K}f = \int \kappa(\cdot,{\bf x}')f({\bf x}')\mathrm{d}{\bf x}'$, can be represented as a formal power series of the Laplace operator $\nabla^2$, i.e., $\mathcal{K}=\sum_{k=0}^\infty \alpha_k (-\nabla^2)^k$ \citep{Solin2020}.
On a compact bounded domain $\mathcal{X}$ with a suitable boundary condition, $-\nabla^2$ has an orthonormal eigendecomposition $-\nabla^2 f= \int l(\cdot,{\bf x}')f({\bf x}')\mathrm{d}{\bf x}'$ with $l({\bf x},{\bf x}') = \sum_{k \in \mathbb{N}} \lambda_k \psi_k({\bf x}) \psi_k({\bf x}')$.
Plugging into the series of the covariance operator $\mathcal{K}$ leads to the following approximation of $\kappa({\bf x},{\bf x}')$ \citep{Solin2020}:
\begin{equation}\label{Hilbertapproximation}
    \kappa({\bf x},{\bf x}') \approx \sum_{k=1}^\infty S(\sqrt{\lambda_k}) \psi_k({\bf x}) \psi_k({\bf x}'),
\end{equation}
where $S(\cdot)$ is the spectral density of the covariance function, see \cref{eqn:Bochner}.
The right-hand side of \cref{Hilbertapproximation} is an approximation of $\kappa({\bf x},{\bf x}')$  because it is restricted to the domain $\mathcal{X}$ and the boundary conditions.
In this case, the features are approximated by $\phi_k({\bf x}) \approx \left[S(\sqrt{\lambda_k})\right]^{1/2} \psi_k({\bf x})$.
While \cite{Solin2020} considered the eigenvalue problem for the Laplacian with homogeneous Dirichlet boundary conditions,
other boundary conditions, such as Neumann or periodic conditions, can also be used to determine $\lambda_k$ and $\psi_k({\bf x})$ with similar approximation errors.

For a unidimensional input domain $\mathcal{X}=[-L,L]$ with $L>0$, the solution to the eigenvalue problem for the Laplacian with homogeneous Dirichlet boundary conditions
is $\lambda_k = {\pi^2 k^2}/(4L^2)$ and $\psi_k(x) = L^{-1/2}\sin\left( \pi k (x+L)/(2L)\right)$, which are independent of the choice of the covariance function.
For a $d$-dimensional input domain, the number
of eigenfunctions for \cref{Hilbertapproximation} is
equal to the number of possible combinations of univariate eigenfunctions over all dimensions. This makes the method impractical when $d>5$ \citep{RiutortMayol2023}. Detailed implementation of the Hilbert space approximation can be found in \cite{RiutortMayol2023}.

\subsubsection{Quasi-Monte Carlo}
Quasi-Monte Carlo (QMC) features \citep{Avron2016qmc,HuangZ2024qmc} can improve the convergence of RFF approximations for stationary covariance functions by using a low-discrepancy sequence (e.g., a Halton or Sobol' sequence) to estimate the integral in \cref{eqn:realCovariance}.
Specifically, if the spectral PDF of the covariance function is separable, i.e., $p(\boldsymbol{\omega}) = \prod_{i=1}^d p_i(\omega_i)$, which is the case of the spectral PDF of SE covariance functions in \cref{eqn:SEspectral}, then low-discrepancy points $\boldsymbol{\omega}_1,\dots,\boldsymbol{\omega}_{N_\phi}$ in the frequency domain can be determined by transforming the corresponding low-discrepancy points ${\bf t}_1,\dots,{\bf t}_{N_\phi}$ in the unit hypercube $[0,1]^d$ via $\boldsymbol{\omega} = \Phi^{-1}({\bf t})$, where ${\bf t} = [t_1,\dots,t_d]^\intercal \in [0,1]^d$ and 
$\Phi^{-1}({\bf t}) = \left[ \Phi^{-1}_1(t_1),\dots, \Phi^{-1}_d(t_d) \right]^\intercal$ with $\Phi_i(\cdot)$ representing the cumulative distribution function of $p_i(\cdot)$.
If one wishes to generate low-discrepancy points for both frequency $\boldsymbol{\omega}$ and phase $b \sim \mathcal{U}(0,2\pi)$ in \cref{eqn:rcovariance}, then the corresponding low-discrepancy points are generated for $[{\bf t},\widetilde{b}]^\intercal \in [0,1]^{d+1}$ before applying the transformations $\boldsymbol{\omega} = \Phi^{-1}({\bf t})$ and $b=2\pi\widetilde{b}$. 

\begin{figure}[t]
	\centering
	\includegraphics[width=\textwidth]{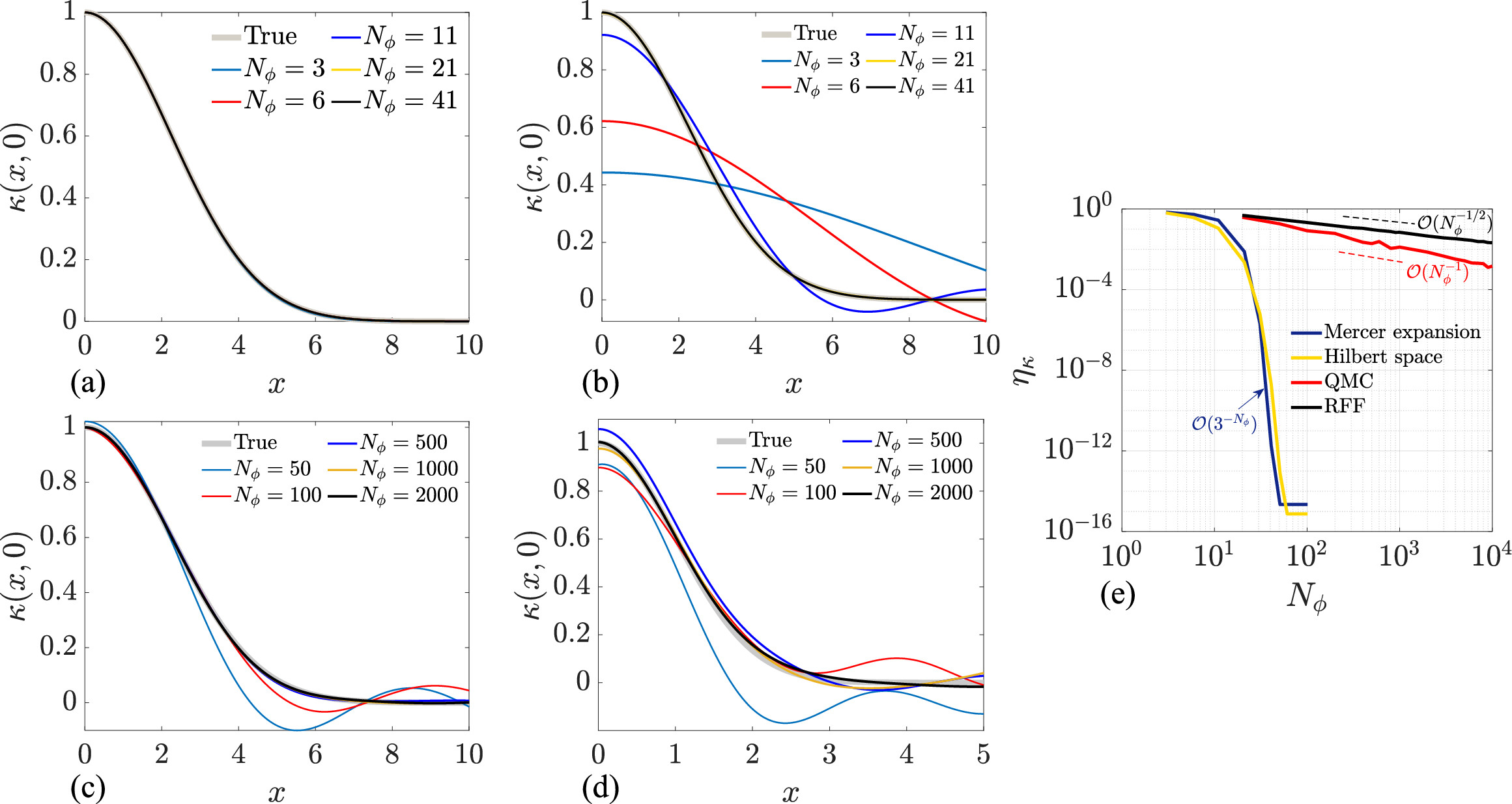}
	\caption{Approximations of the univariate SE covariance function with length scale $l=\sqrt{5}$ and output scale $\sigma_\mathrm{f} = 1$ using (a) Mercer's expansion with $\sigma=\sqrt{3}/2$, (b) Hilbert space approximation, (c) QMC with a Halton sequence, and (d) RFF. (e) Comparison of approximation convergence.}
    \label{fig:approximateSE2}
\end{figure}

\begin{figure}[t]
	\centering
	\includegraphics[width=\textwidth]{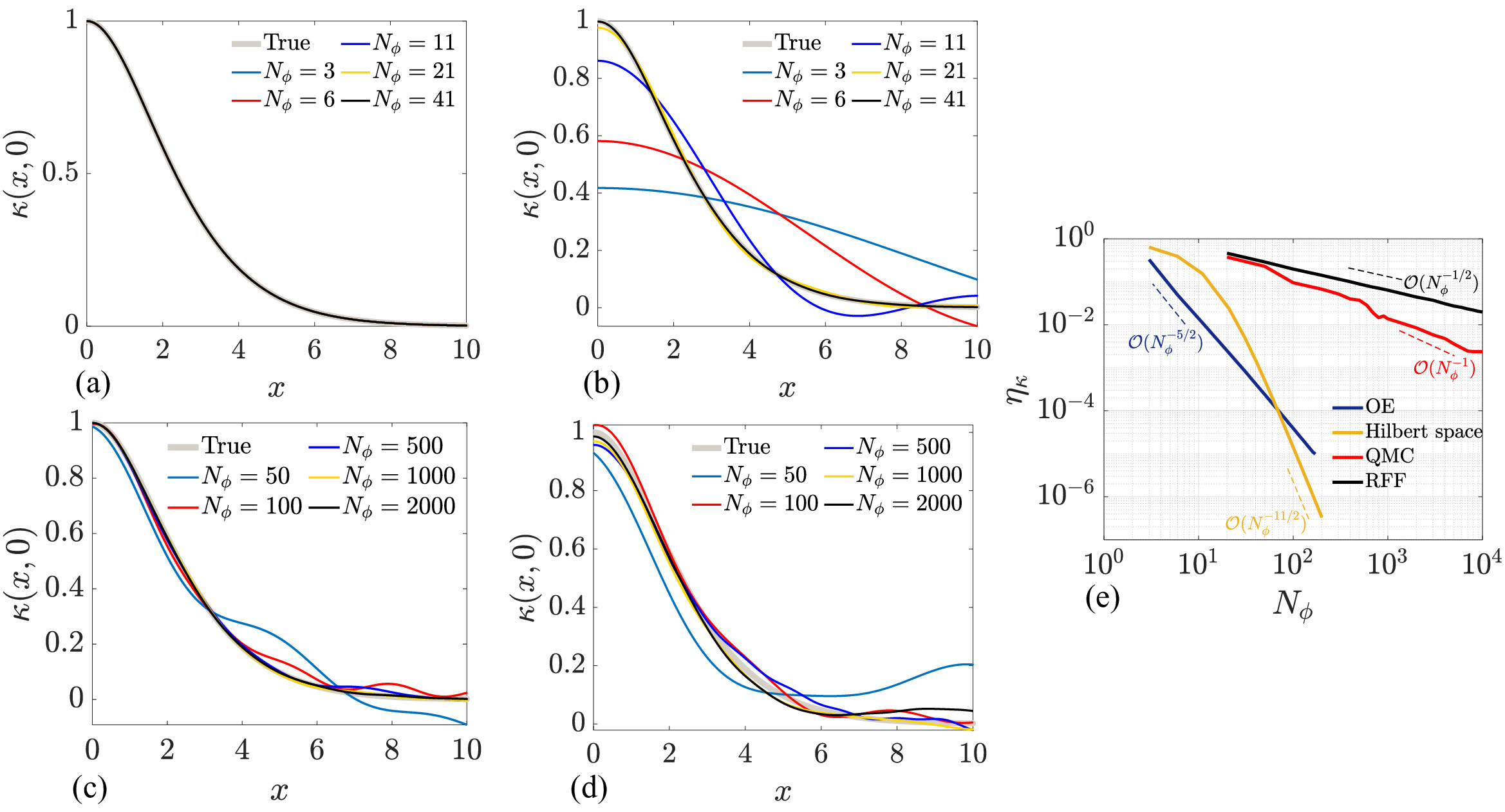}
	\caption{Approximations of the univariate Matérn 5/2 covariance function with length scale $l=\sqrt{5}$ and output scale $\sigma_\mathrm{f}=1$
 using (a) OE, (b) Hilbert space approximation, (c) QMC with a Halton sequence, and (d) RFF. (e) Comparison of approximation convergence.}
    \label{fig:approximateMatern}
\end{figure}

\subsubsection{Comparison of approximation accuracy}
\label{sec444}
We compare the approximation accuracy and convergence of OE, Hibert space approximation, QMC, and RFF methods using a univariate SE covariance function and a univariate Matérn 5/2 covariance function. These covariance functions are defined on $[-10,10]$ with $l=\sqrt{5}$ and $\sigma_\mathrm{f}=1$.
For the OE method, we use a Mercer's expansion with $\sigma = \sqrt{3}/2$ to approximate the SE covariance function, and an almost Mercer's expansion \citep{Tronarp2024} to approximate the Matérn covariance function. 
The homogeneous Dirichlet boundary conditions are used for the Hilbert space approximation.
For the QMC method, a Halton sequence \citep{Halton1960} is generated to determine the low-discrepancy points in the frequency and phase domains.

\Cref{fig:approximateSE2} shows the approximations of the SE covariance function. For the SE covariance function, Mercer's expansion with $\sigma = \sqrt{3}/2$ has an exponential convergence at a rate of $\mathcal{O}(3^{-N_\phi})$, consistent with \cite{Tronarp2024}, Proposition 5.3.
The convergence rate of the Hibert space approximation is on par with that of Mercer's expansion with $\sigma = \sqrt{3}/2$.
The QMC method has a convergence rate of $\mathcal{O}(N_{\phi}^{-1})$ (up to logarithmic factors), outperforming the RFF method that converges at a rate of $\mathcal{O}(N_{\phi}^{-1/2})$.
\Cref{fig:approximateMatern} compares the approximations of the univariate Matérn 5/2 covariance function with $l=\sqrt{5}$ and $\sigma_\mathrm{f}=1$ by the OE (almost Mercer's expansion) \citep{Tronarp2024}, Hilbert space approximation, QMC with a Halton sequence, and RFF.
The convergence rate of the OE method is $\mathcal{O}(N_{\phi}^{-5/2})$, consistent with \cite{Tronarp2024}, Proposition 3.3, while that of the Hibert space approximation reaches $\mathcal{O}(N_{\phi}^{-11/2})$ as the number of basis functions increases.
The QMC and RFF methods require thousands of basis functions to reach a similar level of accuracy.
Interestingly, the OE method provides more accurate approximations than the Hilbert space approximation when using small numbers of basis functions.

\subsection{Comparisons of prediction performance and computational cost}
\label{sec4.4}

We assess the prediction performance of the exhaustive sampling, $\texttt{GP-RFF}$, and $\texttt{GP-PC}$ with RFF prior on the GPs for the $1$D Levy function \citep{Surjanovic2013}, which is defined in $\left[\underline{x},\overline{x}\right] = [-10,10]$. We measure the prediction performance using the $2$-Wasserstein distance between the multivariate Gaussian distribution of the function values at a set of query points evaluated from the exact GP in \cref{eqn:predictiveGP} and that estimated by each sampling method.
The $2$-Wasserstein distance is defined as
\begin{equation}
    d_\mathrm{W} 
     = \left( \left\| {\bf m}_\xi - \widetilde{\bf m}_\xi \right\|_2^2 
     + \mathrm{tr}\left( {\bf K}_{\xi,\xi} +\widetilde{\bf K}_{\xi,\xi} 
     -2 \left( {\bf K}_{\xi,\xi}^{1/2} \widetilde{\bf K}_{\xi,\xi} {\bf K}_{\xi,\xi}^{1/2}\right)^{1/2} \right) \right)^{1/2},
\end{equation}
where ${\bf m}_\xi$ and ${\bf K}_{\xi,\xi}$ are the exact posterior mean vector and exact posterior covariance matrix evaluated at $N_\xi$ query points, 
$\widetilde{\bf m}_\xi$ and $\widetilde{\bf K}_{\xi,\xi}$ the estimates of ${\bf m}_\xi$ and ${\bf K}_{\xi,\xi}$ using each of the sampling methods, and
$\mathrm{tr}(\cdot)$ the trace of a square matrix.

For training, each approximation trial randomly generates a set of input samples over the interval $\left[\underline{x} + 0.2 \left( \overline{x}-\underline{x}\right), \underline{x} + 0.6 \left( \overline{x}-\underline{x}\right)\right]$.
The input training data are then normalized.
The output training data are also normalized using the minimum and maximum values of the $1$D Levy function in $[-10,10]$.
To compute $d_{\text{W}}$, we use $N_\xi = 2000$ query points evenly spaced on the normalized input space.
To consider the variation in the approximation introduced by the use of RFFs, each approximation trial uses 50 independent realizations of the feature maps, resulting in 50 corresponding values of $d_{\text{W}}$ for each set of training points. 
The number of RFFs is $N_\phi = 2000$.

\begin{figure}[t]
	\centering
	\includegraphics[width=\textwidth]{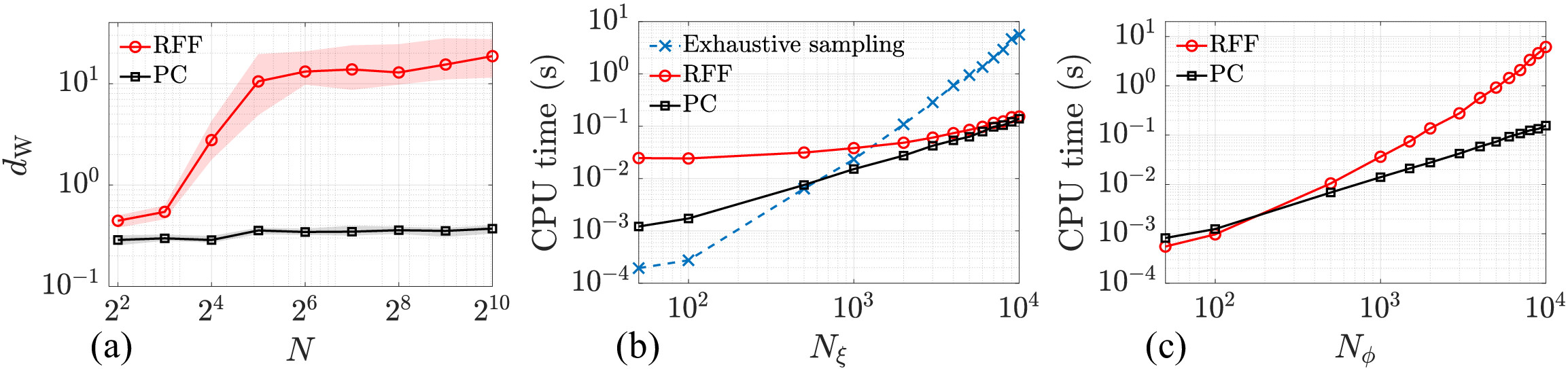}
	\caption{Comparisons of prediction performance and computational cost of exhaustive sampling, $\texttt{GP-RFF}$, and $\texttt{GP-PC}$. (a) Medians and interquartile ranges of the $2$-Wasserstein distance from 50 trials of each approximation. The distance values by the exhaustive sampling are zero and thus omitted from the log-scale plot. (b) CPU times for different numbers of query points. (c) CPU times for different numbers of RFFs.}
    \label{fig:approxerrors}
\end{figure}

\Cref{fig:approxerrors}(a) shows the medians and interquartile ranges of $d_{\text{W}}$ estimates by $\texttt{GP-RFF}$ and $\texttt{GP-PC}$ for different numbers of training points $N \in \{2^2, 2^3, \dots, 2^9, 2^{10} \}$.
The distance values by the exhaustive sampling are zero and thus omitted from the log-scale plot.
The corresponding GP mean and predictive intervals of the approximations by $\texttt{GP-RFF}$ and $\texttt{GP-PC}$ are provided in \cref{fig:prediction_rff_pc} of Appendix~\ref{AppF}.
We see that $d_{\text{W}}$ by $\texttt{GP-RFF}$ tends to increase as the number of training points increases. This behavior can be attributed to the fact that $\texttt{GP-RFF}$ is ill-behaved in extrapolation regions when the number of training points increases \citep{Mutny2018,Wilson2020}, see \cref{fig:prediction_rff_pc}.
In contrast, the distance by $\texttt{GP-PC}$ remains small and stable, regardless of changes in the number of training points.

We now compare the CPU times required by the exhaustive sampling, $\texttt{GP-RFF}$, and $\texttt{GP-PC}$ to draw posterior samples for different numbers of query points $N_\xi$ while fixing the number of RFFs at $N_\phi=1000$.
The sampling costs of $\texttt{GP-RFF}$ and $\texttt{GP-PC}$ for different numbers of RFFs $N_\phi$, with the number of query points fixed at $N_\xi = 1000$, are also compared.
In both cases, the number of training points is $N=20$.
\Cref{fig:approxerrors}(b) confirms that for a small number of training points the computational cost of exhaustive sampling scales cubically in $N_\xi$, while that of $\texttt{GP-RFF}$ and $\texttt{GP-PC}$ is linear in $N_\xi$. 
\Cref{fig:approxerrors}(c) shows that the computational cost of $\texttt{GP-PC}$ is linear in $N_\phi$ while that of $\texttt{GP-RFF}$ scales cubically in $N_\phi$ when the posterior samples of the weights are not generated by an acceleration technique that uses the SMW formula and the Cholesky decomposition of an $N$-by-$N$ matrix \citep{Seeger2007}. While $\texttt{GP-RFF}$ can obtain a linear cost in $N_\phi$ using this acceleration technique, it is still slightly slower than $\texttt{GP-PC}$, see \cref{fig:timeRFFPCacceleration} of Appendix~\ref{AppF}.

To this end, the exhaustive sampling from the function-space view and $\texttt{GP-RFF}$ from the weight-space view exhibit opposing strengths and weaknesses. 
Designed by combining the two perspectives on GPs, $\texttt{GP-PC}$ inherits the accuracy of the exhaustive sampling and improves upon the computational efficiency of $\texttt{GP-RFF}$.

\section{Gaussian process samples for global sensitivity analysis}
\label{sec5}

Consider a model of interest $c({\bf x})$ that can be, for example, a model to evaluate displacements of a structure.
As discussed in \Cref{Sec1}, global sensitivity analysis (GSA) studies how variations in ${\bf x}$ influence the model output $c({\bf x})$ considering the entire range of possible values of ${\bf x}$ \citep{Saltelli2008,Saltelli2010}.
An important result of GSA is the relative importance of each input variable $x_i$, $i \in \{1,\dots,d\}$, in computing $c({\bf x})$.
This section reviews Sobol' indices \citep{Sobol1993,Sobol2001} for GSA and then describes how to compute them using samples from a GP posterior of $c({\bf x})$.

\subsection{Overview of Sobol' indices}

Assuming that $x_i$, $i \in \{1,\dots,d\}$, are independent random variables, each characterized by a known PDF $p(x_i)$. Sobol' indices are defined based on the decomposition of the variance of $c({\bf x})$, denoted as $\mathbb{V} \left[ c({\bf x}) \right]$, into contributions from the input variables and their potential interactions.
The greater the contribution of a variable to $\mathbb{V} \left[ c({\bf x}) \right]$,  the more significant its role in computing $c({\bf x})$.

Following the notations by \cite{DaVeiga2021}, let $\mathcal{S}_d = \{1,\dots,d\}$ be the set of input variable indices and $\mathcal{S} \subseteq  \mathcal{S}_d$ any subset of $\mathcal{S}_d$.
Let ${\bf x}_\mathcal{S}$ represent the vector of input variables with indices in $\mathcal{S}$ and ${\bf x}_{\sim \mathcal{S}}$ the complementary vector with indices not in $\mathcal{S}$.
Sobol' decomposition of $\mathbb{V} \left[ c({\bf x}) \right]$ relies on the following functional analysis of variance (ANOVA) decomposition of $c({\bf x})$ \citep{Sobol1993,Sobol2001,Hooker2004}: there exists functions $c_\mathcal{S}({\bf x}_{\mathcal{S}})$, $\mathcal{S} \subseteq \mathcal{S}_d$,
such that
\begin{equation} \label{eqn:anova}
	c({\bf x}) = \sum_{\mathcal{S} \subseteq \mathcal{S}_d} c_\mathcal{S}({\bf x}_{\mathcal{S}}),
\end{equation}
where
\begin{equation} \label{eqn:orthogonal}
	\mathbb{E}_{x_i}\left[ c_\mathcal{S}({\bf x}_{\mathcal{S}}) \right] = 0, \ \ i \in \mathcal{S},
\end{equation}
which implies that  the terms $c_\mathcal{S}$ in \cref{eqn:anova} are mutually orthogonal.
As a result, we have \citep{Sobol1993,Sobol2001,DaVeiga2021}
\begin{equation} \label{eqn:decompositionproperties}
	\begin{aligned} 
		\mathbb{E}\left[ c({\bf x}) \right]  & = c_\emptyset, \\
		c_\mathcal{S}({\bf x}_{\mathcal{S}}) & = \sum_{\mathcal{V} \subset \mathcal{S}} (-1)^{|\mathcal{S}|-|\mathcal{V}|} \mathbb{E}\left[ c({\bf x})| {\bf x}_{\mathcal{V}}\right].
	\end{aligned}
\end{equation}
Here, the intercept term $c_\emptyset$ corresponds to $\mathcal{S} = \{\emptyset\}$, and $|\mathcal{S}|$ and $|\mathcal{V}|$ the size of $\mathcal{S}$ and $\mathcal{V}$, respectively.

Assuming that $c({\bf x})$ is square integrable. \Cref{eqn:anova,eqn:orthogonal} lead to the following decomposition of $\mathbb{V} \left[ c({\bf x}) \right]$:
\begin{equation} \label{eqn:variancedecomposition}
	\mathbb{V} \left[ c({\bf x}) \right] = \sum_{\mathcal{S} \subseteq \mathcal{S}_d } \mathbb{V} \left[ c_\mathcal{S}({\bf x}_{\mathcal{S}}) \right],
\end{equation}
where 
\begin{equation} \label{eqn:componentvariance}
	\mathbb{V} \left[ c_\mathcal{S}({\bf x}_{\mathcal{S}}) \right] = \sum_{\mathcal{V} \subset \mathcal{S}} (-1)^{|\mathcal{S}|-|\mathcal{V}|} \mathbb{V} \left[\mathbb{E}\left[ c({\bf x})| {\bf x}_{\mathcal{V}}\right] \right].
\end{equation}

From \cref{eqn:variancedecomposition}, we can define the Sobol’ sensitivity index associated with $\mathcal{S}$, as
\begin{equation} \label{eqn:indices}
	S_\mathcal{S} = \frac{\mathbb{V} \left[ c_\mathcal{S}({\bf x}_{\mathcal{S}}) \right]}{\mathbb{V} \left[ c({\bf x}) \right]},
\end{equation}
and the total Sobol’ index associated with $\mathcal{S}$, as
\begin{equation} \label{eqn:totalindices}
	S_\mathcal{S}^T = \sum_{\mathcal{V} \subseteq \mathcal{S}_d, \mathcal{V} \bigcap \mathcal{S} \neq \emptyset} S_\mathcal{V},
\end{equation}
which includes $S_\mathcal{S}$ and the sensitivity indices associated with other sets that are the subsets of $\mathcal{S}_d$ and intersect $\mathcal{S}$. 
 
The \textit{first-order index} measures the contribution of a single input variable $x_i$ to the output variance independently of the other variables. In other words, it measures the expected reduction in $ \mathbb{V} \left[ c({\bf x}) \right]$ when $x_i$ is fixed.
From \cref{eqn:componentvariance,eqn:indices}, the first-order index for $x_i$ is defined as \citep{Sobol1993,Sobol2001,Saltelli2008,Saltelli2010}
\begin{equation} \label{eqn:firstindex}
    S_i = \frac{\mathbb{V} \left[ \mathbb{E} \left[ c({\bf x})|{x_i} \right] \right]}{\mathbb{V} \left[ c({\bf x}) \right]},
\end{equation}
where the inner expectation of the numerator is for a fixed value of $x_i$ and the outer variance considers all possible values of $x_i$.

The \textit{total-effect index} measures the total contribution of a single input variable, including all interactions with other variables.
It encapsulates the first-order effect and all higher-order effects involving $x_i$. 
From \cref{eqn:variancedecomposition,eqn:totalindices,eqn:firstindex}, we have \citep{Saltelli2008,Saltelli2010}
\begin{equation} \label{eqn:totalindex}
        S_i^T  = \sum_{\mathcal{V} \subseteq \mathcal{S}_d, i \in \mathcal{V}} S_i 
         = 1 - \frac{\mathbb{V} \left[ \mathbb{E}\left[ c({\bf x})| {\bf x}_{\sim i} \right] \right]}{\mathbb{V} \left[ c({\bf x}) \right]} 
        = \frac{\mathbb{E} \left[ \mathbb{V}\left[ c({\bf x})| {\bf x}_{\sim i} \right] \right]}{\mathbb{V} \left[ c({\bf x}) \right]},
\end{equation}
where the inner variance of $\mathbb{E} \left[ \mathbb{V}\left[ c({\bf x})| {\bf x}_{\sim i} \right] \right]$ is taken for a fixed value of the complement vector ${\bf x}_{\sim i}$, while the outer expectation considers all possible values of ${\bf x}_{\sim i}$.

\subsection{Numerical approximation of Sobol' indices}

The common practice for computing Sobol' indices is via the Monte Carlo estimates of $S_i$ and $S_i^T$ \citep{Sobol1993,Sobol2001,Saltelli2008,Saltelli2010,WeiP2012}.
Let ${\bf A} \in \mathbb{R}^{N_\text{x} \times d}$ and ${\bf B} \in \mathbb{R}^{N_\text{x} \times d}$ represent two matrices, each containing the Monte Carlo samples of ${\bf x}$ generated from $p(x_i)$, $i \in \{1,\dots,d\}$. The number of input variable samples $N_\text{x}$ can be a few thousands \citep[Section 4.6]{Saltelli2008}.
Define ${\bf A}^{(i)}_{\bf B}$, $i \in \{1,\dots,d\}$, as the matrix whose columns are columns of ${\bf A}$ except the $i$th column, which is from ${\bf B}$.
Let ${\bf c}({\bf A}) \in \mathbb{R}^{N_\text{x}}$, ${\bf c}({\bf B}) \in \mathbb{R}^{N_\text{x}}$, and ${\bf c}({\bf A}^{(i)}_{\bf B}) \in \mathbb{R}^{N_\text{x}}$ represent the vectors containing the values of $c({\bf x})$ evaluated at the input samples in ${\bf A}$, ${\bf B}$, and ${\bf A}^{(i)}_{\bf B}$, respectively.

With ${\bf c}({\bf A})$, ${\bf c}({\bf B})$, and ${\bf c}({\bf A}^{(i)}_{\bf B})$, we can approximate $\mathbb{V} \left[ c({\bf x})\right]$, $\mathbb{V} \left[ \mathbb{E} \left[ c({\bf x})|{x_i} \right] \right]$, and $\mathbb{E} \left[ \mathbb{V}\left[ c({\bf x})| {\bf x}_{\sim i} \right] \right]$ in \cref{eqn:firstindex,eqn:totalindex} to compute the first-order and total-effect indices for $x_i$.
Specifically, $\mathbb{V} \left[ c({\bf x}) \right]$  is approximated as
\begin{equation} \label{eqn:aprxvariance} 
	\mathbb{V} \left[ c({\bf x}) \right] \approx \widehat{\mathbb{V}} \left[ \left[ {\bf c}({\bf A});{\bf c}({\bf B}) \right] \right],
\end{equation}
where $\left[ {\bf c}({\bf A});{\bf c}({\bf B}) \right]$ concatenates ${\bf c}({\bf A})$ and ${\bf c}({\bf B})$, and $\widehat{\mathbb{V}}[\cdot]$ the sample variance.
The estimator for $\mathbb{V} \left[ \mathbb{E} \left[ c({\bf x})|{x_i} \right] \right]$ is \citep{Saltelli2010}
\begin{equation} \label{eqn:aprxnumfirstindex} 
	\mathbb{V} \left[ \mathbb{E} \left[ c({\bf x})|{x_i} \right] \right] \approx \frac{1}{N_\text{x}} \sum_{j=1}^{N_\text{x}} {\bf c}({\bf B})_j \left({\bf c}({\bf A}^{(i)}_{\bf B})_j - {\bf c}({\bf A})_j \right),
\end{equation}
where subscript $j$ represents the $j$th entry of the vector of interest.
Finally, $\mathbb{E} \left[ \mathbb{V}\left[ c({\bf x})| {\bf x}_{\sim i} \right] \right]$ can be estimated by \citep{Jansen1999}
\begin{equation} \label{eqn:aprxnumtotalindex} 
	\mathbb{E} \left[ \mathbb{V}\left[ c({\bf x})| {\bf x}_{\sim i} \right] \right] \approx \frac{1}{2N_\text{x}} \sum_{j=1}^{N_\text{x}} \left( {\bf c}({\bf B})_j - {\bf c}({\bf A}^{(i)}_{\bf B})_j \right)^2
\end{equation}
Detailed derivations of \cref{eqn:aprxnumfirstindex,eqn:aprxnumtotalindex} can be found in \cite{Saltelli2010}.
Alternatives to \cref{eqn:aprxnumfirstindex,eqn:aprxnumtotalindex} are given in \cite{Saltelli2010}, Table~2.

\begin{algorithm}[t]
	\caption{\texttt{GP-GSA}: GP posterior samples for global sensitivity analysis}
	\label{alg:GP-GSA}
	\begin{algorithmic}[1]
		\State \textbf{Input:} dataset $\mathcal{D}$ containing observations of $c({\bf x})$, stationary covariance function $\kappa(\cdot,\cdot)$, number of features $N_\phi$, standard deviation of observation noise $\sigma_\text{n}$, distributions $p(x_i)$, number of input variable samples $N_\text{x}$, number of GP posterior samples $N_\text{s}$
		\State Generate ${\bf A}$, ${\bf B}$, and ${\bf A}_{\bf B}^{(i)}$, $i \in \{1,\dots,d\}$, from $p(x_i)$ and $N_\text{x}$ \label{alg:GP-GSA-line2}
		\For {$k=1:N_\text{s}$}
		
		\State $\widetilde{c}({\bf x}) \gets \begin{cases}
			\texttt{GP-RFF} \left({\bf x}, \mathcal{D},\kappa(\cdot,\cdot), N_\phi, \sigma_\text{n} \right) \\
			\texttt{GP-PC} \left({\bf x}, \mathcal{D},\kappa(\cdot,\cdot), N_\phi, \sigma_\text{n} \right)
		\end{cases}$ \Comment{posterior sample function}  \label{alg:GP-GSA-line4}
		
		\State Compute $\widetilde{{\bf c}}({\bf A})$, $\widetilde{{\bf c}}({\bf B})$, $\widetilde{{\bf c}}({\bf A}_{\bf B}^{(i)})$
		\State Compute $\mathbb{V} \left[ \widetilde{c}({\bf x}) \right]$, see \cref{eqn:aprxvariance} 
		\For {$i=1:d$}
			\State Compute $\mathbb{V} \left[ \mathbb{E} \left[ \widetilde{c}({\bf x})|{x_i} \right] \right]$, see \cref{eqn:aprxnumfirstindex} 
			\State Compute $\mathbb{E} \left[ \mathbb{V}\left[ \widetilde{c}({\bf x})| {\bf x}_{\sim i} \right] \right]$, see \cref{eqn:aprxnumtotalindex} 
			\State Compute $S_i^k$, see \cref{eqn:firstindex} 
			\State Compute $S_i^{T,k}$, see \cref{eqn:totalindex} 
		\EndFor
	
		\EndFor
		
		\State \textbf{return} ${\bf S}_i=\{S_i^k\}_{k=1}^{N_\text{s}}$, ${\bf S}^T_i=\{S_i^{T,k}\}_{k=1}^{N_\text{s}}$ \label{alg:GP-GSA-line14}
	\end{algorithmic}
\end{algorithm}

\subsection{Approximate Sobol' indices via Gaussian process samples}

The main hurdle in computing Sobol' indices via \cref{eqn:aprxvariance,eqn:aprxnumfirstindex,eqn:aprxnumtotalindex} is the large number of $c({\bf x})$ evaluations.
In fact, computing both $\mathbb{V} \left[ c({\bf x}) \right]$ and $\mathbb{V} \left[ \mathbb{E} \left[ c({\bf x})|{x_i} \right] \right]$ (or $\mathbb{E} \left[ \mathbb{V}\left[ c({\bf x})| {\bf x}_{\sim i} \right] \right]$)  requires $2N_\text{x}+N_\text{x}d$ evaluations in which $2N_\text{x}$ is for computing ${\bf c}({\bf A})$ and ${\bf c}({\bf B})$, and $N_\text{x}d$ for computing ${\bf c}({\bf A}^{(i)}_{\bf B})$, $i \in \{1,\dots,d\}$.
This is further emphasized when $c({\bf x})$ is an expensive-to-evaluate black-box function. 
Fortunately, the predictions of Sobol' indices and uncertainty in these predictions can be obtained via GP models of $c({\bf x})$ that require a few to several hundred model observations \citep[see e.g.,][]{LeGratiet2014,LeGratiet2016}.
The idea is to draw multiple GP posterior sample functions as the proxies for $c({\bf x})$ and use them to compute uncertainty-aware sensitivity indices.

Let $N_\text{s}$ denote the number of posterior sample functions drawn from the GP of $c({\bf x})$.
With these functions, a pair of matrices ${\bf A}$ and ${\bf B}$ provides $N_\text{s}$ values of the first-order (or total-effect) sensitivity index for each input variable.
The median and interquartile range computed from these values can be used as an estimate of the sensitivity index of interest and its uncertainty measure, respectively.
\Cref{alg:GP-GSA} summarizes the implementation that outputs $N_\text{s}$ values of each type of Sobol' indices for each input variable (Line 14), given ${\bf A}$, ${\bf B}$, and $N_\text{s}$ posterior sample functions for the model $c({\bf x})$.

Two key factors contribute to uncertainty in the sensitivity indices obtained from \Cref{alg:GP-GSA}: (1) the accuracy of the GP model and (2) the size of matrices ${\bf A}$ and ${\bf B}$.
The former can be mitigated by either increasing the number of training points for the GP model \citep{LeGratiet2014,LeGratiet2016} or utilizing an adaptive sampling technique \citep{LiuH2018,Steiner2019,Chauhan2024}.
The latter, known as the pick-freeze problem, can be addressed by setting $N_\text{x}$ greater or equal to a minimal value at which the pick-freeze approximation error becomes negligible compared to the approximation error \citep{LeGratiet2014}.

With the covariance functions in \cref{SEcovariancefunction,Materncovariancefunction}, one may use values of the separate length scale parameters as an indicator for the relative importance of the input variables \citep{Welch1992,Yousefpour2024}. Intuitively, a large length scale parameter implies a small variation of the model output along the associated input dimension, suggesting a small influence of the corresponding input variable. However, this interpretation makes it difficult to distinguish whether a given length scale parameter reflects the relative importance of the associated input variable or the nonlinearity of the model with respect to that variable, as the length scale parameter is closely related to the nonlinearity in the output model \citep{Paananen2019}.

\section{Gaussian process samples for single-objective optimization}
\label{sec6}

In this section, we focus on solving the following single-objective optimization (SO) problem:
\begin{equation}\label{eqn:SOformulation}
	\underset{{\bf x} \in \mathcal{X}}{\min} \ \  c({\bf x}),
\end{equation}
where $c({\bf x})$ now represents a black-box objective function whose values are evaluated via high-fidelity simulations or physical experiments.
Since this observation mechanism provides no useful information about the derivatives of $c({\bf x})$, it discourages the use of any gradient-based optimizer \citep[see e.g.,][]{Nocedal2006}. 
The feasible set $\mathcal{X} \subset \mathbb{R}^d$ is assumed to be a hypercube $\mathcal{X} = \prod_{i=1}^d \left[\underline{x}_i,\overline{x}_i\right]$, where $\underline{x}_i$ and $\overline{x}_i$ are the lower and upper bounds of $x_i$, respectively.

We start this section with a brief overview of Bayesian optimization (BO) for solving problem~(\ref{eqn:SOformulation}).
We then detail how GP posterior sample functions can be used to solve this problem via a simple yet efficient BO method called Thompson sampling \citep{Thompson1933,Russo2018}.

\subsection{Overview of Bayesian optimization} 

BO is a successful data-driven approach to solving optimization problems with expensive-to-evaluate objective functions \citep[see e.g.,][]{Jones1998,Snoek2012,Shahriari2016,Frazier2018,Balandat2020,Garnett2023}.
Countless applications of BO can be found in science and engineering (see e.g., \cite{Garnett2023}, Appendix D).
At its heart, BO solves the optimization problem using a probabilistic model of the objective function, which is often a GP.
Starting with a prior probabilistic model and a few observations, BO determines a posterior model that represents our beliefs about the characteristics of the objective function. 
From this posterior model, BO formulates an acquisition function that is much easier to evaluate and differentiate than the objective function.
The acquisition function is then optimized, and the resulting solution is assigned as a new candidate solution at which the objective function is evaluated for the next BO iteration. 
The process is repeated until some predefined termination criteria are met.

Each acquisition function is determined by an optimization policy that maps what we value most in the current observations to each point in the feasible space.
What we value in establishing such an optimization policy often represents our expectation for an improvement in the objective function value, a gain in our knowledge of the true optimizer, or a gain in our knowledge of the true optimal value.
To meet such diverse expectations, we need many BO acquisition functions, for example, probability of improvement (PI) \citep{Kushner1964}, expected improvement (EI) \citep{Jones1998}, knowledge gradient \citep{Frazier2008}, upper confidence bound (GP-UCB) \citep{Srinivas2010}---or equivalently lower confidence bound (LCB) for minimization problems, GP Thompson sampling (GP-TS) \citep{Russo2014,Chowdhury2017}, and mutual information \citep{Villemonteix2009,Hennig2012,HernandezLobato2014,WangZ2017,Hvarfner2022}.
More recent developments include likelihood-weighted \citep{Blanchard2021}, LogEI \citep{Ament2023},
epsilon-greedy GP-TS \citep{Do2024jcise}, and GP-TS accelerated by rootfinding \citep{Adebiyi2024bdu,Adebiyi2025tsroots}.
Details of notable BO acquisition functions and their properties are surveyed in \cite{Do2025mfbo}.

In addition to their impressive empirical performance, many BO methods are supported by strong theoretical performance guarantees \citep[see e.g.,][]{Srinivas2010,Bull2011,May2012,Russo2014,Chowdhury2017}.
The reader may consult \cite{Garnett2023}, Chapter 10, for detailed discussions on the theoretical performance of several notable BO algorithms.

\subsection{Thompson sampling for Bayesian optimization} 
GP-TS is an extension of the classical TS in finite-armed bandit problems to continuous settings of BO.
The classical TS selects actions from a finite set of actions over a limited number of iterations.
Each action corresponds to a stochastic reward drawn from an unknown distribution, which is learned after each reward value is revealed.

In the continuous setting of BO, GP-TS generates a sequence of new candidate solutions (i.e., actions) by randomly sampling from unknown posterior distributions $p(\bf{x}^\star|\mathcal{D})$ of the global minimum location $\bf{x}^\star$, which is due to our imperfect knowledge of the objective function encapsulated in the probabilistic model.
As $\bf{x}^\star|\mathcal{D}$ is fully determined by the GP posterior, GP-TS simply draws a posterior sample function and minimizes it to obtain a sample of $\bf{x}^\star|\mathcal{D}$, rather than finding $p(\bf{x}^\star|\mathcal{D})$ and sampling from it.
The objective function value (i.e., reward) for the generated sample is revealed before the posterior is updated.
Thus, we can view GP-TS as a BO method that recommends a new candidate solution by optimizing a stochastic acquisition function drawn from the GP posterior of the objective function.
The advantage of such a stochastic acquisition function is that it is easy to implement, has a closed-form expression of derivatives, and can enjoy a parallel workload to speed up the optimization \citep[see e.g.,][]{Shah2015,HernandezLobato2017,Kandasamy2018}.

\begin{algorithm}[t]
	\caption{\texttt{GP-TS-SO}: GP posterior samples for single-objective optimization}
	\label{alg:GP-TS-SO}
	\begin{algorithmic}[1]
		\State \textbf{Input:} mechanism to compute the objective function $c({\bf x})$, dataset $\mathcal{D}^0 =\{ ({\bf x}^i,y^i)\}_{i=1}^N$, stationary covariance function $\kappa(\cdot,\cdot)$, number of features $N_\phi$, standard deviation of observation noise $\sigma_\text{n}$, threshold for number of optimization iterations $K$
	\State $\left( {\bf x}_{\min},y_{\min}  \right) \gets \min \{y^i, i=1\dots,N \}$

        \For {$k=1:K$}
        
        \State $\alpha({\bf x}) \gets \begin{cases}
        \texttt{GP-RFF} \left({\bf x}, \mathcal{D}^{k-1},\kappa(\cdot,\cdot), N_\phi, \sigma_\text{n} \right) \\
        \texttt{GP-PC} \left({\bf x}, \mathcal{D}^{k-1},\kappa(\cdot,\cdot), N_\phi, \sigma_\text{n} \right)
        \end{cases}$ \Comment{posterior sample function} \label{alg:GP-TS-SO-line4}
        
        \State ${\bf x}^{k} \gets \underset{{\bf x} \in \mathcal{X}}{\mathrm{arg\,min}} \ \ \alpha({\bf x}|\mathcal{D}^{k-1})$ s.t. ${\bf x} \notin \mathcal{D}^{k-1}$ \Comment{new query point} \label{alg:GP-TS-SO-line5}

        \State $y^{k} \gets c({\bf x}^{k}) + \varepsilon_\text{n}^k$, $\varepsilon_\text{n}^k \sim \mathcal{N}\left( 0, \sigma^2_\text{n}\right)$ \Comment{new observation}
        
	\State $\mathcal{D}^k\gets\mathcal{D}^{k-1} \cup \{ \left( {\bf x}^{k},y^{k} \right) \}$
	\State $\left( {\bf x}_{\min},y_{\min} \right) \gets \min\{y_{\min},y^{k}\}$
  
        \EndFor
		
		\State \textbf{return} $\left( {\bf x}_{\min},y_{\min} \right)$
	\end{algorithmic}
\end{algorithm}

\texttt{GP-TS-SO} in \Cref{alg:GP-TS-SO} outlines the implementation of GP-TS.
In each optimization iteration, a posterior sample function $\alpha({\bf x})$ is drawn in Line 4 by either $\texttt{GP-RFF}$ or $\texttt{GP-PC}$.
This function is then minimized in Line 5 for a new candidate solution ${\bf x}^{k}$. The optimization constraint ${\bf x} \notin \mathcal{D}^{k-1}$ is to ensure that the observed points are not reselected.

While a direct search over a discretized domain $\mathcal{X}$ can be used to optimize GP-TS acquisition functions in Line 5 of \Cref{alg:GP-TS-SO} \citep[see e.g.,][]{Balandat2020}, a gradient-based optimizer over a continuous domain $\mathcal{X}$ is often preferred due to its fast convergence and reliable performance \citep[see e.g.,][]{Adebiyi2025tsroots}.
Such an optimizer can also accommodate multi-start strategies, which require a set of starting points to improve the chance of finding a global optimum.
Oftentimes, the starting points are chosen as uniformly distributed samples over the input variable space \citep{Frazier2018} or as random samples from a Latin hypercube design \citep{WangJ2020}.
Notably, \cite{Adebiyi2024bdu,Adebiyi2025tsroots}, based on useful properties of the posterior functions in the interpolation and extrapolation regions, designed effective sets of starting points for gradient-based multistart optimizers, which results in scalable and accurate optimization of GP-TS and excellent performance of BO. 

\begin{figure*}[t]
	\centering
	\includegraphics[width=\textwidth]{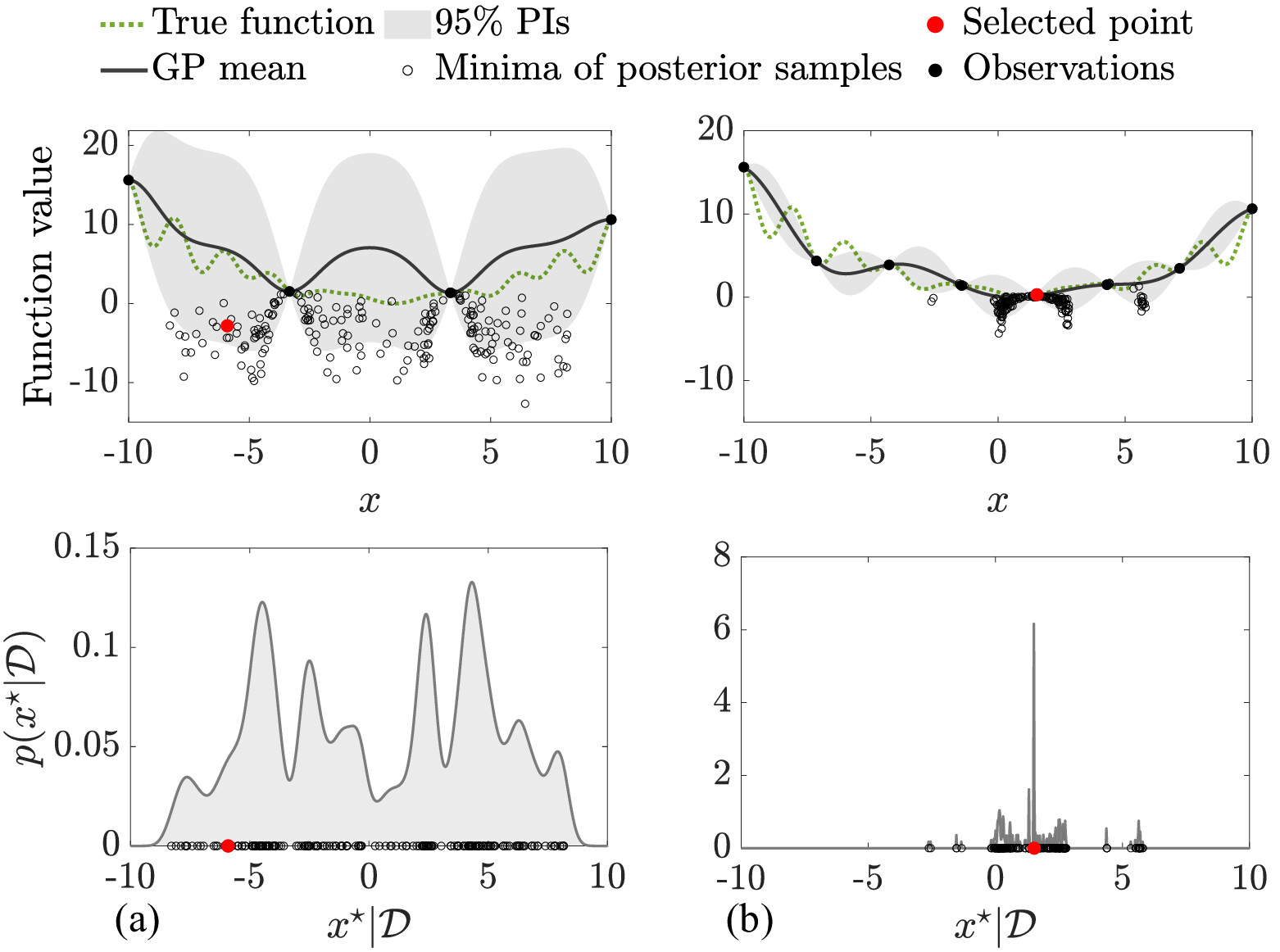}
	\caption{Illustration of GP-TS that selects the new candidate solution as a point randomly sampled from $p(x^\star|\mathcal{D})$, which is equivalent to selecting the global minimum of a function randomly sampled from the posterior of the objective function. (a) Exploration. (b) Exploitation.}
    \label{fig:GPTS}
\end{figure*}

GP-TS naturally manages the exploitation--exploration trade-off. As shown in \cref{fig:GPTS}(a), it favors exploration when the number of observations is insufficient to provide reliable information about the optimal solution $\bf{x}^\star$. During this exploration phase, the algorithm diversifies its selection of the new candidate solutions, which is depicted by the wide spread of $p(\bf{x}^\star|\mathcal{D})$ in \cref{fig:GPTS}(a).
As the number of observations increases and GP-TS gains more knowledge about the objective function, it transitions toward exploitation. In this phase, GP-TS is likely to select new candidate solutions in the neighborhood of the minimum of the GP mean function where $p(\bf{x}^\star|\mathcal{D})$ is concentrated, as shown in \cref{fig:GPTS}(b). 

\section{Gaussian process samples for multi-objective optimization}
\label{sec7}

We now consider the following multi-objective optimization (MO) problem:
\begin{equation}\label{eqn:moproblem}
	\underset{{\bf x} \in \mathcal{X}}{\min} \ \  \left[ c_1({\bf x}),\dots, c_{N_\text{c}}({\bf x})\right],
\end{equation}
where $c_1({\bf x}), \dots, c_{N_\text{c}}({\bf x})$ with $N_\text{c} \geq 2$ are the objective functions we wish to minimize.
Unlike problem (\ref{eqn:SOformulation}), problem (\ref{eqn:moproblem}) generally does not have a unique optimal solution because we cannot guarantee that all the objective functions are minimized at the same input variables ${\bf x}$. Instead, solving problem~(\ref{eqn:moproblem}) provides a set $\mathcal{P}_c$ of objective function values known as the Pareto front that quantifies the trade-off between the objectives. 
The image of $\mathcal{P}_c$ in $\mathcal{X}$ is the Pareto set $\mathcal{P}_x$.
Formal definitions of $\mathcal{P}_c$ and $\mathcal{P}_x$ can be found, for example, in \cite{Kochenderfer2019}, Chapter 12.

There are two main strategies for solving MO problems: decomposition and population methods.
The decomposition methods convert the MO problem with a vector-valued objective function to a series of SO problems, typically using either constraint-based or weight-based approaches \citep[see e.g.,][]{Haimes1971,Chalmet1986,Das1997}.
The population methods, starting from an initial population of candidate solutions, incrementally improve the quality of individuals spanning the Pareto front. 
One of the notable population-based multi-objective algorithms is the non-dominated sorting genetic algorithm II (NSGA-II) \citep{Deb2002}.

\subsection{Overview of multi-objective Bayesian optimization} 

Many multi-objective BO (MOBO) algorithms have been proposed for finding approximate Pareto fronts of problem~(\ref{eqn:moproblem}) when the objective functions are costly \citep[see e.g.,][]{Knowles2006,Couckuyt2012,Bradford2018,KonakovicLukovic2020,Daulton2020,Mathern2021,Do2022,Daulton2022}.
These algorithms can also be categorized into decomposition and population methods.
The former, such as ParEGO \citep{Knowles2006}, converts the objective functions to a single weighted-sum objective function.
This allows for the use of standard BO methods to identify a new candidate solution in each iteration.
By sweeping over the space of the weights, a set of solutions is incrementally built to approximate the Pareto front.

The latter treats the set of initial samples for MOBO as a small population that evolves.
This evolution is typically driven by a non-dominated sorting algorithm that identifies a set of non-dominated solutions from the current population to form the current approximate Pareto front, and an acquisition function that guides the search to improve the quality of the current approximate Pareto front.
Notable MOBO acquisition functions in this sphere include the expected hypervolume improvement \citep{Emmerich2006}, expected Euclidean distance improvement \citep{Keane2006}, $\mathcal{S}$-metric \citep{Ponweiser2008}, Pareto-frontier entropy search \citep{Suzuki2020}, and differentiable expected hypervolume improvement \citep{Daulton2020}.

Efficiently GP-TS via posterior sample functions of the objective functions can be used for MOBO \citep{Bradford2018,Maddox2021}.
One solution strategy is to apply GP-TS to the SO problems associated with all possible combinations of the weights from the weight-based approaches.
In the following section, we describe another solution strategy in which GP-TS is combined with a population algorithm, the hypervolume indicator, and a non-dominated sorting algorithm to find approximate Pareto fronts for MO problems.
Note that while we focus on uncorrelated objective functions, posterior sample functions can also be generated from multi-output GPs to consider correlations between the objective functions \citep{Maddox2021}.

\subsection{Thompson sampling for multi-objective Bayesian optimization}

Starting from initial datasets $\mathcal{D}_j^0 =\{ ({\bf x}^i,y_j^i)\}_{i=1}^N$, $j \in \{1,\dots,N_\text{c} \}$, that consist of $N$ input variable observations and the corresponding objective function observations, GP-TS iteratively finds an approximate Pareto front and the associated Pareto set via the following five steps:
\begin{enumerate}[{(1)}]
    \item Sort non-dominated solutions from the current datasets using a non-dominated sorting algorithm. The resulting solutions span the approximate Pareto front and approximate Pareto set.
    
    \item Draw $N_\text{c}$ sample functions from the posteriors of the objective functions using either $\texttt{GP-RFF}$ or $\texttt{GP-PC}$.

    \item With the resulting posterior sample functions that represent the objective functions, call a population algorithm to find the approximate Pareto front and the corresponding approximate Pareto set. Alternative MO optimizers can also be used.

    \item Use the hypervolume indicator to select a candidate solution from the approximate Pareto set obtained in (3) and determine the associated objective function values.

    \item Update the datasets with the candidate solution and the corresponding objective values, and repeat the optimization process from (1). \smallskip
    
\end{enumerate}

\begin{algorithm}[t]
	\caption{\texttt{GP-TS-MO}: GP posterior samples for multi-objective optimization}
	\label{alg:GP-TS-MO}
	\begin{algorithmic}[1]
		\State \textbf{Input:} mechanism to compute the objective functions $c_j({\bf x})$, $j \in \{1,\dots,N_\text{c}\}$, datasets $\mathcal{D}_j^0 =\{ ({\bf x}^i,y_j^i)\}_{i=1}^N$, stationary covariance function $\kappa(\cdot,\cdot)$, number of features $N_\phi$, standard deviation of observation noise $\sigma_{\text{n},j}$, threshold for number of optimization iterations $K$
	\State $\left( \mathcal{P}^0_x,\mathcal{P}^0_y  \right) \gets 
        \texttt{pareto\_sort}(\mathcal{D}_j^0)$ \Comment{sort non-dominated solutions} \label{alg:GP-TS-MO-line2}
        
        \For {$k=1:K$}
       
        \State $\widetilde{c}_j({\bf x}) \gets \begin{cases}
        \texttt{GP-RFF} \left({\bf x}, \mathcal{D}_j^{k-1},\kappa(\cdot,\cdot), N_\phi, \sigma_{\text{n},j} \right) \\
        \texttt{GP-PC} \left({\bf x}, \mathcal{D}_j^{k-1},\kappa(\cdot,\cdot), N_\phi, \sigma_{\text{n},j} \right)
        \end{cases}$  \Comment{posterior sample functions} \label{alg:GP-TS-MO-line4}
        
        \State $\left( \mathcal{\widetilde{P}}^k_x,\mathcal{\widetilde{P}}^k_c  \right) \gets \texttt{pareto\_sol}(\widetilde{c}_j({\bf x}))$ \label{alg:GP-TS-MO-line5}
        
        \State ${\bf r} \gets \texttt{ref}({\bf y}_j^{k-1})$ \Comment{reference point} \label{alg:GP-TS-MO-line6}
        
        \State ${\bf x}^{k} \gets \texttt{max\_hvi} \left( \mathcal{P}^{k-1}_y, \mathcal{\widetilde{P}}^k_c, \mathcal{\widetilde{P}}^k_x, {\bf r} \right)$ \Comment{new query point} \label{alg:GP-TS-MO-line7} 
        
        \State $y_j^{k} \gets c_j({\bf x}^{k}) + \varepsilon_{\text{n},j}^k$, $\varepsilon_{\text{n},j}^k \sim \mathcal{N}\left( 0, \sigma^2_{\text{n},j}\right)$ \Comment{new observations} \label{alg:GP-TS-MO-line8}
        
	  \State $\mathcal{D}_j^k\gets\mathcal{D}_j^{k-1} \cup \{ \left( {\bf x}^{k},y_j^{k} \right) \}$
 
	  \State $\left( \mathcal{P}^k_x,\mathcal{P}^k_y  \right) \gets 
        \texttt{pareto\_sort}(\mathcal{D}_j^k)$ \label{alg:GP-TS-MO-line10}
  
        \EndFor
		
		\State \textbf{return} $\left( \mathcal{P}^k_x,\mathcal{P}^k_y  \right)$
	\end{algorithmic}
\end{algorithm}

\texttt{GP-TS-MO} in \Cref{alg:GP-TS-MO} summarizes the implementation of the above five steps.
It has the following functions:
\begin{itemize}
    \item \texttt{pareto\_sort} in Lines 2 and 10 corresponds to step (1) that inputs the datasets and outputs $\mathcal{P}_y$, a set of non-dominated solutions, and $\mathcal{P}_x$, a set of the corresponding input variable points \citep[see e.g.,][]{TomR2019,Blank2020}.
    
    \item \texttt{pareto\_sol} in Line 5 implements step (3) that inputs the posterior sample functions $\widetilde{c}_j({\bf x})$ drawn in Line 4 and outputs the associated Pareto front $\mathcal{\widetilde{P}}_c$ and Pareto set $\mathcal{\widetilde{P}}_x$.

    \item \texttt{max\_hvi} in Line 7 corresponding to step (4) is to select a new candidate solution from $\mathcal{\widetilde{P}}_x$ for the objective evaluations in Line 8. This function is built upon the following important concepts in MO: hypervolume indicator, reference point, and hypervolume improvement.\smallskip
    
\end{itemize}

\begin{figure}[t]
	\centering
	\includegraphics[width=0.8\textwidth]{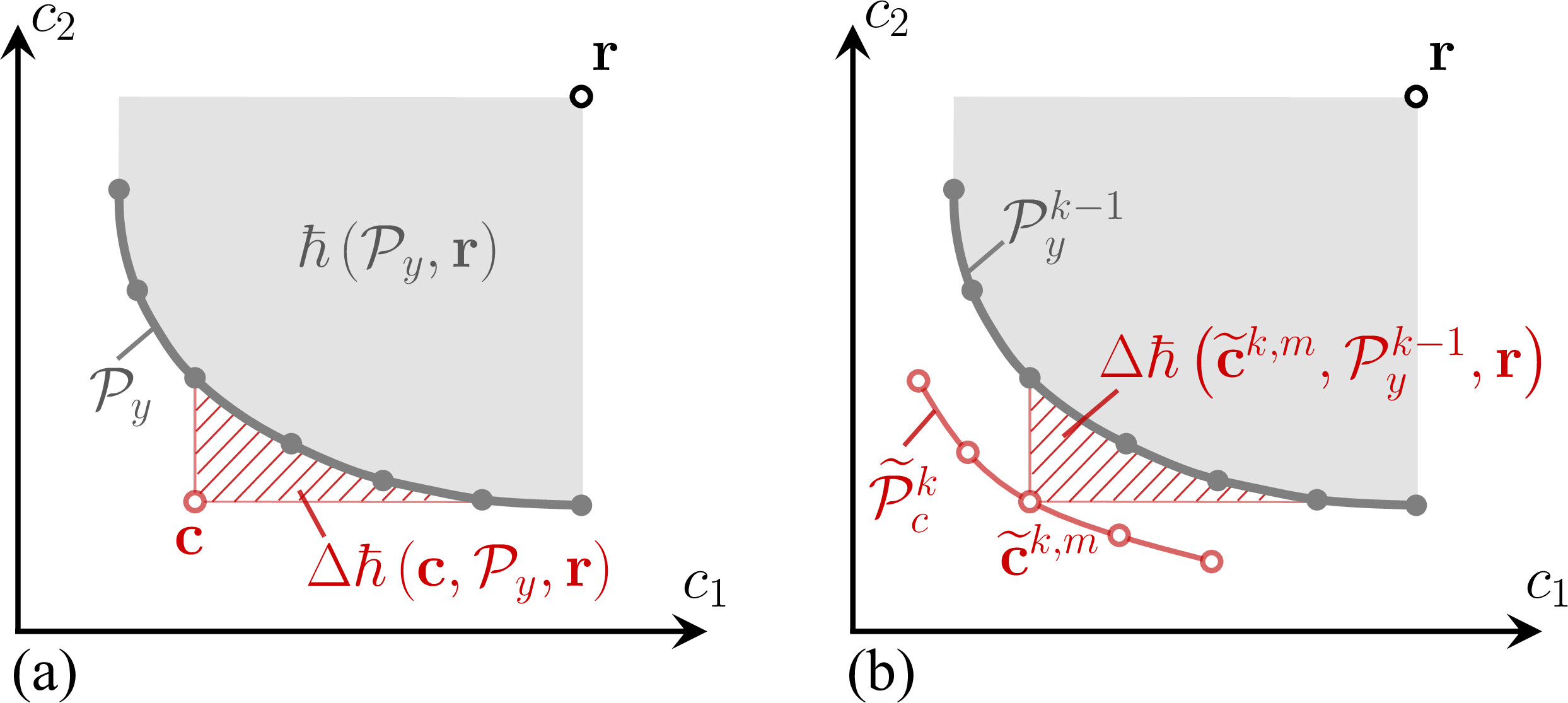}
	\caption{(a) Illustration of hypervolume indicator $\hbar$ and hypervolume improvement $\Delta\hbar$ for a bi-objective minimization problem. (b) Hypervolume improvement associated with a point from the approximate Pareto front for selecting a new query point using \texttt{GP-TS-MO}.}
    \label{fig:hypervolume}
\end{figure}

The hypervolume indicator \citep{Emmerich2006} measures the volume of the space dominated by all solutions spanning the Pareto front and bounded by a reference point ${\bf r} \in \mathbb{R}^{N_\text{c}}$.
It can be evaluated via a Monte Carlo estimation of the volume of the hypercuboid defined by the Pareto front and ${\bf r}$ \citep[see e.g.,][]{CaoY2008}.
For MO minimization problems, each element of ${\bf r}$ can be selected such that it is greater or equal to the maximum value of the corresponding objective values observed from the Pareto front.
Function \texttt{ref} in Line 4 of \texttt{GP-TS-MO} is to determine ${\bf r}$ in each optimization iteration.
It defines each element of ${\bf r}$ as the maximum value of each objective function observed in the current datasets. 
The shaded area in \cref{fig:hypervolume}(a) illustrates the hypervolume $\hbar(\mathcal{P}_y,{\bf r})$ for a Pareto front $\mathcal{P}_y$ of a bi-objective minimization problem.
Let ${\bf c} \in \mathbb{R}^{N_\text{c}}$ be a point in the objective space. The hypervolume improvement formulated for ${\bf c}$, $\mathcal{P}_y$, and ${\bf r}$, indicated by the hatched area in \cref{fig:hypervolume}(a), is defined as
\begin{equation}\label{eqn:hvi}
	\Delta \hbar({\bf c},\mathcal{P}_y,{\bf r}) = \hbar(\mathcal{P}_y \cup {\bf c},{\bf r})  - \hbar(\mathcal{P}_y,{\bf r}).
\end{equation}

Having all the necessary ingredients, we are ready to describe function \texttt{max\_hvi} in Line 7 of \texttt{GP-TS-MO} that is used to select a new query point ${\bf x}^k$ in the $k$th optimization iteration. Let $\widetilde{\bf c}^{k,m} \in \mathcal{\widetilde{P}}_c^k$, $m \in \{1,\dots,M\}$, represent a point in the approximate Pareto front $\mathcal{\widetilde{P}}_c^k$, see \cref{fig:hypervolume}(b), and ${\bf x}^{k,m} \in \mathcal{\widetilde{P}}_x^k$ the corresponding point in the approximate Pareto set $\mathcal{\widetilde{P}}_x^k$, where $\mathcal{\widetilde{P}}_c^k$ and $\mathcal{\widetilde{P}}_x^k$ are obtained from Line 5 of \texttt{GP-TS-MO}.
Function \texttt{max\_hvi} assigns ${\bf x}^{k,m} \in \mathcal{\widetilde{P}}_x^k$ to ${\bf x}^{k}$ such that $\widetilde{\bf c}^{k,m}$ (associated with ${\bf x}^{k,m}$)  has the maximum hypervolume improvement $\Delta \hbar(\widetilde{\bf c}^{k,m},\mathcal{P}_y^{k-1},{\bf r})$ among those computed from the members of $\mathcal{\widetilde{P}}_c^k$, where $\mathcal{P}_y^{k-1}$ is the approximate Pareto front found so far. Notationally, 
\begin{equation}\label{eqn:newquery}
	{\bf x}^{k}= \left\{ {\bf x}^{k,m} \in \mathcal{\widetilde{P}}_x^k \Big| \widetilde{\bf c}^{k,m} \coloneqq  \underset{\widetilde{\bf c}^{k,m} \in \mathcal{\widetilde{P}}_c^k} \argmax \, \Delta \hbar(\widetilde{\bf c}^{k,m},\mathcal{P}_y^{k-1},{\bf r}) \right\}.
\end{equation}
Here, we expect to refine the GPs for the objective functions at locations with the highest score of improving the quality of the current approximate Pareto front.

\section{Numerical examples}
\label{sec8}

In all numerical examples in this section, we use the SE covariance function in \cref{SEcovariancefunction}.
If the goal is to select a covariance function from a set of different covariance functions that best explains the data, a principled approach is to maximize the so-called model evidence, as discussed in \cref{sec101}.
In \cref{sec81,sec83}, we find an optimal set of GP hyperparameters in Line 2 of $\texttt{GP-RFF}$ and $\texttt{GP-PC}$ using the DACE toolbox \citep{Lophaven2002}, which is a MATLAB toolbox.
In \cref{sec82}, we find the optimal hyperparameters using the GPyTorch package \citep{Gardner2018}.

\subsection{Global sensitivity analysis}
\label{sec81}

We first implement $\texttt{GP-GSA}$ (\Cref{alg:GP-GSA}) to compute approximate global sensitivity indices for the input variables of the Ishigami function and a ten-bar truss structure.

\subsubsection{Ishigami function}
\label{sec811}

The Ishigami function \citep{Ishigami1990} is given as follows:
\begin{equation}
    c({\bf x}) = \sin(x_1) + 7 \sin^2(x_2) + 0.1 x_3^4 \sin(x_1),
\end{equation}
where $x_i \sim \mathcal{U}(-\pi,\pi)$, $i \in \{ 1, 2, 3\}$.
The coupling between $x_1$ and $x_3$ in the third term introduces the interaction between these variables.
The exact values of the first-order and total-effect indices for the three variables are $S_1 = 0.3138$, $S_2 = 0.4424$, $S_3 = 0$, $S^T_1 = 0.5574$, $S^T_2 = 0.4424$, and $S^T_3 = 0.2436$ \citep[see e.g.,][]{Sudret2008}.

\begin{figure*}[t]
	\centering
	\includegraphics[width=\textwidth]{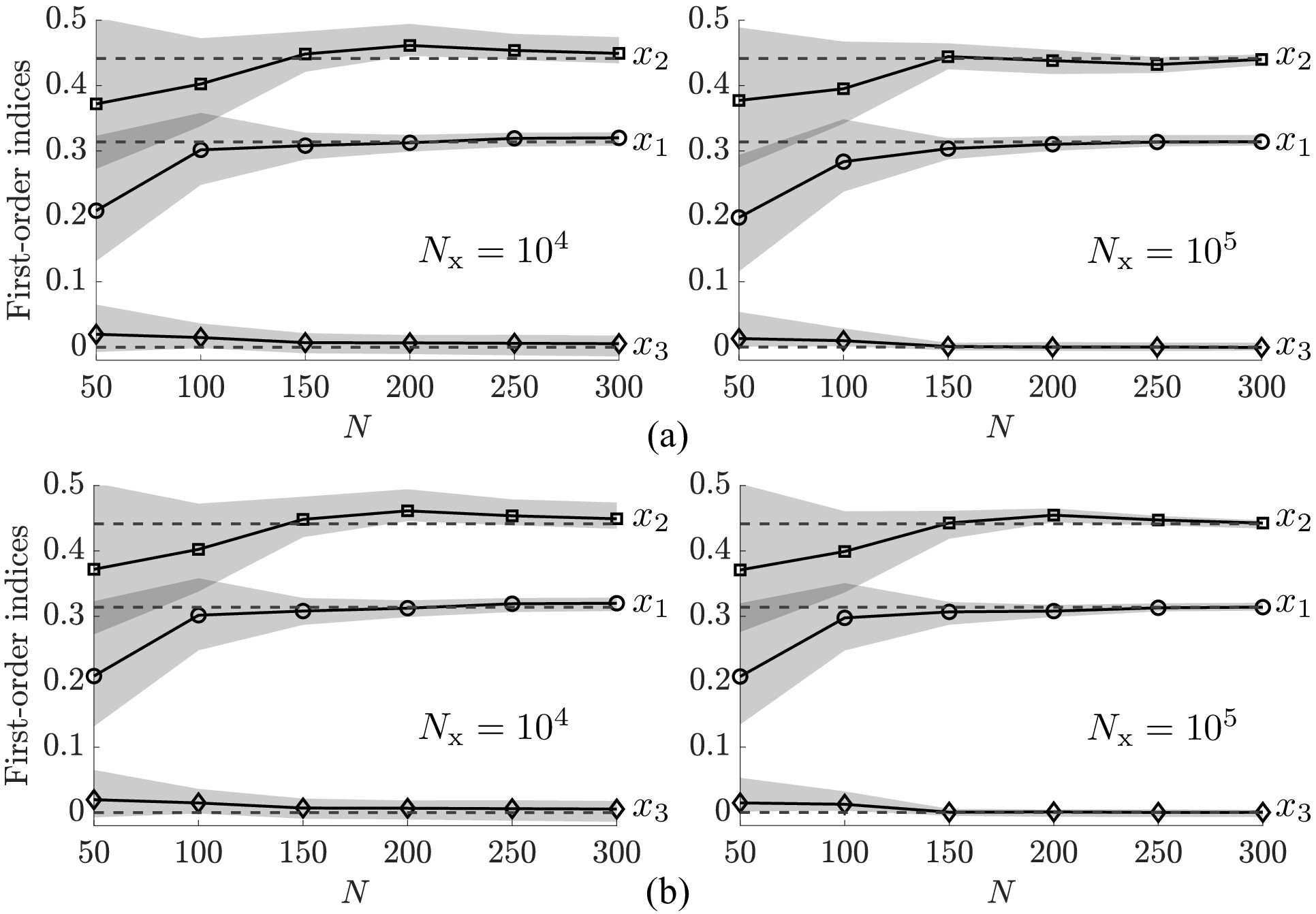}
	\caption{Approximate first-order indices for three variables of the Ishigami function using (a) $\texttt{GP-RFF}$ and (b) $\texttt{GP-PC}$ with different combinations of $N$ and $N_\text{x}$. Solid lines, shaded areas, and dashed lines represent median values, interquartile ranges, and exact values, respectively.}
    \label{fig:SAresultsIshigamifirst}
\end{figure*}

To show how the accuracy and uncertainty of the approximate sensitivity indices depend on the number of training points $N$ and the number of input variable samples $N_\text{x}$, we use all combinations of $N \in \{50,100,\dots,300\}$ and $N_\text{x} \in \{10^4,10^5\}$.
The effect of using random matrices ${\bf A}$ and ${\bf B}$ on the sensitivity analysis results is examined via ten pairs of ${\bf A}$ and ${\bf B}$.
For each trial defined by a combination of $N$ and $N_\text{x}$, we implement $\texttt{GP-GSA}$ from Line 2 ten times, each with a pair of ${\bf A}$ and ${\bf B}$. Other parameters include the number of RFFs $N_\phi = 2000$, the number of posterior sample functions $N_\text{s} = 200$, and the observation noise $\sigma_\text{n} = 10^{-4}$.

With the above setting, a trial provides $2000$ values for each sensitivity index, each corresponding to a pair of ${\bf A}$ and ${\bf B}$, and a posterior sample function. These values are used to compute the median and interquartile range for each sensitivity index. 

\begin{figure*}[t]
	\centering
	\includegraphics[width=\textwidth]{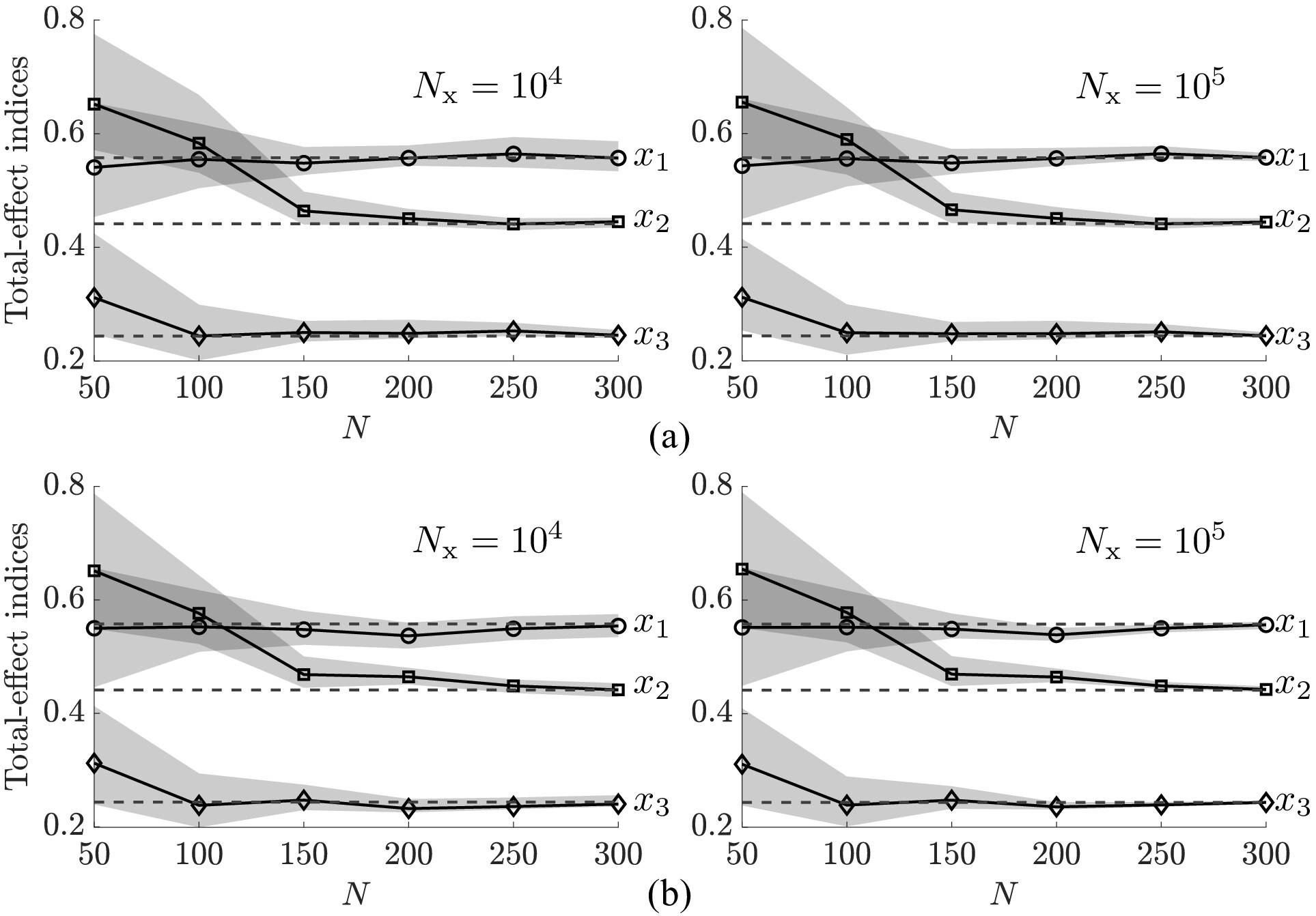}
	\caption{Approximate total-effect indices for three variables of the Ishigami function using (a) $\texttt{GP-RFF}$ and (b) $\texttt{GP-PC}$ with different combinations of $N$ and $N_\text{x}$. Solid lines, shaded areas, and dashed lines represent median values, interquartile ranges, and exact values, respectively.}
    \label{fig:SAresultsIshigamitotal}
\end{figure*}

\Cref{fig:SAresultsIshigamifirst,fig:SAresultsIshigamitotal} show the approximate first-order and total-effect indices for different combinations of $N$ and $N_\text{x}$, respectively.
When $N$ increases from $50$ to $300$, the medians of the approximate indices by both $\texttt{GP-RFF}$ and $\texttt{GP-PC}$ converge to the exact values.
In the right column of \cref{fig:SAresultsIshigamifirst,fig:SAresultsIshigamitotal}, the interquartile ranges of the approximate indices almost vanish for the combination of a sufficiently large value of $N$ and $N_\text{x} = 10^5$.
This is because a large number of observations guarantees a highly accurate GP model, while the number of sample points is large enough to prevent the pick-freeze problem.

\begin{figure}[t]
	\centering
	\includegraphics[width=0.5\textwidth]{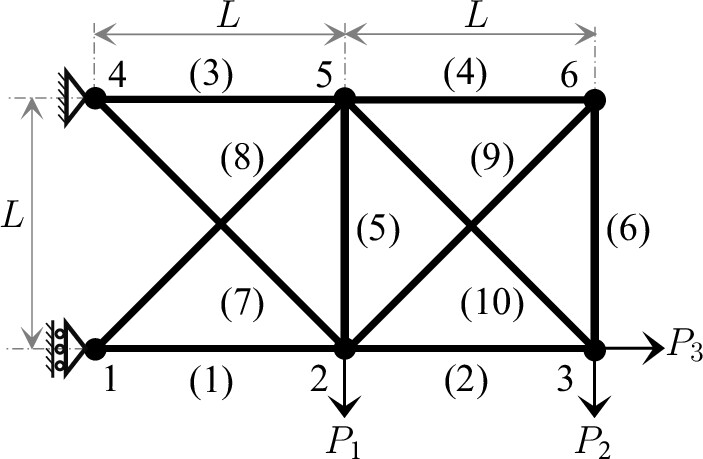}
	\caption{Ten-bar truss.}
    \label{fig:tenbartruss}
\end{figure}

\begin{table*}[t]
    \centering
    \caption{Probabilistic properties of random input variables for the ten-bar truss.}
    \begin{tabular}{cllcc}
        \toprule
       ID & Variable & Distribution &  Mean & Coefficient of variation\\
         & &  &  (or lower value) & (or upper value) \\
        \midrule
        1 & $P_1$ [kN] & Gaussian & 60 & 0.6\\
        2 & $P_2$ [kN] & Gaussian & 40 & 0.4\\
        3 & $P_3$ [kN] & Gaussian & 10 & 0.1\\
        4 & $E$ [GPa] & Gaussian & 200 & 0.2\\
        5 & $L$ [m] & Gaussian & 1 & 0.05\\
        6 & $A_1$ [$10^{-4} \mathrm{m}^2$] & Uniform & 6.5 & 14.5\\
        7 & $A_2$ [$10^{-4} \mathrm{m}^2$] & Uniform & 3.5 & 7.5\\
        8 & $A_3$ [$10^{-4} \mathrm{m}^2$] & Uniform & 10 & 18\\
        9 & $A_4$ [$10^{-4} \mathrm{m}^2$] & Uniform &  0.4 & 1.6\\
        10 & $A_5$ [$10^{-4} \mathrm{m}^2$] & Uniform &  0.4 & 1.6\\
        11 & $A_6$ [$10^{-4} \mathrm{m}^2$] & Uniform &  0.4 & 1.6\\
        12 & $A_7$ [$10^{-4} \mathrm{m}^2$] & Uniform &  3.5 & 7.5\\
        13 & $A_8$ [$10^{-4} \mathrm{m}^2$] & Uniform &  7 & 15\\
        14 & $A_9$ [$10^{-4} \mathrm{m}^2$] & Uniform &  0.4 & 1.6\\
        15 & $A_{10}$ [$10^{-4} \mathrm{m}^2$] & Uniform &  6.5 & 14.5\\
        \bottomrule
    \end{tabular}
    \label{table:1}
\end{table*}

\begin{figure*}[t]
	\centering
	\includegraphics[width=\textwidth]{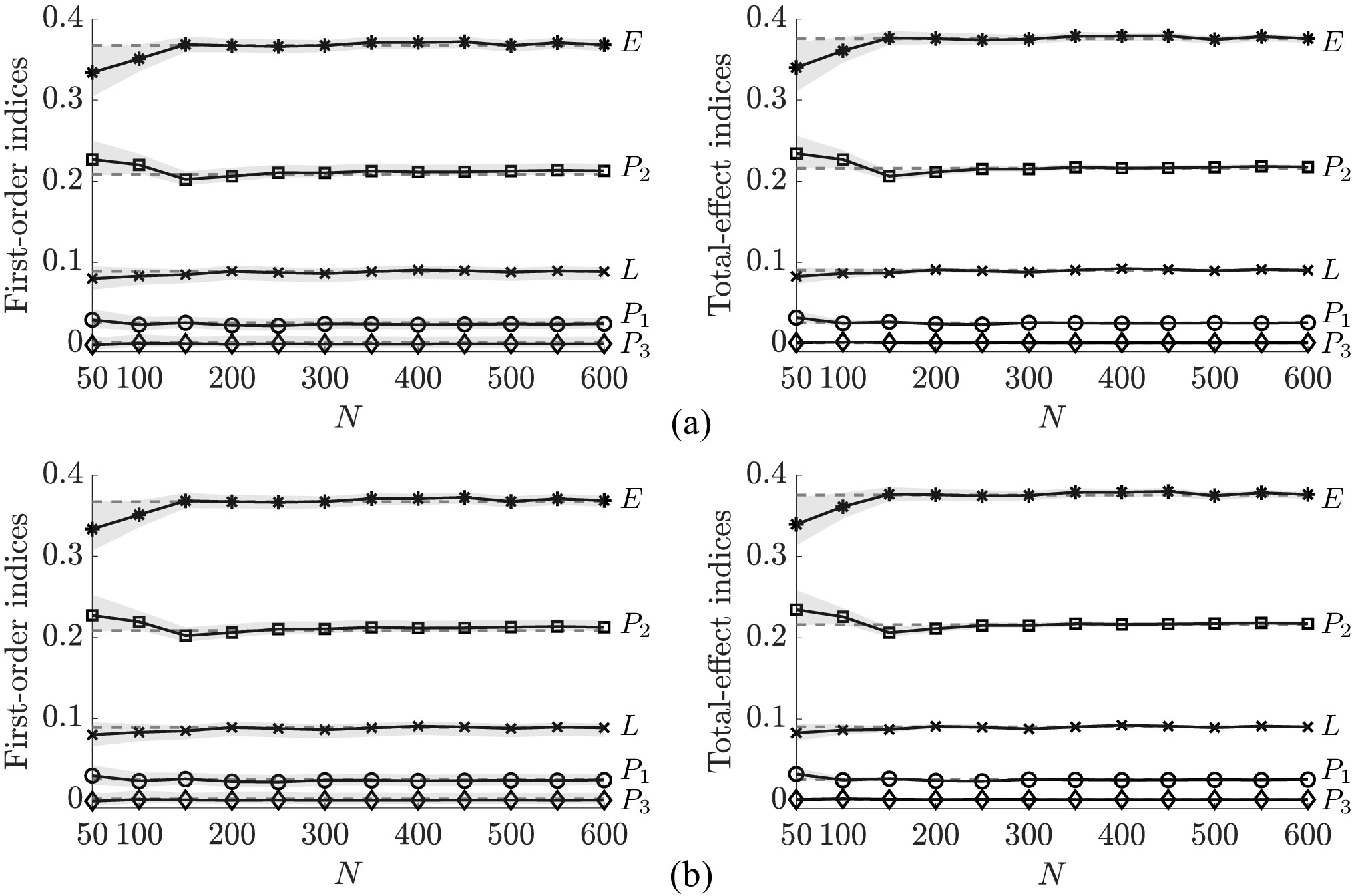}
	\caption{Approximate sensitivity indices for external loads $P_1$, $P_2$, and $P_3$, Young's modulus $E$, and geometric variable $L$ of the ten-bar truss using (a) $\texttt{GP-RFF}$ and (b) $\texttt{GP-PC}$ with different values of $N$. Solid lines, shaded areas, and dashed lines represent median values, interquartile ranges, and Monte Carlo estimates with direct use of the model, respectively.}
    \label{fig:SAresultsTruss1}
\end{figure*}

\begin{table*}[t]
\centering
\caption{Approximate sensitivity indices for the ten-bar truss using (a) $\texttt{GP-RFF}$, and (b) $\texttt{GP-PC}$, when the GP posterior is trained from 600 data points.}
\begin{tblr}{
  hline{1,3,18} = {-}{},
  hline{2} = {3-8}{},
}
ID & Variable &                       & $S_i$                 &                       &            & $S^T_i$                &  \\
 &          & $\texttt{GP-RFF}$,                   & $\texttt{GP-PC}$,                    & Baseline                   & $\texttt{GP-RFF}$,        & $\texttt{GP-PC}$,                    & Baseline     \\
1 & $P_1$      & 0.02418               & 0.02419               & 0.02537               & 0.02548    & 0.02551               & 0.02531 \\
2 & $P_2$      & 0.21286               & 0.21272               & 0.20862               & 0.21763    & 0.21751               & 0.21617 \\
3 & $P_3$      & 0.00030               & 0.00025               & 0.00126               & 0.00115    & 0.00117               & 0.00112 \\
4 & $E$        & 0.36825               & 0.36873               & 0.36740               & 0.37594    & 0.37634               & 0.37563 \\
5 & $L$        & 0.08862               & 0.08861               & 0.08911               & 0.09013    & 0.09015               & 0.09024  \\
6 & $A_1$      & 0.05902               & 0.05899               & 0.05921               & 0.06036    & 0.06037               & 0.06153  \\
7 & $A_2$      & 0.01348               & 0.01355               & 0.01602               & 0.01602    & 0.01610               & 0.01643  \\
8 & $A_3$      & 0.05562               & 0.05569               & 0.05352               & 0.05648    & 0.05659               & 0.05565 \\
9 & $A_4$      & 0.00134               & 0.00133               & 0.00001               & 0.00012    & 0.00012               & 0.00006 \\
10 & $A_5$      & 0.00153               & 0.00150               & 0.00291               & 0.00344    & 0.00340               & 0.00317 \\
11 & $A_6$      & 0.00123               & 0.00123               & 0.00005               & 0.00006    & 0.00006               & 0.00005 \\
12 & $A_7$      & 0.01673               & 0.01659               & 0.01834               & 0.01900    & 0.01882               & 0.01873 \\
13 & $A_8$      & 0.07344               & 0.07326               & 0.07359               & 0.07521    & 0.07499               & 0.07462 \\
14 & $A_9$      & 0.00102               & 0.00101               & 0.00023               & 0.00028    & 0.00028               & 0.00032\\
15 & $A_{10}$   & 0.06978               & 0.06978               & 0.06985               & 0.07192    & 0.07190               & 0.07156               
\end{tblr}
\label{table:2}
\end{table*}

\begin{figure*}[h!]
	\centering
	\includegraphics[width=\textwidth]{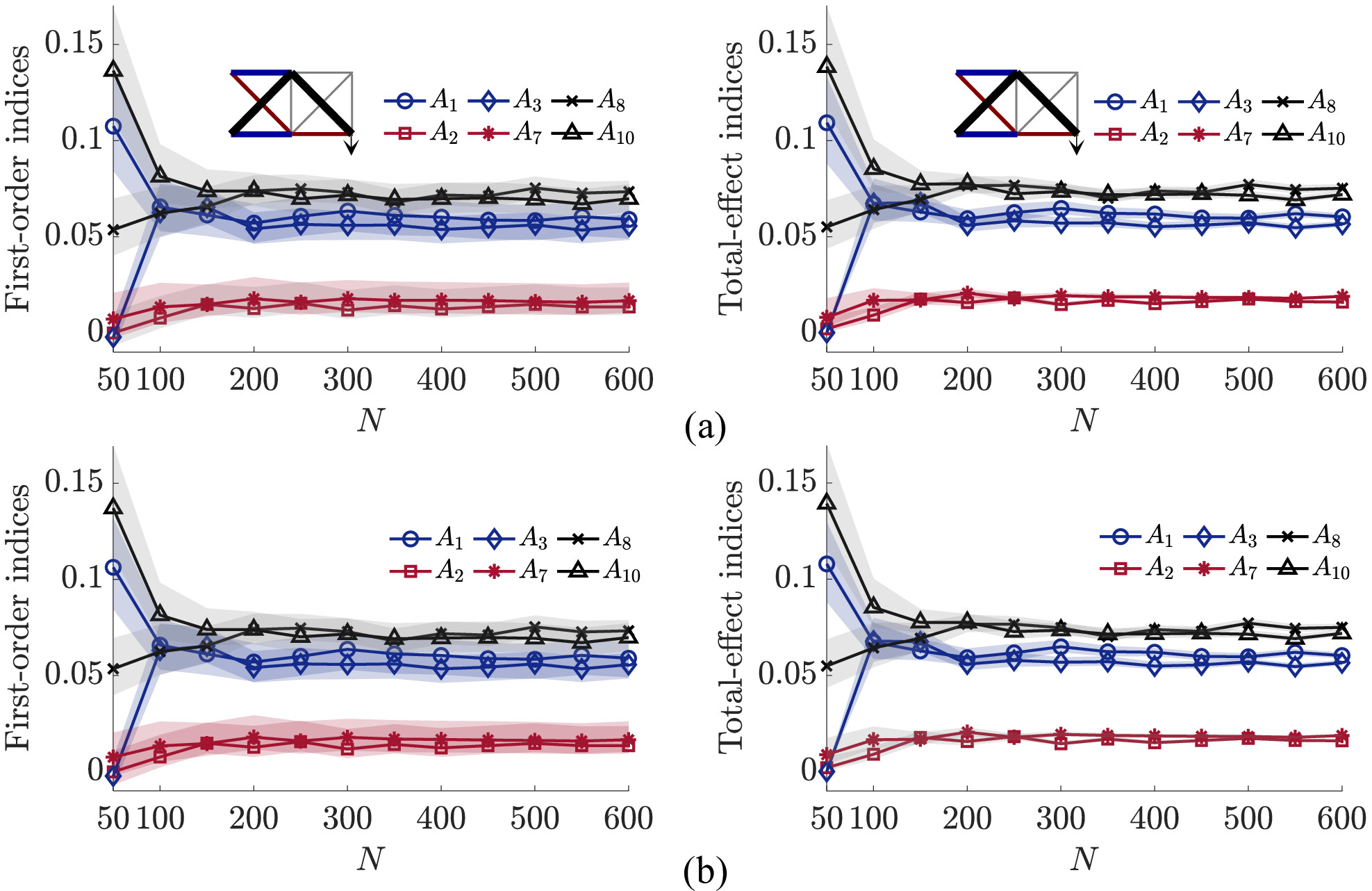}
	\caption{Approximate sensitivity indices for the cross-section areas of six important truss members (i.e., 1, 2, 3, 7, 8, and 10) of the ten-bar truss using (a) $\texttt{GP-RFF}$ and (b) $\texttt{GP-PC}$ with different numbers of training data points. Solid lines and shaded areas represent median values and interquartile ranges, respectively.}
    \label{fig:SAresultsTruss2}
\end{figure*}

\subsubsection{Ten-bar truss}
\label{sec812}

We compute the sensitivity indices for the ten-bar truss structure in \cref{fig:tenbartruss}.
The output of interest $c({\bf x})$ is the vertical displacement at node 3 of the truss.
There are 15 input variables for the truss, including external loads $P_1$, $P_2$, and $P_3$, Young's modulus $E$ of the truss material, the geometrical variable $L$, and the cross-sectional areas $A_1, \dots, A_{10}$ of ten truss members.
The probabilistic properties of these input variables are given in \Cref{table:1}.

To see how the accuracy of the GP model for $c({\bf x})$ influences the sensitivity analysis results, we consider 12 different values of the number of training points $N \in \{50,100,\dots,600\}$.
Other settings for \texttt{GP-GSA} include the number of input variable samples $N_\text{x} = 5 \times 10^4$, the number of RFFs $N_\phi = 2000$, the number of GP posterior sample functions drawn in each trial $N_\text{s} = 500$, and the observation noise $\sigma_\text{n} = 10^{-4}$.
Ten pairs of random matrices ${\bf A}$ and ${\bf B}$ are generated for each trial to see how they affect on the sensitivity analysis results.

To have a baseline for the sensitivity analysis results of the truss, we implement \Cref{alg:GP-GSA} with direct use of $c({\bf x})$ in Line 4, where $c({\bf x})$ is computed from a finite element code of bar elements. We call the resulting sensitivity indices the Monte Carlo estimates with direct use of the model. 

In structural design, we specify boundary conditions, material properties, and geometric properties of our structure before the profiles of its members are selected so that they meet the specified conditions and properties. Thus, we can classify the input variables of the truss into two groups, each corresponding to a distinct stage of the design process: group 1 consists of $P_1$, $P_2$, $P_3$, $E$, and $L$, and group 2 consists of the cross-sectional areas of the truss members.

\Cref{fig:SAresultsTruss1} shows the approximate sensitivity indices for the variables in group 1.
The approximations of each index approach the corresponding Monte Carlo estimate with direct use of the model when $N$ reaches 200.
Additionally, there is no interaction between the variables that influences the vertical displacement at node 3 of the truss as the first-order and total-effect indices for each variable are almost identical.

\Cref{table:2} lists the approximate sensitivity indices for $N=600$. The approximation results from $\texttt{GP-RFF}$ and $\texttt{GP-PC}$ are the same.
From these values, we can rank the most to the least important variables in group 1 for the vertical displacement at node 3 as follows: $E$, $P_2$, $L$, $P_1$, and $P_3$.

\Cref{fig:SAresultsTruss2} shows the approximate sensitivity indices for six important cross-sectional variables in group 2 (i.e., $A_1$, $A_2$, $A_3$, $A_7$, $A_8$, and $A_{10}$).
The sensitivity indices for the remaining four cross-sectional variables (i.e., $A_4$, $A_5$, $A_6$, and $A_9$) reach zero and are omitted from the plots (see \Cref{table:2} for their values).
The ranking of the cross-sectional variables in terms of their influence on the vertical displacement at node 3 is as follows: $(A_8,A_{10})$, $(A_1,A_3)$, $(A_2,A_7)$, and $(A_4,A_5,A_6,A_9)$.

\subsection{Single-objective optimization}
\label{sec82}

We now implement $\texttt{GP-TS-SO}$ (\Cref{alg:GP-TS-SO}) to solve challenging minimization problems for four benchmark functions, namely the $2$D Schwefel, $4$D Rosenbrock, $10$D Levy, and $16$D Ackley functions~\citep{Surjanovic2013}, and for the ten-bar truss in \cref{fig:tenbartruss} and the NACA 4-digit airfoil shape in \cref{fig:airfoilshape}.
The analytical expressions and true global minima of the benchmark functions are given in Appendix~\ref{AppC}.
The shape optimization problem for the airfoil is detailed in Appendix~\ref{AppD}.
The minimization problem formulated for the ten-bar truss is to minimize an objective function $c({\bf x})$ defined as the weighted-sum of the total cross-sectional area and the vertical displacement at node 3 \citep{Adebiyi2025tsroots}. Accordingly, 
\begin{equation}\label{trussSO}
    c({\bf x}) = w_1 \frac{c_1({\bf x})}{c_{1,\max}} + w_2 \frac{c_2({\bf x})}{c_{2,\max}}.
\end{equation}
Here, $c_1({\bf x}) = \sum_{1}^{10}A_i$, where $A_i \in [1,20] \times 10^{-4} \mathrm{m}^2$, $i \in \{1,\dots,10\}$, are cross-sectional areas of the truss members considered as design variables. $c_2({\bf x})$ represents the vertical displacement at node 3. $w_1 = 0.6$, $w_2 = 0.4$, $c_{1,\max} = 200 \times 10^{-4} \mathrm{m}^2$, and $c_{2,\max} = 3 \times 10^{-2} \mathrm{m}$. For this problem, we fix $P_1$, $P_2$, $P_3$, $E$, and $L$ at their mean values as given in \cref{table:1}. 

\begin{figure*}[t]
	\centering
	\includegraphics[width=\textwidth]{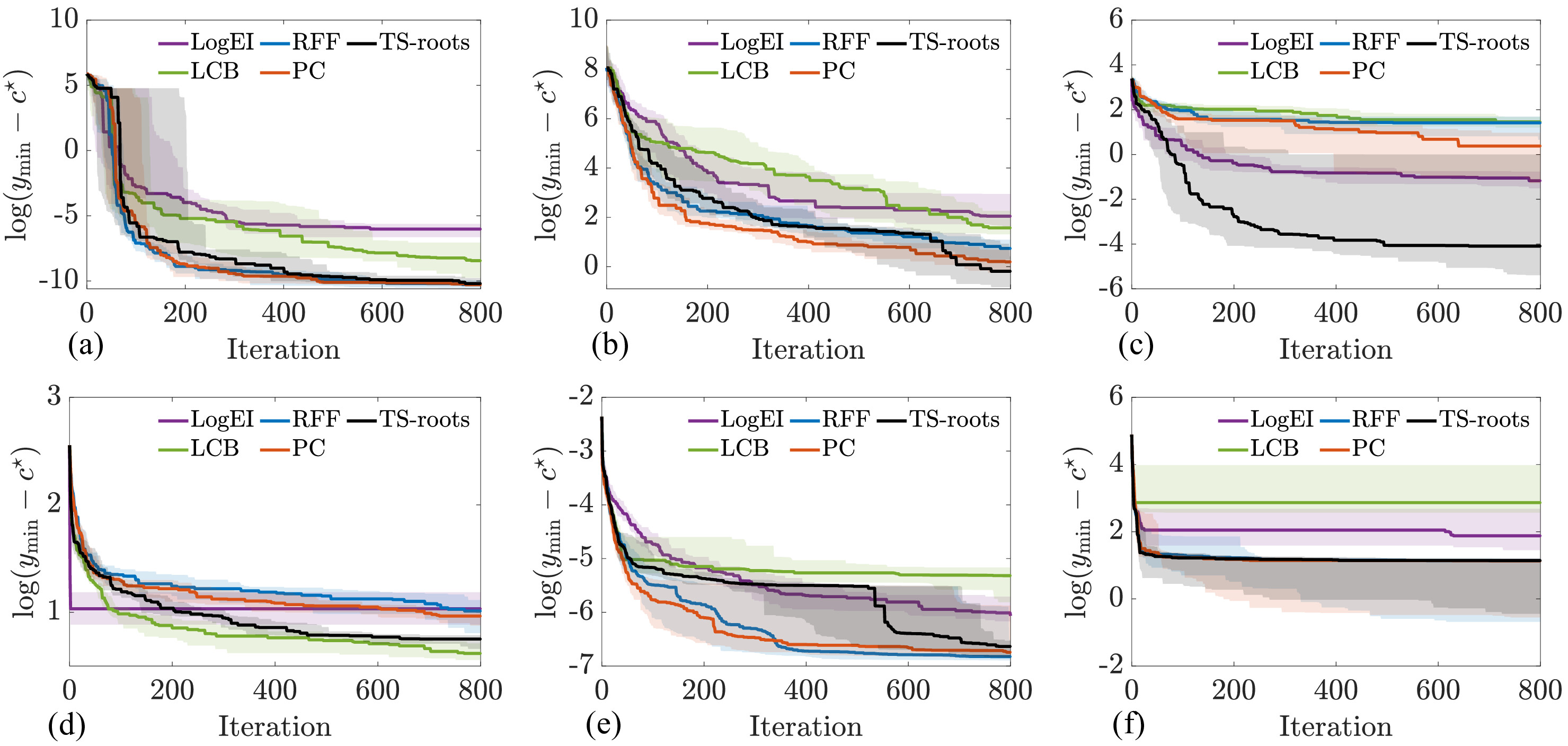}
	\caption{Medians and interquartile solution ranges from 20 trials of each BO method for (a) $2$D Schwefel function, (b) $4$D Rosenbrock function, (c) $10$D Levy function, (d) $16$D Ackley function, (e) ten-bar truss, and (f) airfoil.}
    \label{fig:SObenchmark}
\end{figure*}

We work in the normalized input variable space of $[0,1]^d$. The output observations are scaled using the $z-$score standardization.
The noise is $\sigma_\text{n} = 10^{-3}$ for the scaled output observations.

The number of initial observations and the number of RFFs are $N=10d$ and $N_\phi = 2000$, respectively.
The number of optimization iterations is $K=800$ for all problems.
The posterior function $\alpha({\bf x})$ is minimized in Line 5 of $\texttt{GP-TS-SO}$ using a multi-start gradient-based optimizer whose starting points are generated via the procedure used in BoTorch \citep{Balandat2020}: we first draw 512 Sobol' points in the input space and then select the best 30 candidates via Boltzmann sampling.
Both the function tolerance and the optimality tolerance for the multi-start gradient-based optimizer are set to $10^{-12}$.

We compare the optimization results by $\texttt{GP-TS-SO}$ with those by other well-known BO methods, including LogEI \citep{Ament2023} and LCB \citep{Srinivas2010}.
Hereafter, we use GP-TS-RFF and GP-TS-PC to represent $\texttt{GP-TS-SO}$ with $\texttt{GP-RFF}$ and $\texttt{GP-PC}$, respectively.
To ensure a fair comparison, 20 initial datasets are generated for each optimization problem using Latin hypercube sampling \citep{Owen1992}.
The multi-start gradient-based optimizer described above is also used to optimize LogEI and LCB acquisition functions.
In each optimization iteration, we record the best-found value of $\log(y_{\min}-c^\star)$ and the corresponding input variable vector, where $y_{\min}$ is the best observation of the objective function $c({\bf x})$, $c^\star$ for the benchmark functions is their true minimum value, and $c^\star$ for the ten-bar truss and airfoil is a value smaller than the best objective function value we observed from all trials.

To show how accurate optimization of GP-TS acquisition functions affects the performance of BO, especially in high-dimensional settings, we compare the optimization performance of three GP-TS variants, including GP-TS-RFF, GP-TS-PC, and TS-roots \citep{Adebiyi2025tsroots}. TS-roots enhances the gradient-based multistart optimization of \texttt{GP-PC} by designing a small, effective set of starting points consisting of exploration and exploitation points.
To ensure a fair comparison, the number of exploration and exploitation points is set at 20 and 10, respectively, resulting in a total of 30 starting points for TS-roots. 

\Cref{fig:SObenchmark} shows the medians and interquartile ranges of solutions obtained from 20 independent trials of each BO method.
At least one of the GP-TS variants (GP-TS-RFF, GP-TS-PC, or TS-roots) provides a good solution to each problem. LogEI performs poorly on most problems, with the exception of the 10D Levy function.
LCB performs well on the 16D Ackley function but degrades on the remaining problems.
Notably, TS-roots is competitive with GP-TS-RFF and GP-TS-PC on the 2D Schwefel, 4D Rosenbrock, ten-bar truss, and airfoil problems, while achieving better solutions on the 10D Levy and 16D Ackley functions. These results indicate that accurately optimizing GP-TS acquisition functions, especially in high-dimensional settings, can substantially improve the overall performance of BO.
In most cases, GP-TS-RFF and GP-TS-PC perform better in later stages. This is because $\texttt{GP-TS-SO}$ delays the reward as it prioritizes exploration in the initial stages.
\subsection{Multi-objective optimization}
\label{sec83}

Finally, we implement $\texttt{GP-TS-MO}$ (\Cref{alg:GP-TS-MO}) to solve four benchmark MO problems: KNO1, VLMOP2, VLMOP3, and DTLZ2a \citep[see e.g.,][]{vanVeldhuizen1999,Knowles2006}, and a bi-objective minimization problem formulated for the ten-bar truss in \cref{fig:tenbartruss}.
KNO1 and VLMOP2 are bi-objective minimization problems of two input variables.
VLMOP3 and DTLZ2a are tri-objective minimization problems. The former has two input variables and the latter has eight input variables.
The formulation and input variable domain of each MO benchmark problem are provided in Appendix~\ref{AppE}.
The bi-objective minimization problem for the ten-bar truss is to simultaneously minimize the total cross-sectional area $c_1({\bf x})$ and the vertical displacement $c_2({\bf x})$ at node 3, which are defined in \cref{trussSO}.

Like what we did in \Cref{sec82}, we implement $\texttt{GP-TS-MO}$ in the normalized input variable space of $[0,1]^d$ and the $z-$score scaled output space.
The noise is set as $\sigma_\text{n} = 10^{-3}$ for the scaled output observations.

For all problems, we use NSGA-II \citep{Deb2002} as the solver underlying function \texttt{pareto\_sol} in Line 5 of $\texttt{GP-TS-MO}$.
This solver is characterized by a population size of $500$, a number of generations of $100$, a crossover fraction of $65\%$, a constraint tolerance of $10^{-12}$, and a function tolerance of $10^{-12}$.
The parameters for $\texttt{GP-TS-MO}$ are $N=100$, $N_\phi = 2000$, and $K=500$.

We implement 20 trials for each of the $\texttt{GP-TS-MO}$ methods (i.e., $\texttt{GP-RFF}$ and $\texttt{GP-PC}$). These trials differ in their initial datasets.
For comparison, we perform 20 trials of NSGA-II for each problem. Each of these NSGA-II trials uses a population size that is equal to the number of initial data points $N$ in each $\texttt{GP-TS-MO}$ method, and is run for a number of generations so that the total number of objective function evaluations is equal to that required by each $\texttt{GP-TS-MO}$ method.
Baseline solutions for each problem, denoted as $(\mathcal{P}_x^\star,\mathcal{P}_y^\star)$, are obtained from another NSGA-II run with a substantially larger population size of $5000$ and $1000$ generations.

\begin{figure*}[t]
	\centering
	\includegraphics[width=0.99\textwidth]{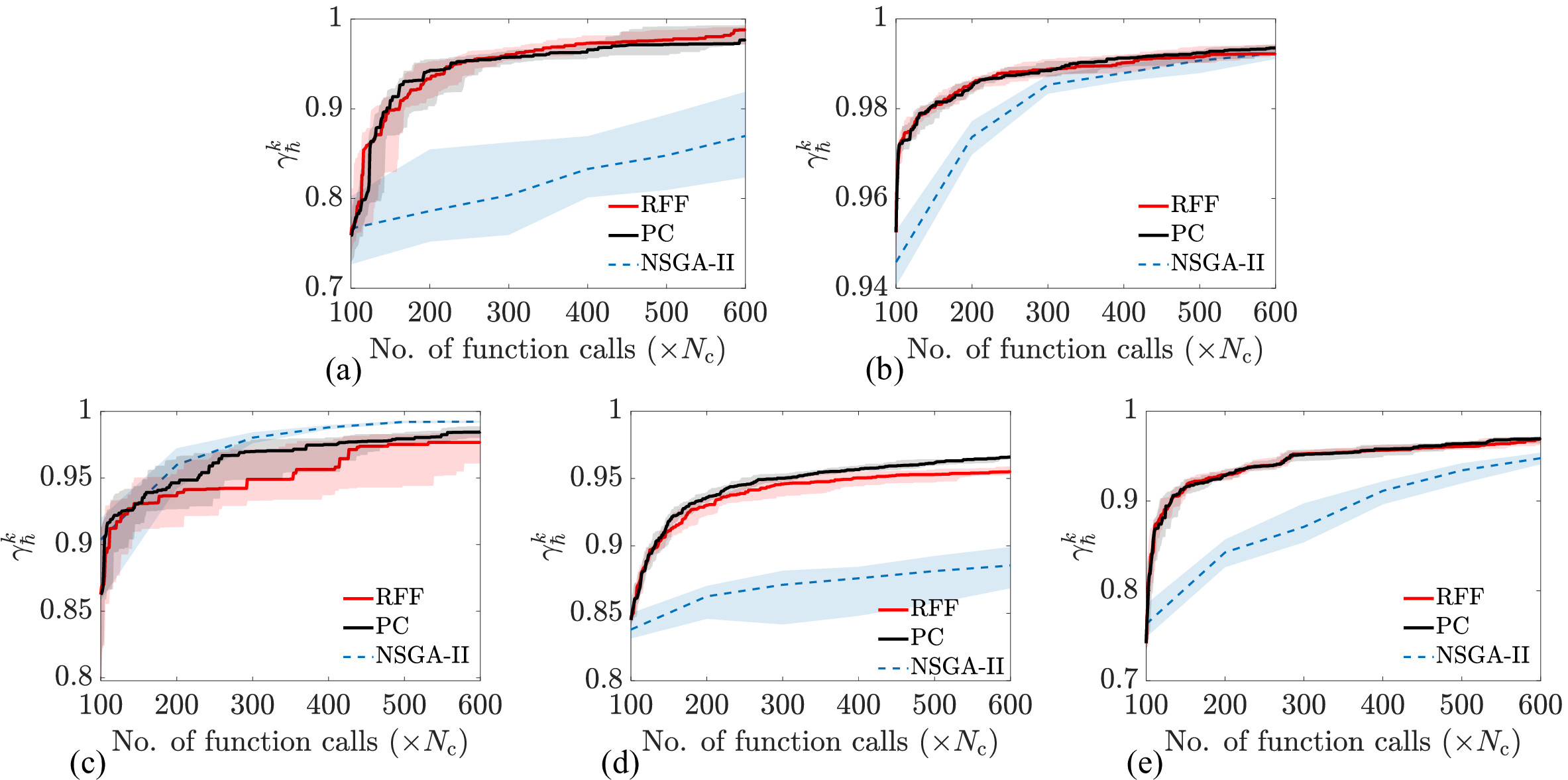}
	\caption{Medians and interquartiles of hypervolume ratio values from 20 trials of each GP-TS method for (a) KNO1, (b) VLMOP2, (c) VLMOP3, (d) DTLZ2a, and (e) ten-bar truss.}
	\label{fig:mohvi}
\end{figure*}

Let ${\bf r} = [25,25]^\intercal$, $[2,2]^\intercal$, $[10, 18, 0.2]^\intercal$, $[2, 2, 2]^\intercal$, and $[200, 2.5]^\intercal$ represent reference points defined for the KNO1, VLMOP2, VLMOP3, DTLZ2a, and ten-bar truss problems, respectively.
These reference points are used to assess the improvement in the hypervolume indicator for each problem after we obtain the optimization results, and they are independent of these results. 
Note that it is not necessary to fix the reference points at the aforementioned values, as we are free to define other reference points for each problem as long as each of their component values is greater than the corresponding value in the given reference points.
To show how each of the $\texttt{GP-TS-MO}$ methods improves the hypervolume indicator during the optimization process, we compute $\hbar\left( \mathcal{P}_y^k,{\bf r} \right)$ for the approximate Pareto front $\mathcal{P}_y^k$ in each optimization iteration, and define $\gamma_\hbar^k = \hbar\left( \mathcal{P}_y^k,{\bf r} \right)/\hbar\left( \mathcal{P}_y^\star,{\bf r} \right)$ as the hypervolume ratio. If $\gamma_\hbar^k$ approaches $1$, then the approximate Pareto front $\mathcal{P}_y^k$ well captures the baseline Pareto front $\mathcal{P}_y^\star$. We also expect that each of the $\texttt{GP-TS-MO}$ methods continually improves $\gamma_\hbar^k$ during the optimization process.

\begin{figure*}[t]
	\centering
	\includegraphics[width=0.99\textwidth]{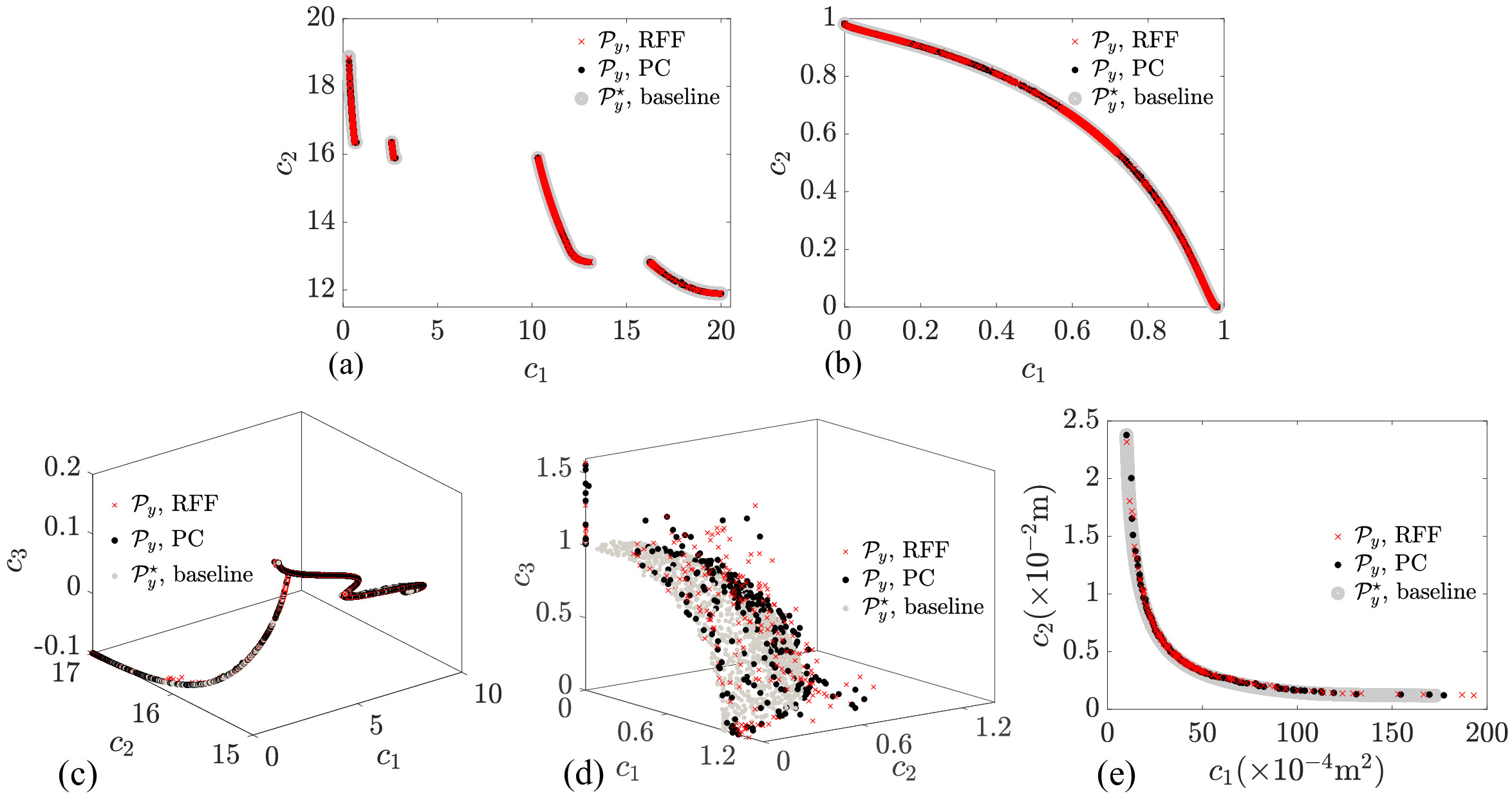}
	\caption{Approximate Pareto front from 20 trials of each GP-TS method for (a) KNO1, (b) VLMOP2, (c) VLMOP3, (d) DTLZ2a, and (e) ten-bar truss.}
	\label{fig:mopareto}
\end{figure*}

With an abuse of abbreviations, we use GP-TS-RFF and GP-TS-PC to represent $\texttt{GP-TS-MO}$ with $\texttt{GP-RFF}$ and $\texttt{GP-PC}$, respectively. \Cref{fig:mohvi} shows the improvement of the hypervolume indicator during the optimization by each of the $\texttt{GP-TS-MO}$ methods.
Both GP-TS-RFF and GP-TS-PC show considerable improvements in the hypervolume indicator for each problem.
The resulting approximate Pareto fronts are better or on par with those of NSGA-II when using the same number of objective function evaluations.

\Cref{fig:mopareto} compares the combined Pareto front of the approximate Pareto fronts by 20 trials of each $\texttt{GP-TS-MO}$ method with the baseline Pareto front $\mathcal{P}_y^\star$.
While 20 trials of each $\texttt{GP-TS-MO}$ method require $20 \times 600 \times N_\text{c}$ objective function evaluations, they provide high-quality approximate Pareto fronts that well capture the baseline Pareto front obtained from $5000 \times 1000 \times N_\text{c}$ objective function evaluations.

\section{Discussion}
\label{sec9}
In the previous sections, we presented various methods for drawing posterior samples from GPs, with a focus on RFF and PC methods.
We showed how the generated samples can be used to solve GSA and optimization problems, and provided numerical examples illustrating their effectiveness in these tasks.
In this section, we summarize and discuss some key takeaways obtained from the previous sections.

For approximating covariance functions and GP prior samples, the OE and Hilbert space approximation show excellent performance in low-dimensional settings. Unfortunately, they are subject to the curse of dimensionality.
In higher-dimensional settings, the QMC and RFF methods are more suitable.
While the QMC method has faster convergence than the RFF method, it has a slightly higher sampling overhead for generating low-discrepancy sequences, and its advantage tends to degrade when the dimension exceeds ten \citep{HuangZ2024qmc}.

Compared to the exhaustive sampling from the function-space view of GPs, the RFF method is computationally more efficient when the number of query points increases.
However, its extrapolation performance degrades when the number of training points grows within the interpolation region.
The PC method inherits the accuracy of exhaustive sampling and improves upon the computational efficiency of the RFF method.

In the implementation of the RFF method and PC with RFF priors, careful attention should be given to selecting the number of features and setting an observation noise level.
Based on the results presented and our experience, using a few thousand RFFs to approximate SE and Matérn covariance functions can provide a satisfactory balance between approximation accuracy and computational efficiency.
Additionally, for the use of these sampling methods in GSA and optimization, we recommend setting the noise level at a small value, such as $\sigma_\text{n} = 10^{-4}$ or $10^{-3}$, if one wishes to assume noiseless observations.
Otherwise, the noise level can be specified based on prior knowledge or optimized automatically during the training process.
For training stability, we recommend normalizing the input variable space to the unit hypercube $[0,1]^d$ and scaling the output observations using the $z$-score standardization.
It is worth mentioning that the current implementation of the sampling methods assumes homoscedastic noise, which means all observations share the same variance. Extending these methods to accommodate heteroscedastic noise that varies according to observation regions is still an open problem.
This is important because our knowledge often varies in different domains of a problem.

The accuracy of GPs and the number of random input variables are two main factors that influence the uncertainty in the GP sample-based global sensitivity indices.
The unfavorable impact of the model accuracy can be mitigated by simply increasing the number of training data points.
When GP posterior samples are available, increasing the number of random input variables to a sufficiently large value can mitigate the pick-freeze problem while introducing a modest increase in computational cost.

For optimization, GP-TS via posterior sample functions, as a randomized algorithm, can lead to delayed rewards.
This is because GP-TS tends to favor exploration in the early stages of the optimization process when observations are limited.
It is also important to have a robust global algorithm to optimize GP-TS acquisition functions \citep{Adebiyi2025tsroots}, because its convergence guarantee relies on the premise that the acquisition functions are globally optimized \citep{Wilson2018}.

\section{Other considerations}
\label{sec10}

\subsection{Selection of covariance function}
\label{sec101}
In GPs, the covariance function encodes key structural assumptions about the latent function such as smoothness, additivity, or stationarity. The choice of covariance function therefore affects both predictive accuracy and uncertainty calibration, which is critical for downstream tasks like GSA and optimization.
Selecting a covariance function is a special case of model selection whose objective is to choose a model structure $\mathcal{M}$, determined by a particular combination of observation model as in \cref{eqn:marginallikelihood}, prior mean function, and prior covariance function, such that 
the chosen $\mathcal{M}$ well explains the observed data $\mathcal{D}$. When the observation model and prior mean objective are specified, model selection reduces to choosing a covariance function and its hyperparameters $\boldsymbol{\phi}_\kappa$.

Given a set $\mathcal{M}$ of candidate covariance functions, we can perform Bayesian inference over 
the hyperparameters $\boldsymbol{\phi}_\kappa$ for each candidate using the data $\mathcal{D}$. Let $(\boldsymbol{\phi}_\kappa,\mathcal{M})$ index the compound space of candidates. The size of $\boldsymbol{\phi}_\kappa$ may differ across candidates. We define a model posterior over this space as $p(\boldsymbol{\phi}_\kappa, \mathcal{M} | \mathcal{D}) = p(\mathcal{M} | \mathcal{D}) \, p(\boldsymbol{\phi}_\kappa| \mathcal{D}, \mathcal{M})$, where $p(\boldsymbol{\phi}_\kappa| \mathcal{D}, \mathcal{M})$ is the hyperparameter posterior and $p(\mathcal{M} | \mathcal{D})$ is the model posterior. The latter can be further written as $p(\mathcal{M} | \mathcal{D}) \propto p(\mathcal{M}) \, p(\mathbf{y} | \mathbf{X}, \mathcal{M})$, with the model evidence
$p(\mathbf{y} | \mathbf{X}, \mathcal{M}) = \int p(\mathbf{y} | \mathbf{X}, \boldsymbol{\phi}_\kappa, \mathcal{M}) \, p(\boldsymbol{\phi}_\kappa|\mathcal{M})\, \mathrm{d}\boldsymbol{\phi}_\kappa$.
If the candidate prior $p(\mathcal{M})$ is flat, then we can select a covariance function from $\mathcal{M}$ that maximizes $p(\mathbf{y} | \mathbf{X}, \mathcal{M})$ \citep{Malkomes2016}. In practice, the model evidence is generally intractable, so we approximate it via a second-order Laplace approximation of $\log p(\boldsymbol{\phi}_\kappa | \mathcal{D}, \mathcal{M})$ \citep{Malkomes2016}. This leads to an efficient workflow: compute the MAP estimate ${\boldsymbol{\phi}}^\star_{\kappa}$ for each candidate, approximate the corresponding evidence model, and select the model that is best supported by the data.
Finally, because $\mathcal{D}$ evolves in many downstream settings like BO, the selection of the best covariance function should be revisited as a new observation is revealed.

\subsection{Numerical stability of RFF and PC methods}
\label{sec102}
The stability of the RFF method mainly concerns the ill-posedness of the random feature matrix $\boldsymbol{\Phi}$ in \cref{eqn:postweightmeancov}. In our examples, $\boldsymbol{\Phi}$ is well-conditioned when $N_\phi \gg N$, where $N_\phi$ is the number of features and $N$ is the number of data points. \cite{ChenZ2024} showed that $\boldsymbol{\Phi}$ is well-conditioned when the ratio $N_\phi/N$ scales like $\log(N)$ and becomes ill-conditioned in the interpolation regime when $N_\phi=N$. Meanwhile, the stability of the PC method concerns the ill-conditioning of covariance matrix $\mathbf{C}$ (see \cref{eqn:pathwise}). By using a small jitter $\sigma_\text{n} = 10^{-4}$ or $10^{-3}$, we have not observed the numerical instability in our examples. Another strategy that can improve the numerical stability and avoid direct inversion of $\mathbf{C}$ is to use an iterative solver for solving a system of linear equations. This solver can be preconditioned conjugate gradient \citep{Gardner2018} or stochastic gradient descent without preconditioning \citep{LinJA2023}. 

\subsection{Handling discrete input variables and design constraints}
\label{sec103}
We may extend the sampling methods to handle categorical or integer-valued input variables or to enforce design constraints on posterior samples.
To handle categorical or integer-valued variables, a simple approach is to learn a continuous latent representation of the original input space \citep{Gomez-Bombarelli2018,Oune2021}, and then generate sample paths from the GP trained in that latent space.
Another approach is to transform the input variables before computing the covariance function so that the modeled function can obtain constant values in specific regions of the input variable space, for example, those with the same one-hot encoding value of categorical variables or those with the same integer value of integer-valued variables \citep{GarridoMerchan2020}.
To impose global inequality constraints (e.g.,  boundedness, monotonicity, or convexity), an approach is to place a multivariate Gaussian prior on a class of spline functions, where the constraints are incorporated through constraints on the coefficients of the spline functions \citep{Maatouk2017}. It is also possible to enforce bound constraints with high probability by constraining the optimization of the GP hyperparameters \citep{Pensoneault2020}.

\section{Conclusion}
\label{sec11}

We presented the formulation and implementation of RFF and PC methods to generate GP posterior samples.
We also briefly described alternative approaches for generating GP prior samples that are useful for the PC method in lower-dimensional settings.
Importantly, we detailed how the generated samples can be used in GSA, SO optimization, and MO optimization.
Through a series of numerical experiments, we showed that the presented sampling methods consistently produced reliable global sensitivity indices and higher-quality optimization solutions. Below, we suggest several promising research directions.

For practical GSA in engineering applications, it is important to extend the presented sampling methods to support importance measures that consider dependent and possible data.
For optimization, the presented sampling methods are not suitable for problems with multiple correlated objective functions.
 
Although not discussed in this paper, the presented sampling methods are well-suited for parallelization, which can significantly accelerate both GSA and optimization tasks.
Extensions of these sampling methods to problems with  multiple fidelities \citep{Kennedy2000,Do2025mfbo} may be another promising direction.
The effects of misspecified prior and likelihood models on the prediction performance of GP posterior samples, and consequently on the results of GSA and optimization, and how to manage them are also important research topics.

\section*{Acknowledgments}
The authors acknowledge the support of the Research Computing Data Core at the University of Houston, USA for assistance with the calculations carried out in this work.
\section*{Statements and Declarations}
\textbf{Conflict of interest.}
The authors declare that they have no conflict of interest.

\textbf{Funding.}
This work was funded by the University of Houston through the SEED program no. 000189862.

\textbf{Data Availability.}
Not applicable.

\textbf{Replication of results.}
The random Fourier feature and pathwise conditioning methods are transparent.
The code implementing them is available at \url{https://github.com/UQUH/GPSampling}.
All necessary information has been provided in the paper to enable replication of results for the global sensitivity analysis of the Ishigami function in \Cref{sec811} and the benchmark optimization problems in \Cref{sec82,sec83}.
The numerical solver for the ten-bar truss in \Cref{sec812,sec82} is available from the first author upon request.

\textbf{Ethics approval and Consent to participate.}
Not applicable.

\textbf{Author contributions.}
Bach Do: Conceptualization, Investigation, Software, Visualization, Writing--original draft, Review \& editing.
Nafeezat A. Ajenifuja: Investigation, Validation, Writing--review \& editing.
Taiwo A. Adebiyi: Investigation, Software, Data curation, Writing--review \& editing.
Ruda Zhang: Supervision, Funding acquisition, Writing--original draft, Review \& editing.

\newpage
\appendix
\section{Geometric explanation of conditional means and variances from bivariate Gaussian distributions}\label{AppA}
\setcounter{equation}{0} %
\renewcommand{\theequation}{A.\arabic{equation}} %
\setcounter{figure}{0} %
\renewcommand{\thefigure}{A.\arabic{figure}}

To provide an intuitive understanding of how to derive the conditional mean and conditional variance in \cref{eqn:conditionalmeancovariance}, we present a geometric explanation for bivariate Gaussian distributions.
Let ${\bf f}- {\bf m} = [f_1 - m_1, f_2 - m_2]^\intercal \sim \mathcal{N}\left( {\bf 0}, \boldsymbol{\Sigma}\right)$ represent a bivariate joint Gaussian distribution of $f_1-m_1$ and $f_2-m_2$, where
\begin{equation} \label{eqn:bivariatecovariance}
    \boldsymbol{\Sigma} = \begin{bmatrix}
    \Sigma_{11} &  \Sigma_{12}\\
    \Sigma_{21} & \Sigma_{22}
    \end{bmatrix} = 
    \begin{bmatrix}
    \sigma_1^2 &  \rho \sigma_1 \sigma_2\\
    \rho \sigma_1 \sigma_2 & \sigma_2^2
    \end{bmatrix}, 
\end{equation}
with $\left| \rho \right| \leqslant 1$, $\sigma_1>0$, and $\sigma_2>0$.

\begin{figure*}[t]
	\centering
	\includegraphics[width=\textwidth]{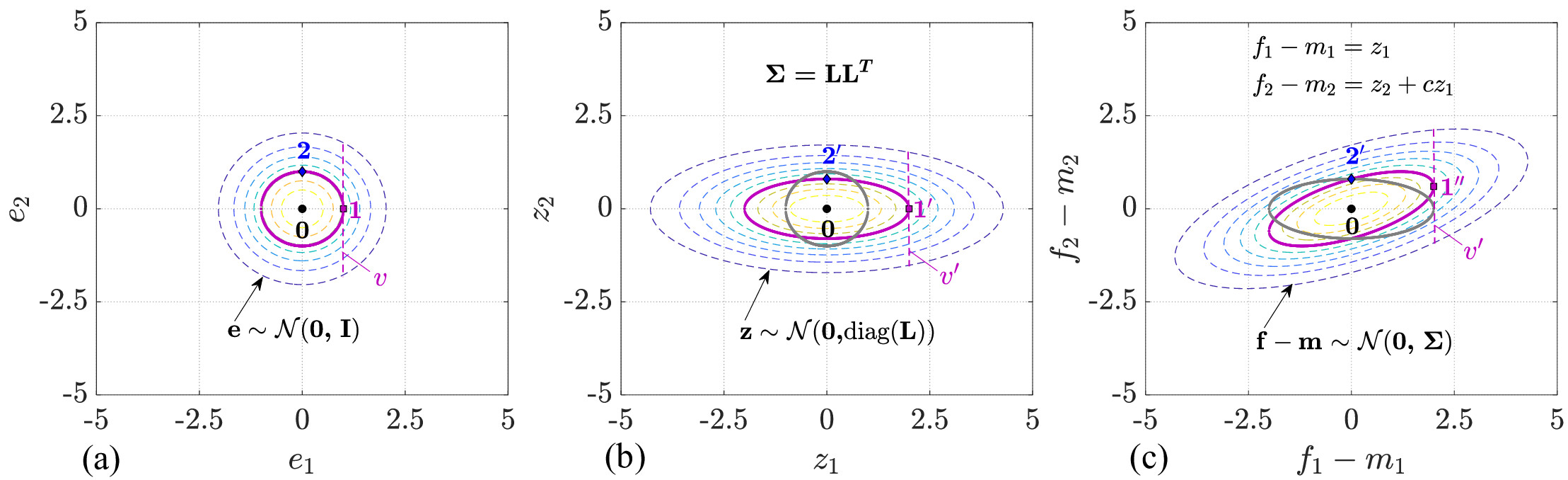}
	\caption{A geometric way to derive a conditional Gaussian distribution of $f_2-m_2|f_1-m_1$ (indicated by line segment $v'$ in panel (c)) from the corresponding joint Gaussian distribution ${\bf f}- {\bf m}\sim \mathcal{N}\left( {\bf 0}, \boldsymbol{\Sigma}\right)$. (a) Circular contours of $\mathcal{N}\left( {\bf 0}, {\bf I}\right)$, two reference points ${\bf 1}$ and ${\bf 2}$ on a contour, and the length of line segment $v$ proportional to the standard deviation of $e_2|e_1 \sim \mathcal{N}\left( 0, 1\right)$. Transformation from $\mathcal{N}\left( {\bf 0}, {\bf I}\right)$ to $\mathcal{N}\left( {\bf 0}, \boldsymbol{\Sigma}\right)$ can be decomposed into two steps. The first step (b) defines ${\bf z}=\diag({\bf L}) {\bf e}$ with ${\bf L}{\bf L}^\intercal = \boldsymbol{\Sigma}$, scaling the vertical and horizontal axes of the circular contours in (a) such that $z_1 = L_{11} e_1$ and $z_2 = L_{22} e_2$, where $L_{11}$ and $L_{22}$ are the $(1,1)$ and $(2,2)$th entries of ${\bf L}$, respectively. This step results in $|v'| = L_{22} |v|$ and $z_2|z_1 \sim \mathcal{N}\left( 0, L_{22}^2\right)$. The second step (c) lifts any points $(z_1,z_2)$ in (b) by an amount proportional to $z_1$, thereby transforming $(z_1,z_2)$ into $(z_1,z_2+cz_1)=(f_1-m_1,f_2-m_2)$, where $c$ can be computed directly from ${\bf L}$. This step not only shifts the mean of the conditional Gaussian distribution from (b), resulting in the target conditional mean $\mathbb{E}[f_2-m_2|f_1-m_1] = cz_1 = c(f_1-m_1)$, but also preserves the variance of the conditional Gaussian distribution as $\mathbb{V}[f_2-m_2|f_1-m_1] = \mathbb{V}[z_2|z_1] = L_{22}^2$.}
    \label{fig:ConditionalGaussian}
\end{figure*}

\Cref{fig:ConditionalGaussian} shows how we can derive the conditional Gaussian of $f_2-m_2|f_1-m_1$ from ${\bf f}- {\bf m} \sim \mathcal{N}\left( {\bf 0}, \boldsymbol{\Sigma}\right)$.
The idea is to compute the mean and variance of the conditional Gaussian distribution of $f_2-m_2|f_1-m_1$, indicated by the vertical-dashed line segment $v'$ in \cref{fig:ConditionalGaussian}(c), by transforming the conditional standard Gaussian distribution $e_2|e_1 \sim \mathcal{N}(0,1)$, indicated by the vertical-dashed line segment $v$ in \cref{fig:ConditionalGaussian}(a).
To deploy this idea, we first compute matrix ${\bf L}$ from the Cholesky decomposition of $\boldsymbol{\Sigma}$ in \cref{eqn:bivariatecovariance}. We have
\begin{equation}
    {\bf L} = \begin{bmatrix}
    \sigma_1 &  0\\
    \rho \sigma_2 & \sigma_2 \sqrt{1-\rho^2}
    \end{bmatrix}.
\end{equation}
We then define ${\bf f}- {\bf m} = {\bf L} {\bf e}$ to transform ${\bf e} \sim \mathcal{N}\left( {\bf 0},{\bf I} \right)$ into ${\bf f}- {\bf m}  \sim \mathcal{N}\left( {\bf 0}, \boldsymbol{\Sigma}\right)$, see \cref{fig:GaussianSampling}(b). This transformation reads
\begin{equation} \label{eqn:ConditionalGaussian}
        {\bf f}-{\bf m}  = 
        \begin{bmatrix}
            \sigma_1 &  0\\
            \rho \sigma_2 & \sigma_2 \sqrt{1-\rho^2}
        \end{bmatrix}
        \begin{bmatrix} 
            e_1 \\
            e_2
        \end{bmatrix} 
         = 
        \begin{bmatrix}
            \sigma_1 e_1 \\
            \sigma_2 \sqrt{1-\rho^2} e_2
        \end{bmatrix} + 
        \begin{bmatrix}
            0 \\
            \rho \sigma_2 e_1
        \end{bmatrix}.
\end{equation}

Each term of the right-hand side of \cref{eqn:ConditionalGaussian} corresponds to an affine transformation. The first term transforms ${\bf e} \sim \mathcal{N}({\bf 0},{\bf I})$ into ${\bf z} = [z_1, z_2]^\intercal = \diag({\bf L}) {\bf e} \sim \mathcal{N}({\bf 0}, \diag\left( \boldsymbol{\Sigma})\right)$, where $z_1 = \sigma_1 e_1$ and $z_2 = \sigma_2 \sqrt{1-\rho^2} e_2$ (i.e., $|v'| = \sigma_2 \sqrt{1-\rho^2} |v|$ with $|\cdot|$ representing the length of line segments), see \cref{fig:ConditionalGaussian}(b).
This transformation changes the scale of horizontal and vertical axes of circular contours of $\mathcal{N}({\bf 0},{\bf I})$. Thus, the means conditioned on $e_1$ (marked by ${\bf 1}$ in \cref{fig:ConditionalGaussian}(a)) remains unchanged and leads to the conditional Gaussian distribution $z_1|z_2 \sim \mathcal{N}(0,\sigma_2^2 (1-\rho^2))$.
Since $z_1 = \sigma_1 e_1$, resulting in $\rho \sigma_2 e_1 = \rho \sigma_2 \frac{z_1}{\sigma_1}$, the second term transforms ${\bf z} = [z_1, z_2]^\intercal \sim \mathcal{N}({\bf 0}, \diag\left( \boldsymbol{\Sigma})\right)$ obtained from the first transformation into ${\bf f}- {\bf m} = [f_1-m_1, f_2-m_2]^\intercal =  [z_1,z_2 + \rho \frac{\sigma_2}{\sigma_1} z_1]^\intercal \sim \mathcal{N}\left( {\bf 0}, \boldsymbol{\Sigma}\right)$ by vertically lifting any points $(z_1,z_2)$ an amount of $\rho \frac{\sigma_2}{\sigma_1} z_1$, see \cref{fig:ConditionalGaussian}(c).
This adds $\rho \frac{\sigma_2}{\sigma_1} z_1 = \rho \frac{\sigma_2}{\sigma_1} (f_1-m_1)$ to the mean of $z_2|z_1$, marked by ${\bf 1}'$ in \cref{fig:ConditionalGaussian}(b), and remains the variance unaltered.
As a result, the conditional Gaussian distribution is $f_2-m_2|f_1-m_1 \sim \mathcal{N}\left(\rho \frac{\sigma_2}{\sigma_1} (f_1-m_1),\sigma_2^2 (1-\rho^2)\right)$ and constant $c$ in \cref{fig:ConditionalGaussian}(c) is $\rho \sigma_2 / \sigma_1$.

Having the conditional Gaussian distribution $f_2-m_2|f_1-m_1 \sim \mathcal{N}\left(\rho \frac{\sigma_2}{\sigma_1} (f_1-m_1),\sigma_2^2 (1-\rho^2)\right)$, it is straightforward to recover the conditional mean and conditional variance of $f_2|f_1 \sim \mathcal{N}\left( \mu_{2|1},\Sigma_{2|1} \right)$. 
Specifically, we have 
\begin{equation}
	\begin{aligned}
		\mu_{2|1} & = m_2 + \rho \frac{\sigma_2}{\sigma_1} (f_1-m_1)  
                   = m_2 + \Sigma_{21} \Sigma_{11}^{-1} (f_1-m_1), \\
		\Sigma_{2|1} & = \sigma_2^2 (1-\rho^2)  
                       = \Sigma_{22} (1-\Sigma_{21} \Sigma_{11}^{-1} \Sigma_{22}^{-1} \Sigma_{12}) 
                      = \Sigma_{22} - \Sigma_{21} \Sigma_{11}^{-1} \Sigma_{12}.
	\end{aligned}
\end{equation}
These results agree with those provided in \cref{eqn:conditionalmeancovariance}.

\section{Spectral density of Matérn covariance functions}\label{AppB}
\setcounter{equation}{0} %
\renewcommand{\theequation}{B.\arabic{equation}} %

Consider the Matérn class of covariance functions $\kappa({\bf x},{\bf x}')$ in \cref{Materncovariancefunction}.
Since $\kappa({\bf x},{\bf x}')$ is stationary, its spectral density can be derived from the second expression of \cref{eqn:Bochner}, as \citep[Theorem 6.13]{Wendland2004}
\begin{equation}
    S(\boldsymbol{\omega}) = \sigma_\mathrm{f}^2 \frac{(2\pi)^{d/2} (2\nu)^\nu \Gamma\left(\nu + \frac{d}{2}\right)}{\Gamma(\nu)} |\boldsymbol{\Lambda}|^{1/2} \left( 2\nu + \boldsymbol{\omega}^\intercal \boldsymbol{\Lambda} \boldsymbol{\omega} \right)^{-\left(\nu + \frac{d}{2}\right)},
\end{equation}
where $|\boldsymbol{\Lambda}|$ is the determinant of $\boldsymbol{\Lambda}$.

The spectral PDF corresponding to $S(\boldsymbol{\omega})$ is
\begin{equation}
\begin{aligned}
    p(\boldsymbol{\omega}) = \frac{S(\boldsymbol{\omega})}{\sigma_\mathrm{f}^2 (2\pi)^d} & =  \frac{(2\nu)^\nu \Gamma\left(\nu + \frac{d}{2}\right)}{(2\pi)^{d/2} \Gamma(\nu)} |\boldsymbol{\Lambda}|^{1/2} \left( 2\nu + \boldsymbol{\omega}^\intercal \boldsymbol{\Lambda} \boldsymbol{\omega} \right)^{-\left(\nu + \frac{d}{2}\right)}\\
    &= \frac{(2\nu)^\nu \Gamma\left(\nu + \frac{d}{2}\right)}{(2\pi)^{d/2} \Gamma(\nu)} |\boldsymbol{\Lambda}|^{1/2} \left( 2 \nu\right)^{-\left(\nu + \frac{d}{2}\right)} \left( 1 + \frac{1}{2\nu} \boldsymbol{\omega}^\intercal \boldsymbol{\Lambda} \boldsymbol{\omega} \right)^{-\left(\nu + \frac{d}{2}\right)}\\
    &= \frac{\Gamma\left(\nu + \frac{d}{2}\right)}{(2\pi)^{d/2} \Gamma(\nu) (2\nu)^{d/2}} |\boldsymbol{\Lambda}|^{1/2} \left( 1 + \frac{1}{2\nu} \boldsymbol{\omega}^\intercal \boldsymbol{\Lambda} \boldsymbol{\omega} \right)^{-\left(\nu + \frac{d}{2}\right)}.
\end{aligned}
\end{equation}
Let $\boldsymbol{\Sigma} = \boldsymbol{\Lambda}^{-1}$ and $\nu_1 = 2\nu$. The spectral PDF can be written as
\begin{equation} \label{MaternspectralPDF}
    p(\boldsymbol{\omega}) = \frac{\Gamma\left(\frac{\nu_1+d}{2}\right)}{(2\pi)^{d/2} \Gamma(\nu_1/2) \nu_1^{d/2} |\boldsymbol{\Sigma}|^{1/2}} \left( 1 + \frac{1}{\nu_1} \boldsymbol{\omega}^\intercal \boldsymbol{\Sigma}^{-1} \boldsymbol{\omega} \right)^{-\left(\frac{\nu_1+d}{2}\right)}.
\end{equation}
This is a $d$-variate Student's $t$-distribution with degrees of freedom $\nu_1=2\nu$, location vector $\boldsymbol{\mu}={\bf 0}$, and scale matrix $\boldsymbol{\Sigma} = \mathrm{diag}\left([l_1^{-2}, \dots, l_d^{-2}]^\intercal\right)$.

\section{Benchmark functions for single-objective optimization}\label{AppC}
\setcounter{equation}{0} %
\renewcommand{\theequation}{C.\arabic{equation}} %

The analytical expressions and the global minima of the benchmark functions used in \Cref{{sec82}} are given below \citep{Surjanovic2013}.

\paragraph{Schwefel function}
\begin{equation}
	c({\bf x}) = 418.9829 d - \sum_{i=1}^{d} x_i \sin \left( \sqrt{|x_i|} \right).
\end{equation}
This function is evaluated on $\mathcal{X}=[-500,500]^d$ and has a global minimum $c^\star = c({\bf x}^\star) = 0$ at ${\bf x}^\star = [420.9687,\dots,420.9687]^\intercal$ .

\paragraph{Rosenbrock function}
\begin{equation}
	c({\bf x}) = \sum_{i=1}^{d-1} \left[100(x_{i+1}-x_i^2)^2 + (x_i - 1)^2\right].
\end{equation}
This function is evaluated on $\mathcal{X}=[-5,10]^d$ and has a global minimum at ${\bf x}^\star = [1,\dots,1]^\intercal$ with $c^\star = c({\bf x}^\star) = 0$.

\paragraph{Levy function}
\begin{equation}
f(\mathbf{x}) = \sin^2 (\pi w_1) + \sum_{i=1}^{d-1} (w_i - 1)^2 \left[ 1 + 10 \sin^2 (\pi w_i +1) \right] + (w_d - 1)^2 \left[ 1 +  \sin^2 (2 \pi w_d) \right],
\end{equation}
where $w_i = 1 + \frac{x_i-1}{4}$, $i = 1,\cdots,d$.
The function is evaluated on $\mathcal{X}=[-10,10]^d$
and has a global minimum $c^\star = c({\bf x}^\star)= 0$ at $\mathbf{x}^\star = [1,\cdots,1]^\intercal$.

\paragraph{Ackley function}
\begin{equation}
	c({\bf x}) =  -a \exp\left(-b\sqrt{\frac{1}{d} \sum_{i=1}^{d}x^2_i}\right) \\
    - \exp\left(\frac{1}{d} \sum_{i=1}^{d}\cos(hx_i)\right) +a + \exp(1),
\end{equation}
where $a = 20$, $b = 0.2$, and $h = 2\pi$.
The function is evaluated on $\mathcal{X}=[-10,10]^d$ and has a global minimum $c^\star = c({\bf x}^\star) = 0$ at ${\bf x}^\star = [0,\dots,0]^\intercal$.

\section{Airfoil shape optimization problem} \label{AppD} 
\setcounter{equation}{0} %
\setcounter{table}{0}
\renewcommand{\theequation}{D.\arabic{equation}} 
\setcounter{figure}{0} %
\renewcommand{\thefigure}{D.\arabic{figure}}
\renewcommand{\thetable}{D.\arabic{table}}

\begin{figure}[t]
	\centering
	\includegraphics[width=0.5\textwidth]{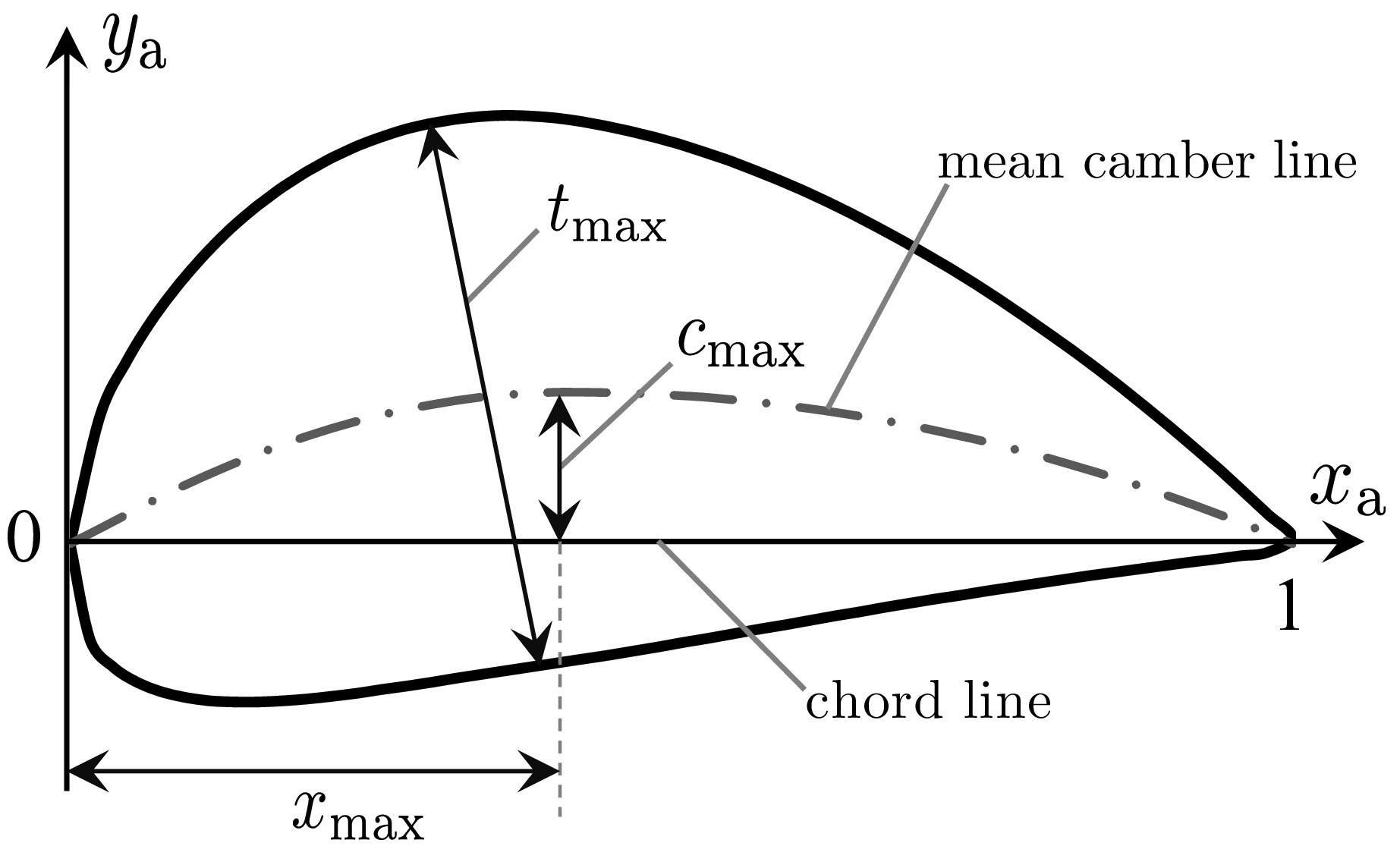}
	\caption{Shape parameters for NACA 4-digit airfoil, adapted from \cite{Do2025mfbo}.}
    \label{fig:airfoilshape}
\end{figure}

The shape optimization problem formulated for the NACA 4-digit airfoil \citep{Jacobs1933} in \cref{sec82} is to maximize its lift-to-drag ratio for finding its optimal shapes.
As shown in \cref{fig:airfoilshape}, the airfoil shape is parameterized by three parameters: $c_{\max}$, $x_{\max}$, and $t_{\max}$. Here, $c_{\max}$ is the height measured from the chord line to the point where the mean camber line has the largest absolute curvature value, $x_{\max}$ the position of $c_{\max}$ in horizontal coordinates, and $t_{\max}$ the maximum thickness of the airfoil.
The parameterization equations are provided in \cite{Do2025mfbo}, Appendix D.
In this problem, we express the parameters in terms of their ratios to the chord length with design domains of $c_{\max} \in [0, 0.08]$, $x_{\max} \in [0, 0.8]$, and $t_{\max} \in [0.1, 0.25]$.

\begin{table*}[t]
    \centering
    \caption{Operating conditions and weightings for calculating the multi-point lift and drag coefficients.}
    \begin{tabular}{ccccc}
         \toprule
      Condition & Mach no. & Angle of attack ($^{\circ}$) & Reynolds no. & Weight value $w_i$\\
      \midrule
        $1$ & $0.5$ & $0$ & $6.3 \times 10^6$ &  0.4 \\
        $2$ & $0.55$ & $0.2$ & $6.3 \times 10^6$ &  0.2 \\
        $3$ & $0.5$ & $0.5$ & $6.3 \times 10^6$ &  0.2 \\
        $4$ & $0.55$ & $0$ & $6.3 \times 10^6$ &  0.2 \\
        \bottomrule
    \end{tabular}
    \label{table:3}
\end{table*}

We adopt the multi-point design philosophy \citep{Toal2023} for the airfoil. 
This design philosophy is to find a shape that performs well across various operating conditions, rather than at a specific design condition.
More specifically, we consider four distinct operating conditions, as listed in \cref{table:3}, where the airfoil is expected to operate well in four different combinations of speeds and angles of attack while the Reynolds number remains constant.
As a result, we define the objective function $c({\bf x})$ for the shape optimization problem as the weighted sum of negative lift-to-drag ratios associated with the given operating conditions, which reads
\begin{equation}
    c({\bf x})= -\sum_{i=1}^{4} w_i \frac{c_{\text{L},i}({\bf x})}{c_{\text{D},i}({\bf x})},
\end{equation}
where ${\bf x} = \left[ c_{\max}, x_{\max}, t_{\max} \right]^\intercal$, $c_{\text{L},i}({\bf x})$ and $c_{\text{D},i}({\bf x})$ represent the lift and drag coefficients associated with the $i$th operating condition, respectively, $w_i$ listed in \cref{table:3} is the weight corresponding to the $i$th operating condition, and the negative sign is to transform a maximization problem of the lift-to-drag ratio into a minimization problem of $c({\bf x})$.
To compute the lift and drag coefficients at each operating condition, we implement the 2D-panel code XFOIL \citep{Drela1989}.
Details on the fluid flow model for XFOIL can be found in \cite{Drela1987,Drela1989}.

\newpage
\section{Benchmark problems for multi-objective optimization}\label{AppE}
\setcounter{equation}{0} %
\renewcommand{\theequation}{E.\arabic{equation}} %
The benchmark multi-objective optimization problems used in \cref{sec83} are given below.

\paragraph{KNO1} 
This problem has two objective functions as follows \citep{Knowles2006}:
\begin{equation} 
    \begin{aligned}
        c_1({\bf x}) & = 20 - r \cos(\phi),\\
        c_2({\bf x}) & = 20 - r \sin(\phi),
    \end{aligned}
\end{equation}
where ${\bf x} = [x_1,x_2]^\intercal$ with $x_1, x_2 \in [0,3]$, and
\begin{equation} 
    \begin{aligned}
        r = & 9 - \left[ 3 \sin \left( \frac{5}{2(x_1 + x_2)^2}\right) + 3 \sin\left( 4(x_1 + x_2) \right)
         + 5 \sin\left( 2(x_1 + x_2) +2\right)\right],\\
       \phi = & \frac{\pi}{12 \left( x_1 - x_2 + 3\right)}.
    \end{aligned}
\end{equation}

\paragraph{VLMOP2} 
Two objective functions of this problem are \citep{vanVeldhuizen1999}
\begin{equation} 
    \begin{aligned}
        c_1({\bf x}) & = 1 - \exp \left( - \sum_{i=1}^{2} \left( x_i - \frac{1}{\sqrt{2}} \right)^2 \right),\\
        c_2({\bf x}) & = 1 - \exp \left( - \sum_{i=1}^{2} \left( x_i + \frac{1}{\sqrt{2}} \right)^2 \right),
    \end{aligned}
\end{equation}
where ${\bf x} = [x_1,x_2]^\intercal$ with $x_1, x_2 \in [-2,2]$.

\paragraph{VLMOP3} 
Three objective functions of this problem are \citep{vanVeldhuizen1999}
\begin{equation} 
    \begin{aligned}
        c_1({\bf x}) & = 0.5(x_1^2 + x_2^2) + \sin(x_1^2 + x_2^2),\\
        c_2({\bf x}) & = \frac{(3x_1 - 2x_2 + 4)^2}{8} + \frac{(x_1 - x_2 +1)^2}{27}+ 15,\\
        c_3({\bf x}) & = \frac{1}{x_1^2 + x_2^2 +1} - 1.1 \exp (-x_1^2 - x_2^2),
    \end{aligned}
\end{equation}
where ${\bf x} = [x_1,x_2]^\intercal$ with $x_1, x_2 \in [-3,3]$.

\paragraph{DTLZ2a} 
Three objective functions of this problem are given as follows \citep{Knowles2006}:
\begin{equation} 
    \begin{aligned}
        c_1({\bf x}) & = (1+g) \cos \left( \frac{x_1 \pi}{2} \right) \cos \left( \frac{x_2\pi}{2} \right),\\
        c_2({\bf x}) & = (1+g) \cos \left( \frac{x_1 \pi}{2} \right) \sin \left( \frac{x_2\pi}{2} \right),\\
        c_3({\bf x}) & = (1+g) \sin \left( \frac{x_1 \pi}{2} \right),
    \end{aligned}
\end{equation}
where $x_i \in [0,1]$, $i \in \{ 1,\dots,8\}$ and $g = \sum_{i=3}^8 (x_i - 0.5)^2.$ 

\newpage

\section{Additional results}\label{AppF}
\setcounter{figure}{0} %
\renewcommand{\thefigure}{F.\arabic{figure}}

\begin{figure}[h!]
	\centering
	\includegraphics[width=0.55\textwidth]{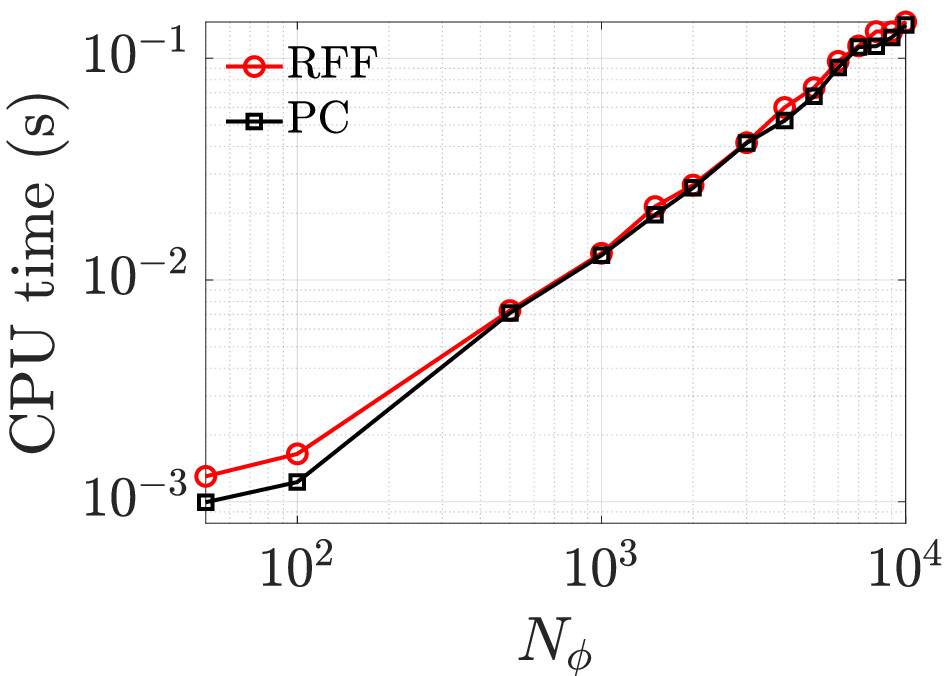}
	\caption{Comparison of computational cost of \texttt{GP-RFF} and \texttt{GP-PC} for different numbers of RFFs when the posterior samples of the weights are generated by an acceleration technique that uses the SMW formula and the Cholesky decomposition of an $N$-by-$N$ matrix.}
    \label{fig:timeRFFPCacceleration}
\end{figure}

\Cref{fig:timeRFFPCacceleration} compares the computational cost of \texttt{GP-RFF} and \texttt{GP-PC} for different numbers of RFFs when the posterior samples of the weights are generated by an acceleration technique that uses the SMW formula and the Cholesky decomposition of an $N$-by-$N$ matrix \citep{Seeger2007}.
We see that $\texttt{GP-RFF}$ can obtain a linear cost in $N_\phi$. However, it is still slightly slower than $\texttt{GP-PC}$.

\begin{figure*}[t]
	\centering
	\includegraphics[width=0.825\textwidth]{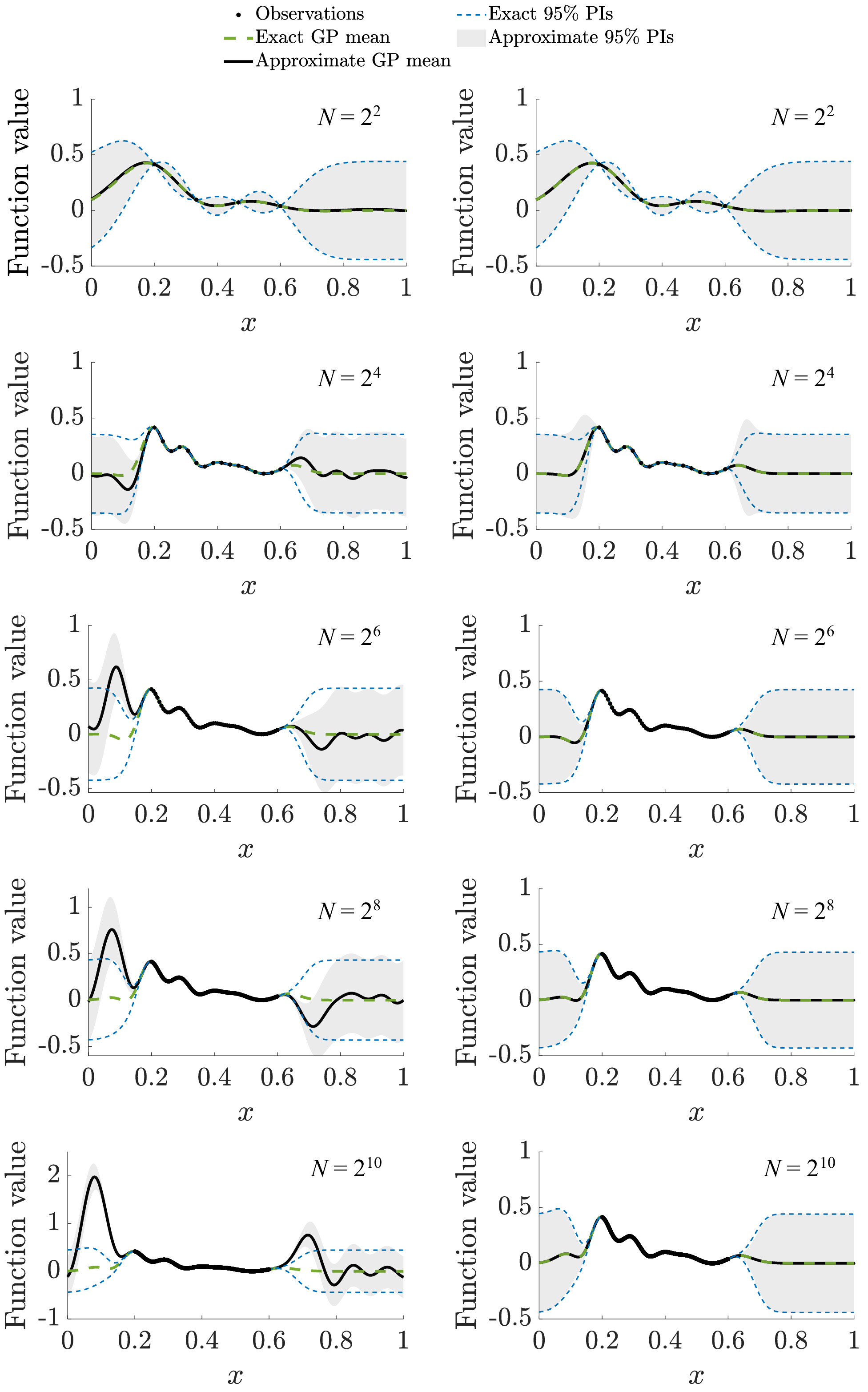}
	\caption{Mean and predictive intervals of the exact posterior compared with those of approximations by $\texttt{GP-RFF}$ and $\texttt{GP-PC}$ for different numbers of training points in the interpolation region. \textit{Left column:} Approximations by $\texttt{GP-RFF}$. \textit{Right column:} Approximations by $\texttt{GP-PC}$.}
    \label{fig:prediction_rff_pc}
\end{figure*}

\Cref{fig:prediction_rff_pc} compares the prediction performance of $\texttt{GP-RFF}$ and $\texttt{GP-PC}$ on the 1D Levy function described in \cref{sec4.4} for different numbers of training points.
While the extrapolation performance of $\texttt{GP-RFF}$ deteriorates as the number of training points increases, that of $\texttt{GP-PC}$ remains accurate and stable.

\clearpage
\bibliographystyle{elsarticle-num-names}
\bibliography{GPSampling} 

\end{document}